\PassOptionsToPackage{dvipsnames}{xcolor}
\documentclass[sensors,article,accept,pdftex,moreauthors]{Definitions/mdpi} 

\firstpage{1} 
\pubvolume{26}
\issuenum{6}
\articlenumber{1798}
\pubyear{2026}
\copyrightyear{2026}
\externaleditor{Jun Chen and Wen-Chiao Lin} 
\datereceived{30 January 2026 } 
\daterevised{5 March 2026 } 
\dateaccepted{9 March 2026 } 
\datepublished{12 March 2026} 
\usepackage{algorithmic}
\usepackage{algorithm}

\makeatletter
\newcommand*{\textoverline}[1]{$\overline{\hbox{#1}}\m@th$}
\makeatother

\Title{PLM-Net: Perception Latency Mitigation Network for Vision-Based Lateral Control of Autonomous Vehicles}

\Author{{Aws} 
 Khalil \orcidA{} and Jaerock Kwon *\orcidB{}}

\AuthorNames{Aws Khalil and Jaerock Kwon}

\address[1]{%
Department of Electrical and Computer Engineering, University of Michigan-Dearborn, \mbox{Dearborn, MI 48128, USA;} awskh@umich.edu\\
}
\corres{\hspace{-0.8em}Correspondence: jrkwon@umich.edu}

\abstract{
This study introduces the Perception Latency Mitigation Network (PLM-Net), a modular deep learning framework designed to mitigate perception latency in vision-based imitation-learning lane-keeping systems. Perception latency, defined as the delay between visual sensing and steering actuation, can degrade lateral tracking performance and steering stability. 
While delay compensation has been extensively studied in classical predictive control systems, its treatment within vision-based imitation-learning architectures under constant and time-varying perception latency remains limited.
Rather than reducing latency itself, PLM-Net mitigates its effect on control performance through a plug-in architecture that preserves the original control pipeline.
The framework consists of a frozen Base Model (BM), representing an existing lane-keeping controller, and a Timed Action Prediction Model (TAPM), which predicts future steering actions corresponding to discrete latency conditions. Real-time mitigation is achieved by interpolating between model outputs according to the measured latency value, enabling adaptation to both constant and time-varying latency.
The framework is evaluated in a closed-loop deterministic simulation environment under fixed-speed conditions to isolate the impact of perception latency. Results demonstrate significant reductions in steering error under multiple latency settings, achieving up to $62\%$ and $78\%$ reductions in Mean Absolute Error (MAE) for constant and time-varying latency cases, respectively. These findings demonstrate the architectural feasibility of modular latency mitigation for vision-based lateral control under controlled simulation settings.
The project page including video demonstrations, code, and dataset is publicly released.
}


\begin{document}



\section{Introduction}
\label{sec:introduction}

\subsection{Motivation}
\label{subsec:1-a}

Vision-based Autonomous Vehicle (AV) control follows a perception--planning--control (sense–think–act) cycle, where visual observations are processed to generate control actions.
In this cycle, there is a latency between sensing the environment and applying a corresponding action, which makes the human reaction time always higher than zero~\cite{kwon2007enhanced}. 
Similarly, it is challenging to completely eliminate this latency in AV control \cite{li2020towards}, and reducing it through powerful GPUs and FPGAs is impractical in automotive platforms. 
If not properly mitigated, this latency can degrade lateral tracking performance and steering stability, potentially affecting ride comfort and control reliability.
Human drivers implicitly anticipate future vehicle states when reacting to visual stimuli \cite{ho2022cognitive}. Inspired by this predictive behavior, we propose a deep neural network architecture that forecasts future steering actions to mitigate perception latency.

\begin{figure}[H]
\includegraphics[width=0.75\columnwidth]{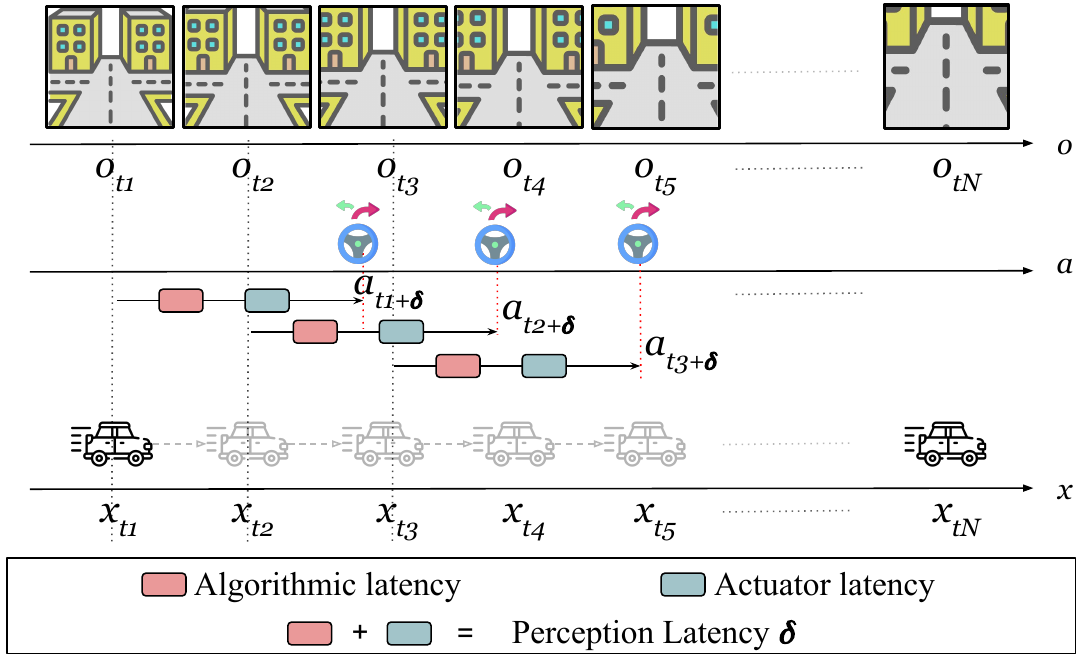}
\caption{Perception 
 latency definition. When vehicle state is \textit{$x_t$} and we have observation \textit{$o_t$}, the corresponding action \textit{$a_t$} is applied at time \textit{$t+\delta$} not \textit{$t$} ($o_t \rightarrow a_{t+\delta}$). By the time this action is applied, the vehicle state has changed, and we have a new observation. The perception latency has two components: the algorithmic latency and the actuator latency.}
\label{fig:perception-latency-def}
\end{figure}

In this paper, we refer to this latency as the perception 
 latency $\delta$, as shown in Figure~\ref{fig:perception-latency-def}.
When vehicle state is \textit{$x_t$} and we have observation \textit{$o_t$}, the corresponding action \textit{$a_t$} is applied at time \textit{$t+\delta$} rather than at time \textit{$t$} ($o_t \rightarrow a_{t+\delta}$). By the time this action is applied, the vehicle state has changed and we have a new observation.
The perception latency has two components: the algorithmic latency, which is the time required for the algorithm to infer an action from an observation, and the actuator latency, which is the time required to apply the inferred action. The actuator latency, known as steering lag in lateral control \cite{xu2020preview}, can be considered constant \cite{xu2020preview}. However, as mentioned in \cite{liu2018hierarchical}, the high processing cost of visual algorithms leads to uneven time delays based on the driving scenario, which leads to an overall perception latency that is time-varying.

To address both constant and time-varying perception latency encountered during lane keeping, we propose a deep-learning-based approach, focusing primarily on vision-based AV lateral control. 
The contribution of this paper will be discussed after explaining the effect of the perception latency on AV lateral control.

\subsection{Latency Effect on AV Lateral Control}
\label{subsec:latency-effect-on-av-lateral-control}
Vision-based AV lateral control for lane keeping can be achieved using various methods. The traditional approach involves incorporating a computer vision module for lane-marking detection alongside a classical control module for path planning and control. Alternatively, a deep-learning-based approach, such as imitation learning \cite{chen_end--end_2017}, directly maps visual input to control actions, like steering angle. This paper adopts the latter method.

The effect of perception latency on AV lateral control during lane keeping is highly dependent on vehicle speed. 
In this study, vehicle speed is held constant during evaluation in order to isolate the effect of perception latency independently of speed-induced dynamic variations.
At low speeds, the scene does not change significantly between the time the vehicle receives the observation at time \textit{$t$} and the time it applies the action at time \textit{$t+\delta$}, because the traveled distance during the latency period is minimal ($d = \delta v$). In this scenario, it is reasonable to assume that $a_t \approx a_{t+\delta}$, making the effect of \textit{$\delta$} negligible. Thus, at low speeds, we can consider that at time $t$, the action $a_t$ corresponding to observation $o_t$ is applied immediately ($o_t \rightarrow a_t$).

\begin{figure}[H]
\includegraphics[width=0.6\columnwidth]{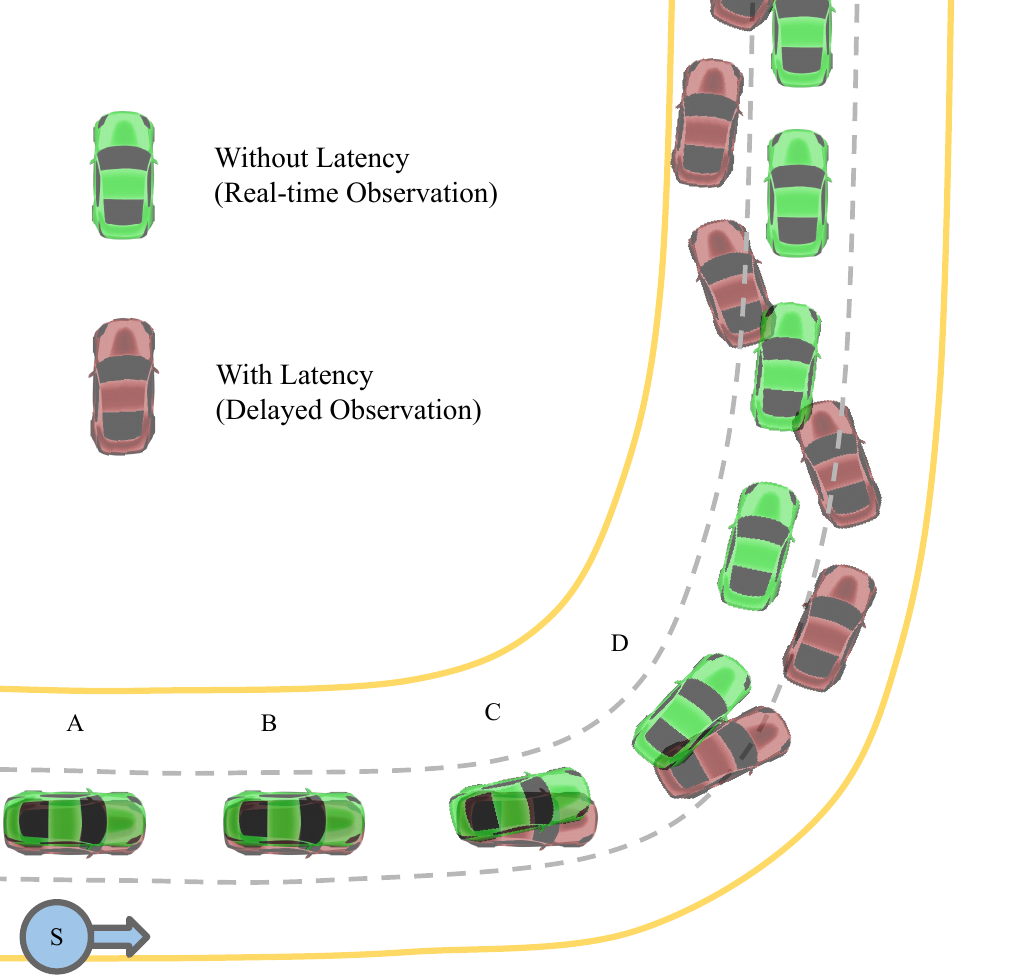}
\caption{Illustration 
 of perception latency effect on lateral control during lane keeping. The green vehicle exhibits driving behavior with real-time observations, while the red vehicle demonstrates driving behavior with a delayed observation input. 
Both vehicles maintain a constant low speed. $S$ denotes the start point, and the arrow indicates the direction of vehicle motion. Motion starts at position $A$. 
Positions $A$ and $B$ depict both vehicles driving straight, staying within their lanes, with minimal steering adjustments ($a_{t_A} \approx a_{t_B} \approx 0.0$). 
At position $C$, where the first curve is encountered, the green vehicle successfully adjusts its trajectory by turning left to stay in the lane, while the red vehicle continues straight due to the delayed observation input. Subsequently, from position $D$ onward, the red vehicle struggles to recover from its zigzag-shaped trajectory, highlighting the impact of incorrect actions on subsequent observations.}
\label{fig:driving-with-latency}
\end{figure}

The effect of latency on AV control during lane keeping is illustrated in Figure~\ref{fig:driving-with-latency}. 
Two vehicles drive at a constant low speed: the green vehicle uses the normal real-time observation, assuming zero latency, while the red vehicle uses the delayed observation, mimicking the perception latency effect. 
Both vehicles start from position $A$ and move towards position $D$. If we timestamp each position, then $t_A < t_B < t_C < t_D$.
For the green vehicle at position $B$, the available observation is $o_{t_B}$ and the corresponding action is $a_{t_B}$. For the red vehicle at the same position, the available observation is $o_{t_B-\delta}$ and the corresponding action is $a_{t_B-\delta}$. This pattern continues for the other positions.
Both vehicles remain in their lanes at positions $A$ and $B$, before encountering any curves, because they started within the lane and continued straight, where the steering angle is zero. Thus, at positions $A$ and $B$, the actions $a_t$ and $a_{t-\delta}$ are effectively the same.
At position $C$, the first curve is encountered. The green vehicle successfully turns left to stay in the lane, but the red vehicle continues straight. This deviation occurs because the red vehicle's input observation at position $C$ is $o_{t_C-\delta}$, not $o_{t_C}$, leading it to apply the action $a_{t_C-\delta}$ at position $C$. After this first curve, the red vehicle's zigzag trajectory becomes difficult to correct, even on a straight road, due to the incorrect action taken at position $C$, which causes subsequent incorrect observations.

In this approach, to mitigate the perception latency and avoid this unstable driving behavior, the main objective would be to predict the correct action from the delayed observation input.

\subsection{Contribution}

\begin{figure}[H]
\includegraphics[width=1\linewidth]{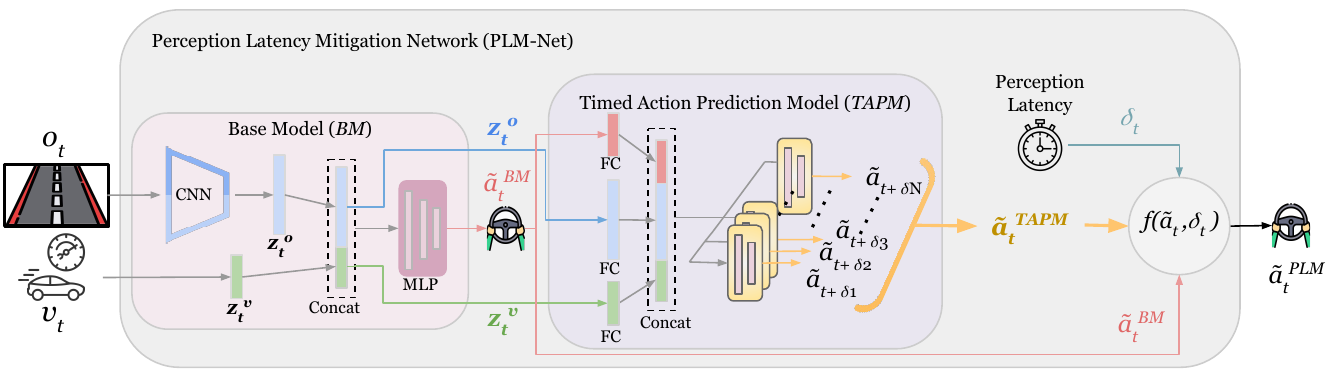}
\caption{
 Overview of the perception latency mitigation network (PLM-Net). By leveraging a Timed Action Prediction Model (TAPM) alongside the Base Model (BM), PLM-Net enhances the system's ability to mitigate perception latency. The TAPM incorporates predictive modeling to anticipate future actions based on current visual observations. Additionally, the integration of multiple sub-models within the framework enables adaptation to varying latency levels, as the final action value $\widetilde{a}^{\text{PLM}}_t$ is determined through the function $f(\widetilde{a}_t,\delta_t)$ where it performs linear interpolation based on the real-time latency value $\delta_t$ given all the predictive action values provided by the TAPM ($\boldsymbol{\widetilde{a}}^{\text{TAPM}}_t$) and the current action value provided by the BM ($\widetilde{a}^{\text{BM}}_t$).
 See Section \ref{sec:method} for detailed explanation.}
\label{fig:method-intro}
\end{figure}

We introduce the Perception Latency Mitigation Network (PLM-Net), outlined in Figure~\ref{fig:method-intro}. 
This novel deep learning approach is intended to work easily and without requiring any changes to the original vision-based LKA system of the AV. 
As depicted in Figure~\ref{fig:method-intro}, PLM-Net leverages a Timed Action Prediction Model (TAPM) alongside the Base Model (BM), where the latter represents the preexisting vision-based LKA system. 
The design of the TAPM is inspired by our prior work, \textit{ANEC} \cite{khalil2023anec}, and the Branched Conditional Imitation Learning model (\textit{BCIL}) proposed by Codevilla et al. \cite{codevilla2018ConditionalImitation}. 
It combines the concept of a predictive model capable of forecasting future action from current visual observation, akin to \textit{ANEC} \cite{khalil2023anec} (originally inspired by the human driver capability of dealing with the perception latency \cite{salvucci2006modeling}), with the notion of employing multiple sub-models within the \textit{BCIL} framework \cite{codevilla2018ConditionalImitation} to provide different predictive action values corresponding to different latency levels.
Additionally, similar to the ‘command’ input used in \textit{BCIL} to select a sub-model, the real-time perception latency $\delta_t$ is used to determine the final action value.
The final action value $\widetilde{a}^{\text{PLM}}_t$ is determined through the function $f(\widetilde{a}_t,\delta_t)$ where it performs linear interpolation based on the real-time latency value $\delta_t$ given all the predictive action values provided by the TAPM ($\boldsymbol{\widetilde{a}}^{\text{TAPM}}_t$) and the current action value provided by the BM ($\widetilde{a}^{\text{BM}}_t$).
A comprehensive explanation of our proposed method is provided in Section \ref{sec:method}. The main contributions of this paper are summarized below.

\begin{enumerate}[labelsep=-2mm]

\item[]{Main contributions:}

\begin{itemize}
    \item We formulate perception latency in vision-based imitation-learning lane keeping as a time-offset control problem, and analyze how delayed observations degrade steering stability and lateral tracking performance;
    
    \item We propose PLM-Net, a modular plug-in latency-mitigation framework that augments an existing imitation-learning lane-keeping controller without modifying or retraining the base policy, thereby preserving its deployment characteristics;
    
    \item We introduce the Timed Action Prediction Model (TAPM), a latency-conditioned multi-head predictive module that produces discrete future steering actions indexed by delay values, enabling mitigation of both constant and time-varying latency through runtime interpolation based on measured latency;
    
    \item We validate the proposed framework in a deterministic closed-loop simulation environment under fixed-speed conditions to isolate latency effects, demonstrating substantial improvements in steering similarity and trajectory stability across multiple latency settings.
\end{itemize}\end{enumerate}

The structure of this paper proceeds as follows: Section~\ref{sec:related-work} presents an overview of related research, while Section~\ref{sec:method} outlines the proposed methodology. Following this, Sections~\ref{sec:setup} and \ref{sec:results} present the experimental setup and the experimental findings within the simulation environment, accompanied by a comprehensive analysis. Finally, Section~\ref{sec:conclusion} summarizes the core findings of the study and provides valuable insights into potential avenues for future research.

\section{Related Work}
\label{sec:related-work}

In the domain of vision-based control, the explicit modeling and mitigation of perception latency have received comparatively limited attention relative to other aspects of autonomous driving control.
In classical control methods, discussions surrounding latency in autonomous driving have largely revolved around computational delays associated with hardware deployment \cite{berntorp2018control, lee2022probabilistically, strobel2020accurate, vedder2018low} and communication delays related to network performance~\cite{ge2019ultra, el2020smooth, pokhrel2021towards, zhang2020ac4av, gorsich2018evaluating, yaqoob2019autonomous}.
While several classical control studies have addressed perception or input delay in driving systems, these efforts are relatively sparse compared to the broader literature on delay-aware predictive control.
Xu et al. \cite{xu2020preview} modeled steering lag as a fixed 200 ms delay, while Liu et al. \cite{liu2018hierarchical} proposed a hierarchical MPC framework to compensate for time-variant input delays on the order of several hundred milliseconds. More recently, Kalaria et al. \cite{kalaria2022delay, kalaria2024delay} developed robust control strategies to handle fixed perception delays (e.g., 200 ms), demonstrating safety improvements for both lane-keeping and racing tasks.
In the context of networked and teleoperated autonomous vehicles, Kamtam et al. \cite{kamtam2024network} reviewed delay patterns between 100 and 350 ms across vision and communication channels.
Recent system-level studies have also begun to address the impact of latency variability on autonomous driving pipelines. Han and Kim \cite{han2023minimizing} propose probabilistic scheduling techniques to minimize end-to-end perception--planning--control latency in real-time AV systems.
While valuable, these approaches primarily aim to reduce latency, while our work focuses on learning to compensate for its effect on control behavior. Unlike classical or system-level approaches, PLM-Net introduces a learning-based method that adaptively mitigates both fixed and time-varying latency without relying on handcrafted dynamics or delay-tuned controllers.
Classical delay-aware control methods, including preview control and delay-compensated MPC, explicitly incorporate system dynamics and delay models into the control law and typically require accurate vehicle modeling, as well as controller redesign. In contrast, PLM-Net does not modify the original control structure nor require explicit vehicle dynamics modeling. Instead, it augments an already trained vision-based imitation-learning policy with a latency-conditioned predictive layer that operates as a plug-in module. Therefore, rather than replacing model-based delay compensation strategies, the proposed framework provides a complementary learning-based solution tailored to data-driven lateral control systems where analytical models or controller redesign may not be feasible.

Within vision-based neural network control models, research efforts have primarily focused on model architecture and performance optimization, with comparatively limited attention given to explicit modeling of the perception latency issue \cite{pomerleau_alvinn_1989, net-scale_autonomous_2004, chen_end--end_2017, bojarski2017explaining, wang_end--end_2019-1, wu_end--end_2019, kwon2022incremental, weiss2020deepracing, cultrera2023addressing, jamgochian2023shail, cheng2024pluto, delavari2025caril}.

Only few studies have discussed perception latency.
Li et al. \cite{li2020towards} underscored the significance of addressing latency in online vision-based perception systems. To tackle this issue, they introduced a methodology for assessing the real-time performance of perception systems, effectively balancing accuracy and latency. 
However, it is important to note that the paper primarily focuses on proposing a metric and benchmark for evaluating the real-time performance of perception systems, rather than offering a direct solution for vehicle control. While their approach provides valuable insights into quantifying the trade-off between accuracy and latency in perception systems, it does not directly address the challenges associated with mitigating latency in autonomous driving scenarios.
Khalil et al. \cite{khalil2023anec} addressed the perception latency issue by proposing the \textit{Adaptive Neural Ensemble Controller (ANEC)}. However, \textit{ANEC} assumed perception latency to be constant and did not address time-varying latency. Weighted sum was used to combine the output from the two driving models, but the weight function parameters had to be carefully chosen and adjusted by hand. 
This introduces environment-specific parameter tuning, which may limit straightforward deployment across different scenarios. Furthermore, ANEC does not explicitly model latency as a runtime-conditioned variable nor provide discrete latency-indexed predictive heads. Instead, it blends multiple policies through adaptive weighting. In contrast, PLM-Net formulates latency mitigation as an explicit mapping between measurable delay values and predicted future steering actions. This design separates the base driving policy from the latency-compensation layer, enabling deterministic interpolation across latency conditions while preserving the original controller parameters.
Mao et al. \cite{mao2019delay} explored latency in the context of video object detection by analyzing the detection latency of various video object detectors. While they introduced a metric to quantify latency, their study focused on measurement rather than proposing solutions to mitigate latency.
Kocic et al. \cite{kocic_end--end_2019} tried to decrease the latency in driving by altering the neural network architecture. Although this approach could potentially diminish latency, preserving the original accuracy poses a significant challenge.
Wu et~al.~\cite{wu2022trajectory} emphasized that control-based driving models, which convert images into control signals, inherently exhibit perception latency and are susceptible to failure due to their focus on the current time step. In response, they developed Trajectory-guided Control Prediction (TCP), a multi-task learning system integrating a control prediction model with a trajectory planning model. However, this approach necessitates the extraction of precise trajectories, presenting a notable challenge.
Popov et al. \cite{popov2024mitigating} introduced a latent space generative world model that, while not explicitly addressing latency, exhibited inherent robustness to it during deployment. This suggests that certain architectures may inadvertently compensate for latency, though without targeted mechanisms. In contrast, our method proactively models latency during both training and inference, allowing it to adaptively adjust actions based on real-time delay conditions.
Tampuu et al. \cite{tampuu2024effects} examined how delays and vehicle speed affect the test-time performance of end-to-end driving models, highlighting that even modest delay values can significantly impact behavior unless label alignment is addressed. However, their method focuses on mitigating symptoms of latency degradation, not latency modeling itself.

In our approach, we acknowledge the inevitability of latency and explicitly model its effect during both training and inference. By integrating latency-indexed predicted future actions with the current action based on the real-time latency value, the proposed method mitigates the impact of latency on steering behavior in vision-based imitation-learning lane~keeping.

\section{Method}
\label{sec:method}

This section describes the proposed Perception Latency Mitigation Network (PLM-Net) and its evaluation methodology. We first introduce the conceptual framework and explain how latency mitigation is achieved through the interaction between the Base Model (BM) and the Timed Action Prediction Model (TAPM). We then detail the architectural design of both models, followed by the training procedure and the performance metrics used to evaluate latency mitigation in vision-based lane keeping.

\subsection{PLM-Net Framework}

\begin{figure}[H]
\includegraphics[width=0.9\linewidth]{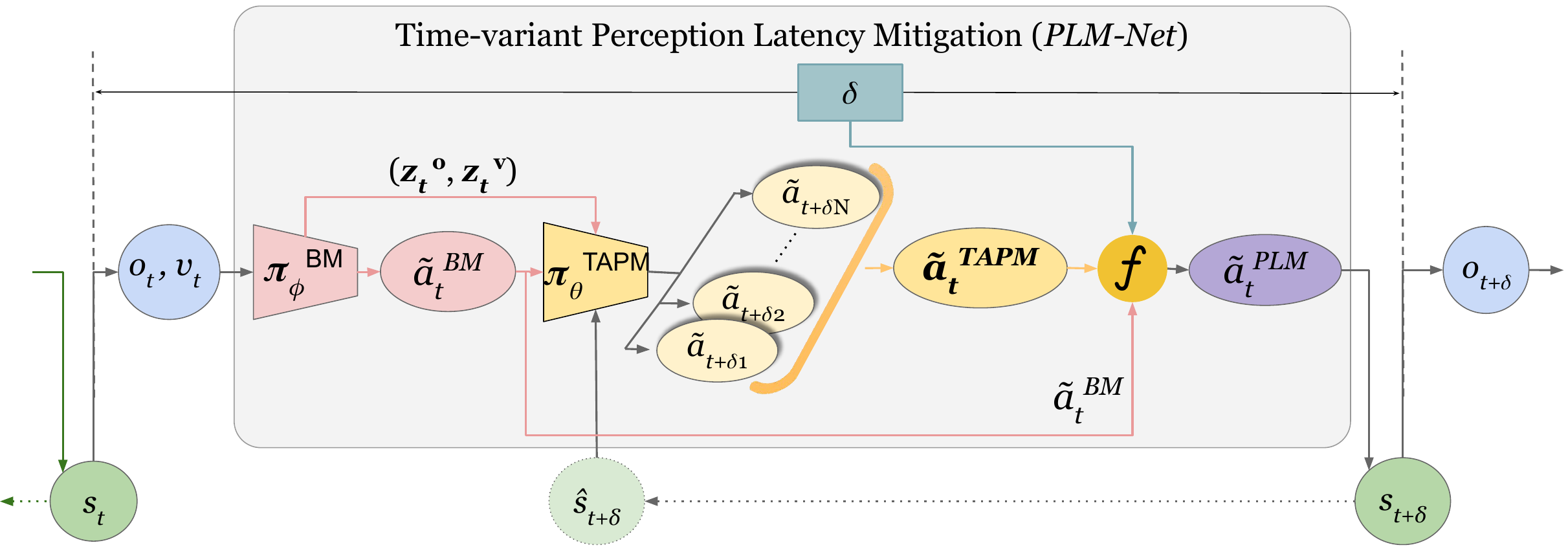}
\caption{PLM-Net 
diagram illustrating the interaction between the Base Model (BM) and the Timed Action Prediction Model (TAPM) in mitigating perception latency.
The BM, governed by policy $\pi^{\text{BM}}_\phi$, processes visual observation $o_t$ and vehicle speed $v_t$ to generate the nominal steering action $\widetilde{a}^{\text{BM}}_t$, as described in Equation~\eqref{eq:a_t-bm}. 
The TAPM, governed by policy $\pi^{\text{TAPM}}_\theta$, generates a set of latency-indexed predictive steering actions corresponding to predefined latency offsets in $\boldsymbol{\delta}^{\text{TAPM}}$ (e.g., $\widetilde{a}_{t+\delta_1}$ associated with the delayed operating state $s_{t+\delta_1}$), as detailed in Equation~\eqref{eq:a_t-tapm}. 
The inputs to $\pi^{\text{TAPM}}_\theta$ include $\widetilde{a}^{\text{BM}}_t$, along with feature vectors $\boldsymbol{z^{o}_t}$ and $\boldsymbol{z^{v}_t}$ extracted from the BM. 
The zero-latency action $\widetilde{a}^{\text{BM}}_t$ is appended to the latency grid to form the reference set used for interpolation.
Finally, linear interpolation (Algorithm~\ref{alg:linear_interpolation_latency}) combines the latency-indexed action grid according to the measured perception latency $\delta$ to produce the final mitigated action $\widetilde{a}^{\text{PLM}}_t$.
Green nodes represent vehicle states, blue nodes represent observations, red blocks indicate the Base Model processing stage, yellow blocks denote TAPM processing and predictive heads, and the purple node represents the final action produced by PLM-Net. Arrows illustrate the flow of information between modules.
}
\label{fig:PLM-Net-diagram}
\end{figure}

We begin by describing the overall framework of PLM-Net and the interaction between its components.
As shown in Figure~\ref{fig:method-intro}, the PLM-Net has two major components, the Base Model (BM) and the Timed Action Prediction Model (TAPM).
This novel deep learning approach smoothly integrates the TAPM with the original vision-based LKA system of the AV, represented by the BM. 
Figure~\ref{fig:PLM-Net-diagram} shows how these two models mitigate the perception latency, where $\pi^{\text{BM}}_\phi$ is the BM policy and $\pi^{\text{TAPM}}_\theta$ is the TAPM policy. 
As explained in \eqref{eq:a_t-bm}, given the vehicle state $s_t$ at time $t$, $\pi^{\text{BM}}_\phi$ takes the input 
 $\boldsymbol{i_t} = \{o_t, v_t\}$, where $o_t$ is the visual observation and $v_t$ is the vehicle speed at time $t$, and provides the action $\widetilde{a}^{\text{BM}}_t$. 

\begin{equation}
    \widetilde{a}^{\text{BM}}_t = \pi^{\text{BM}}_\phi(\boldsymbol{i_t}) =  \pi^{\text{BM}}_\phi(o_t, v_t).
\label{eq:a_t-bm}
\end{equation}

The TAPM is a predictive action model, meaning that the policy $\pi^{\text{TAPM}}_\theta$ generates a set of predictive action values 
 $\boldsymbol{\widetilde{a}}^{\text{TAPM}}_t$, where each action corresponds to a future operating state associated with a specific latency value in $\boldsymbol{\delta}^{\text{TAPM}}$.

The number of predictive actions in the vector $\boldsymbol{\widetilde{a}}^{\text{TAPM}}_t$ depends on the number of sub-models in TAPM. 
If there are $N$ sub-models, then $\boldsymbol{\delta}^{\text{TAPM}} = [\delta_1, \delta_2, \ldots , \delta_N]$. For example, if $\delta_1 = 0.15$ s, then the action $\widetilde{a}_{t+0.15}$ represents the action that the BM would take if the vehicle was in the state $s_{t+0.15}$.
The process of obtaining the vector $\boldsymbol{\widetilde{a}}^{\text{TAPM}}_t$ is described by \eqref{eq:a_t-tapm}, where the inputs to $\pi^{\text{TAPM}}_\theta$ are the output of the $\pi^{\text{BM}}_\phi$ (i.e., the action $\widetilde{a}^{\text{BM}}_t$) along with two feature vectors, the image feature vector $\boldsymbol{z^{o}_t}$ and the vehicle velocity vector $\boldsymbol{z^{v}_t}$, both derived from $\pi^{\text{BM}}_\phi$.
\begin{equation}
\begin{aligned}
    \boldsymbol{\widetilde{a}}^{\text{TAPM}}_t
    &= [\widetilde{a}_{t+\delta_1}, \widetilde{a}_{t+\delta_2}, \ldots , \widetilde{a}_{t+\delta_N}] \\ 
    &= \pi^{\text{TAPM}}_\theta(\widetilde{a}_t, \boldsymbol{z^{o}_t}, \boldsymbol{z^{v}_t}).
\end{aligned}
\label{eq:a_t-tapm}
\end{equation}

The final action of the PLM-Net, denoted by $\widetilde{a}^{\text{PLM}}_t$, is obtained through linear interpolation, as detailed in Algorithm~\ref{alg:linear_interpolation_latency}. 
This interpolation combines the outputs of both $\pi^{\text{BM}}_\phi$ and $\pi^{\text{TAPM}}_\theta$ according to the real-time perception latency value $\delta$. In this work, the latency $\delta$ is assumed to be measurable or monitored at inference time (e.g., through system-level latency tracking mechanisms) and is explicitly injected in the controlled simulation environment used for~evaluation.

\begin{algorithm}[H]
\caption{Linear Interpolation for Latency}
\label{alg:linear_interpolation_latency}
\begin{algorithmic}[1]
\REQUIRE $\delta, \delta_{\text{ref}}$ \quad \textit{// Target latency value and list of known latency.}
\REQUIRE $a_{\delta_{\text{ref}}}$ \quad \textit{// Corresponding action values for known latency.}
\ENSURE $a_{\delta}$ \quad \textit{// Interpolated action value for target latency}
\FOR{$j=1$ \TO $N$}
    \IF{$\delta_{\text{ref}}[j] < \delta < \delta_{\text{ref}}[j+1]$}
        \STATE $a_{\delta} \leftarrow a_{\delta_{\text{ref}}}[j] + \frac{\delta - \delta_{\text{ref}}[j]}{\delta_{\text{ref}}[j+1] - \delta_{\text{ref}}[j]} \times (a_{\delta_{\text{ref}}}[j+1] - a_{\delta_{\text{ref}}}[j])$
    \ENDIF
\ENDFOR
\RETURN $a_{\delta}$
\end{algorithmic}
\end{algorithm}
Since $\widetilde{a}^{\text{BM}}_t$ represents the steering action corresponding to the zero-latency case; it is equivalent to $\widetilde{a}_{t+0.0}$. Therefore, the value $0.0$ is prepended to $\boldsymbol{\delta}^{\text{TAPM}}$ and the action $\widetilde{a}^{\text{BM}}_t$ is appended to $\boldsymbol{\widetilde{a}}^{\text{TAPM}}_t$ to construct the reference latency vector $\boldsymbol{\delta}_{ref}$ and its associated steering vector $\boldsymbol{\widetilde{a}}_{\delta_{ref}}$, as shown in \eqref{eq:algo}.
\begin{equation}
\begin{split}
    \boldsymbol{\delta}_{ref} &= [0.0, \boldsymbol{{\delta}}^{\text{TAPM}}] = [0.0, \delta_1, \delta_2, \ldots , \delta_N], \\
    \boldsymbol{\widetilde{a}_{\delta_{ref}}} 
    &= [\widetilde{a}^{\text{BM}}_t, \boldsymbol{\widetilde{a}}^{\text{TAPM}}_t] \\
    &= [\widetilde{a}_{t+0.0}, \widetilde{a}_{t+\delta_1}, \widetilde{a}_{t+\delta_2}, \ldots , \widetilde{a}_{t+\delta_N}].
\end{split}
\label{eq:algo}
\end{equation}

Given a measured latency value $\delta$, the algorithm identifies the two adjacent latency entries in $\boldsymbol{\delta}_{ref}$ that bound $\delta$ and performs linear interpolation between their corresponding steering predictions. If $\delta$ exactly matches one of the predefined latency values, the corresponding steering action is selected directly without interpolation.

At inference time, the Base Model first computes the nominal steering action, the TAPM generates the latency-indexed predictive actions, and the measured latency $\delta$ determines the interpolated final output. This procedure enables smooth mitigation of both constant and time-varying perception latency within the predefined latency range.

\subsection{PLM-Net Models Architecture}
We now detail the internal architecture of the Base Model (BM) and the Timed Action Prediction Model (TAPM).

\begin{figure}[H]
\includegraphics[width=0.85\linewidth]{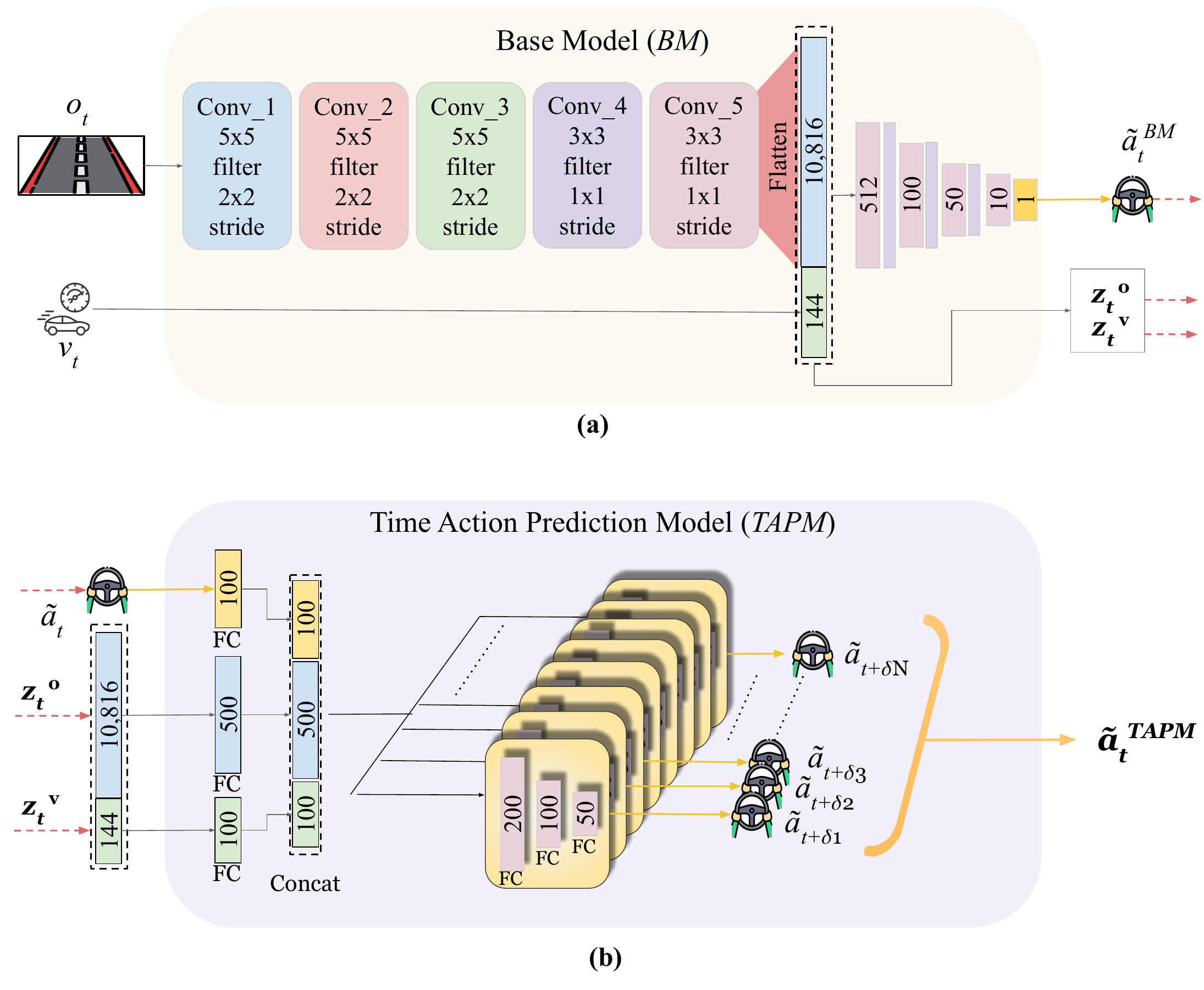}
\caption{
PLM-Net 
architecture.
(\textbf{a}) The BM network design, inspired by the NVIDIA PilotNet structure \cite{bojarski2016end}, processes visual observation $o_t$ and vehicle speed $v_t$ to predict steering angle $\widetilde{a}^{\text{BM}}_t$. It utilizes five convolutional layers and a multi-layer perceptron network with dropout layers, including fully connected layers with neuron counts of 512, 100, 50, and 10.
(\textbf{b}) The TAPM network design, inspired by \textit{ANEC} \cite{khalil2023anec} and \textit{BCIL} \cite{codevilla2018ConditionalImitation}, processes inputs forwarded by the BM ($\widetilde{a}_t$, $\boldsymbol{z^{o}_t}$, $\boldsymbol{z^{v}_t}$) through fully connected layers with 100, 500, and 100 neurons, respectively. These layers are then concatenated and forwarded to sub-models, including fully connected layers with neuron counts of 200, 100, and 50, and dropout layers with a rate of 0.3. Each sub-model will provide a distinct future action for a certain latency value. The outputs of the sub-models form the TAPM output $\boldsymbol{\widetilde{a}}^{\text{TAPM}}_t$, which is a set of predictive action values corresponding to distinct latency values.
In the diagram, arrows indicate the direction of data flow. Blue fully connected layers represent the image-feature pathway ($\boldsymbol{z^{o}_t}$), green layers correspond to the velocity pathway ($\boldsymbol{z^{v}_t}$), and pink layers denote the action-prediction MLP blocks. The different colors used for convolutional layers serve only for visual distinction between stages of the network.
}
\label{fig:PLM-Net-arch}
\end{figure}

The architecture of the PLM-Net models is illustrated in Figure~\ref{fig:PLM-Net-arch}. The network design of the BM is presented in Figure~\ref{fig:PLM-Net-arch}a, and the network design of the TAPM is presented \mbox{in Figure~\ref{fig:PLM-Net-arch}b.}

The BM network design is inspired by the NVIDIA PilotNet structure \cite{bojarski2016end}, with modifications tailored to our requirements. Specifically, we adapt the network to accept visual observations ($o_t$) and vehicle speed ($v_t$) as inputs and to predict steering angle (action $\widetilde{a}^{\text{BM}}_t$) as output, and we add dropout layers to improve generalization and avoid overfitting.
The BM features two primary inputs: the visual observation $o_t$ and the vehicle speed $v_t$. 
The visual observation undergoes processing through five convolutional layers, then a flatten layer to obtain the image feature vector $\boldsymbol{z^{o}_t}$. 
Simultaneously, the vehicle speed input is directed through a fully connected layer that has $144$ neurons, resulting in the formation of the vehicle speed vector $\boldsymbol{z^{v}_t}$.
Subsequently, the image feature vector $\boldsymbol{z^{o}_t}$ and the vehicle speed vector $\boldsymbol{z^{v}_t}$ are concatenated and forwarded to a multi-layer perceptron (MLP) network. 
This MLP configuration consists of four fully connected layers interspersed with three dropout layers. The fully connected layers contain $512$, $100$, $50$, and $10$ neurons, respectively. The dropout layers maintain a dropout rate of $0.3$.
The final output of the BM $\widetilde{a}^{\text{BM}}_t$, the image feature vector $\boldsymbol{z^{o}_t}$, and the vehicle speed vector $\boldsymbol{z^{v}_t}$, are forwarded to the TAPM network.

The TAPM network design is the result of fusing two key ideas. 
Firstly, it incorporates a predictive model, similar to \textit{ANEC} \cite{khalil2023anec}, which was inspired by human drivers' ability to mitigate perception latency.
Secondly, it utilizes the \textit{BCIL} framework \cite{codevilla2018ConditionalImitation}, employing multiple sub-models to provide a range of predictive action values that align with different latency levels, and adding a ‘command’ input, representing the perception latency $\delta$, to influence the final action value. 
The TAPM network inputs are $\widetilde{a}^{\text{BM}}_t$, $\boldsymbol{z^{o}_t}$, and $\boldsymbol{z^{v}_t}$, forwarded from the BM. These three inputs go to 100-neuron, 500-neuron, and 100-neuron fully connected layers, respectively. 
The outputs of these three layers are then concatenated to be forwarded to all sub-models. 
Each sub-model consists of three fully connected layers, with $200$, $100$, and $50$ neurons, interspersed with two dropout layers with a dropout rate of $0.3$. 
Sub-models outputs will result in $\boldsymbol{\widetilde{a}}^{\text{TAPM}}_t$, shown in \eqref{eq:a_t-tapm}.
Although a single neural network could, in principle, learn a continuous mapping between latency values and steering actions, prior work \cite{codevilla2018ConditionalImitation, cultrera2023addressing} has shown that branched architectures improve stability and performance when handling condition-dependent outputs. This motivated our adoption of a BCIL-inspired design for TAPM, where separate sub-models specialize in distinct latency~conditions.

\subsection{PLM-Net Models Training}

\begin{figure}[H]
\includegraphics[width=0.75\linewidth]{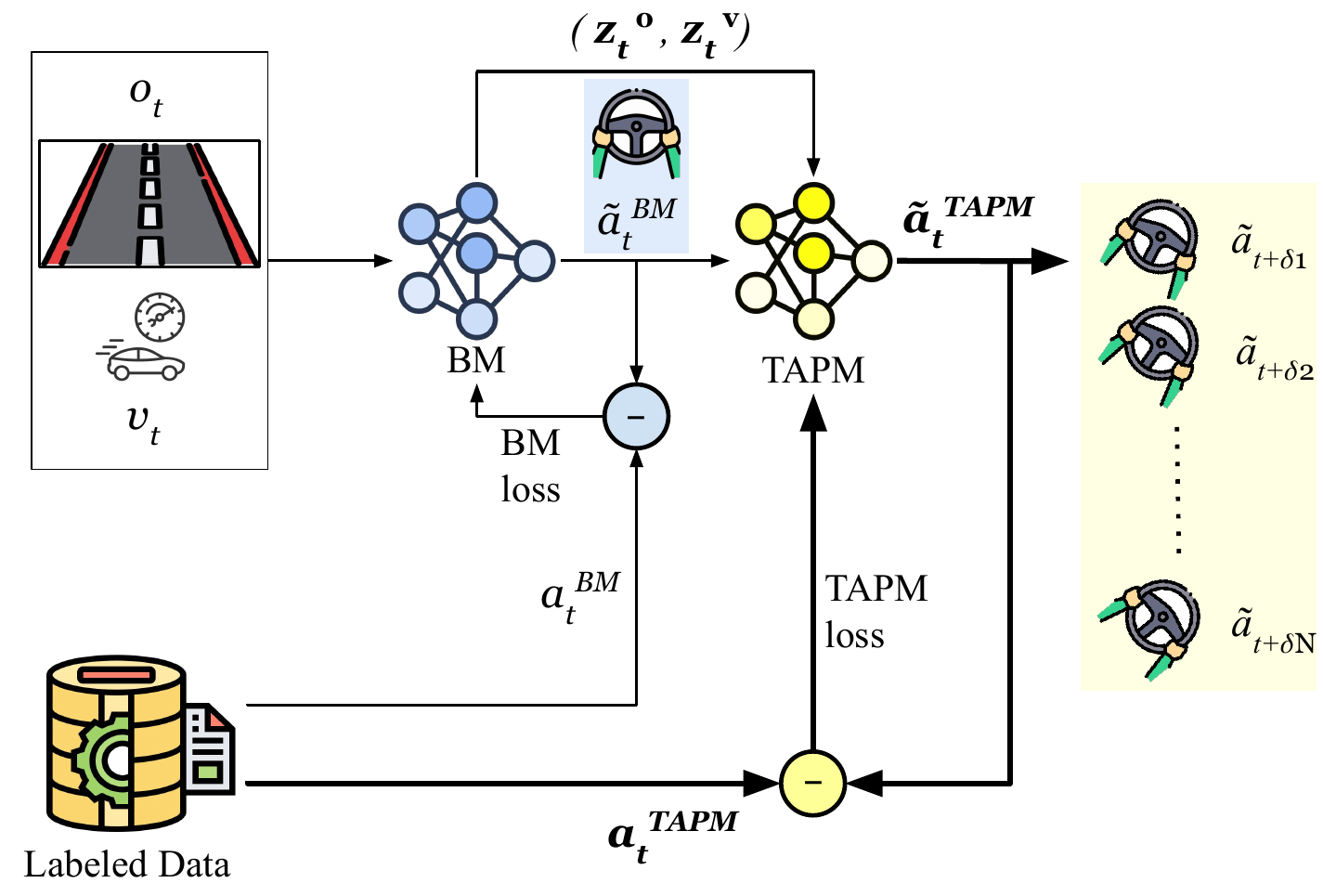}
\caption{
The 
training 
process for PLM-Net models. 
Initially, the BM policy $\pi^{\text{BM}}_\phi$ is trained using supervised learning on dataset $\boldsymbol{D}$, optimizing parameters $\phi$ to minimize prediction error (Equation~\eqref{eq:objective_fun_BM}). 
Subsequently, a new dataset $\boldsymbol{D}^{\text{TAPM}}$ is generated from $\boldsymbol{D}$ to train the TAPM policy $\pi^{\text{TAPM}}_\theta$, optimizing parameters $\theta$ to minimize error between predicted and ground truth future actions (Equation~\eqref{eq:objective_fun_TAPM}). 
During TAPM training, the BM remains non-trainable to ensure TAPM's reliance on its output.
This modular design supports integration into existing control pipelines.
Input data pass through the BM, generating output action $\widetilde{a}^{\text{BM}}_t$ and input vectors $\boldsymbol{z^{o}_t}$ and $\boldsymbol{z^{v}_t}$, which feed into the TAPM. 
The TAPM utilizes these inputs to generate future action values ($\boldsymbol{\widetilde{a}}^{\text{TAPM}}_t$).
In the diagram, the blue network represents the Base Model (BM), while the yellow network represents the TAPM. Circles denote loss computation nodes used during training. Black arrows indicate the flow of data and supervision signals through the training pipeline, and the shaded panel on the right illustrates the set of predicted future actions corresponding to different latency offsets.
}
\label{fig:PLM-Net-train}
\end{figure}

This subsection explains the supervised training procedures for both the BM and the TAPM.
The functionality of a vision-based \textit{LKA} system can be achieved through imitation learning, where we directly map the steering angle (i.e., action $a_t$) to the input $\boldsymbol{i_t} = \{o_t, v_t\}$, where $o_t$ is the visual observation and $v_t$ is the vehicle speed at time $t$. 
The training dataset collected by an expert driver can be defined as $\boldsymbol{D}=\{\boldsymbol{i_t}, a_t\}^M_{t=1}$, where $M$ is the total number of time-steps. 
Figure~\ref{fig:PLM-Net-train} explains the training process of the PLM-Net models.
Learning the BM policy $\pi^{\text{BM}}_\phi$ is a supervised learning problem. The parameters $\phi$ of the policy are optimized by minimizing the prediction error of $a_t$ given the input $i_t$, as shown in \eqref{eq:objective_fun_BM}, where we use the Mean Squared Error (MSE) to calculate the loss per sample. Once optimized, the BM can predict the action $\widetilde{a}^{\text{BM}}_t$ given the input $\boldsymbol{i_t} = \{o_t, v_t\}$ at time $t$, as shown in \eqref{eq:a_t-bm}.
 
\begin{equation}
\underset{\phi}{\arg\min} \frac{1}{M} \sum_{t=1}^{M} (\widetilde{a}^{\text{BM}}_{t} - a_{t})^2
\label{eq:objective_fun_BM}
\end{equation}

\vspace{-12pt}

To learn the TAPM policy $\pi^{\text{TAPM}}_\theta$, we generate a new dataset $\boldsymbol{D}^{\text{TAPM}}$ from the original dataset $\boldsymbol{D}$, mapping the input $\boldsymbol{i_t} = \{o_t, v_t\}$ to $N$ distinct future actions $\boldsymbol{a}^{\text{TAPM}}_t$ corresponding to $N$ distinct latency values $\boldsymbol{\delta}^{\text{TAPM}}$ \eqref{eq:a_t-tapm}, where 
\[
\boldsymbol{D}^{\text{TAPM}}=\{\boldsymbol{i_t}, \boldsymbol{a}^{\text{TAPM}}_t\}^M_{t=1}=\{\boldsymbol{i_t}, [a_{t+\delta_1},\ldots,a_{t+\delta_N}]\}^M_{t=1}
\]

The optimization of the parameters $\theta$ of the TAPM policy $\pi^{\text{TAPM}}_\theta$ is explained in \eqref{eq:objective_fun_TAPM}. We minimize the prediction error of $\boldsymbol{a^{\text{TAPM}}_t}$ given the input $\boldsymbol{i_t}$, where $\boldsymbol{a}^{\text{TAPM}}_t$ indicates the ground truth values of $\boldsymbol{\widetilde{a}}^{\text{TAPM}}_t$. We use MSE to calculate the loss per sample.

\begin{equation}
\underset{\theta}{\arg\min} \frac{1}{M} \sum_{t=1}^{M} \sum_{j=1}^{N} (\widetilde{a}_{t+\delta_j} - a_{t+\delta_j})^2
\label{eq:objective_fun_TAPM}
\end{equation}

To integrate the TAPM with the existing LKA system without modifying it, we set the BM to be non-trainable during the training of the TAPM. This modular separation allows PLM-Net to serve as a plug-in latency mitigation layer, enabling deployment without retraining the base controller.
This design ensures that latency mitigation is achieved without altering the original control policy, preserving the integrity and deployment characteristics of the base lane-keeping controller.

\subsection{Performance Metrics}
\label{subsec:performance-metrics}
Finally, we describe the quantitative metrics used to evaluate latency mitigation performance.
In the context of lateral control for AVs, perception latency can lead to delayed or incorrect actions, particularly affecting the steering angle, which in turn impacts the overall trajectory of the vehicle, as explained in Section \ref{subsec:latency-effect-on-av-lateral-control}. Therefore, to assess PLM-Net's ability to mitigate this latency, two primary evaluation criteria can be employed: steering angle similarity and trajectory similarity. Accurate steering is essential for maintaining lane position and following the desired path, ensuring precise vehicle control even under latency conditions. Meanwhile, trajectory similarity evaluates how closely the vehicle's path adheres to the intended trajectory, providing insight into the broader impact of perception latency on vehicle navigation and the efficacy of PLM-Net in mitigating it.

\section{Simulation-Based Validation Setup}
\label{sec:setup}

This section outlines the experimental setup for validating the Perception Latency Mitigation Network (PLM-Net). 
We begin by describing the simulator used for creating a controlled testing environment. 
Next, we detail the dataset for training and evaluation. 
We then discuss the methods for measuring driving performance, followed by the parameter tuning process. 
After that, an ablation study is presented to highlight key development decisions. 
Finally, we specify the computational environment and machine learning framework, including hardware and software configurations.

\subsection{Simulator}

To simulate and evaluate the effect of perception latency on lateral control, we used the OSCAR simulator \cite{kwon_oscar_2021}, which provides a simulated Ford Fusion vehicle and multiple test tracks. OSCAR is tightly integrated with an ROS (Robotic Operating System) \cite{quigley2009ros} and a Gazebo  multi-robot 3D simulator \cite{koenig2004design-gazebo}, enabling real-time control testing and closed-loop behavior evaluation. These capabilities made it well-suited for our goal of studying perception latency under controlled and repeatable conditions.

While other simulators such as CARLA \cite{Dosovitskiy17} or NVIDIA Drive SIM \cite{nvidia_drive_sim} offer photorealistic rendering and advanced sensor simulation, the focus of our work does not require such features. Our primary goal is to explore the learning and integration of latency-aware control models, and OSCAR provided a practical and efficient platform for this investigation.
In our experiments, perception latency was explicitly injected within the control loop to emulate delay, enabling precise and repeatable evaluation under both constant and time-varying latency profiles.

The simulated environment consisted of a single ego vehicle without surrounding traffic, and the human driver followed the predefined lane centerline as the reference trajectory during data collection.

\subsection{Dataset}

\subsubsection{Training and Test Tracks}
\begin{figure}[H]
\includegraphics[width=0.6\linewidth]{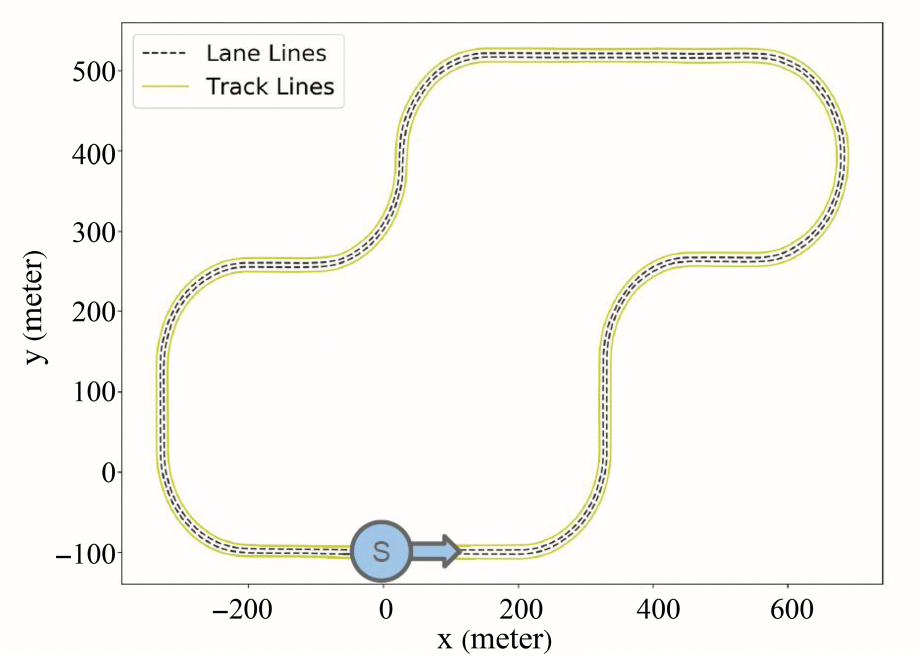}
\caption{Test 
 track used for evaluating the PLM-Net performance. The track features a combination of straight sections and curves to simulate real-world driving conditions. The starting point is marked as $S$ and the arrow indicates the driving direction. This track was chosen to test the vehicle's ability to maintain lane keeping and handle perception latency during both straight and curved segments.}
\label{fig:test-track}
\end{figure}

{The training track used for PLM-Net is identical to the one employed in our prior work} \cite{khalil2023anec}{, ensuring consistency in base controller learning. To evaluate generalization and latency mitigation performance, a separate three-lane test track was designed containing a combination of straight segments and both left and right turns, as shown in Figure}~\ref{fig:test-track}. {This track was not used during training and was selected to expose the controller to multiple curvature conditions so that the interaction between perception latency and different road geometries could be examined during evaluation.}
\vspace{-6pt}

\subsubsection{Data Collection}

Our training dataset, denoted by $\boldsymbol{D}$, was collected by a human driver navigating the training track using an OSCAR simulator. This simulator captures visual observations via a mounted camera on the vehicle, alongside critical control values such as steering angle, throttle position, braking pressure, time, velocity, acceleration, and position. High-quality data was collected using the Logitech G920 dual-motor feedback driving force racing wheel with pedals and a gear shifter, resulting in approximately 115,000 clean training samples. Table \ref{tbl:dataset} provides detailed statistics on the steering and velocity data.

The steering angle for the vehicle is represented on a scale from $-1$ to $1$, where \mbox{$0$ denotes} the center position. According to the right-hand rule, positive steering angles (\mbox{$0$ to $1$}) correspond to rotations to the left, and negative steering angles ($0$ to $-1$) correspond to rotations to the right. 
The steering wheel has a maximum rotation angle of $\pm 450^\circ$. This mapping implies that a steering angle of $1$ corresponds to a $450^\circ$ rotation to the left, while a steering angle of $-1$ corresponds to a $450^\circ$ rotation to the right. Thus, the steering angle range is scaled such that $0$ to $1$ maps to $0^\circ$ to $450^\circ$ and $0$ to $-1$ maps to $0^\circ$ to $-450^\circ$.

During both training and evaluation, the vehicle operated at a constant predefined speed to isolate the effect of perception latency on lateral control. The relationship between speed and latency impact was previously analyzed in \cite{khalil2023anec}; in this work, speed was held constant to focus specifically on latency mitigation behavior.

\begin{table}[H]
\caption{Steering and velocity statistics.}
\label{tbl:dataset}

\begin{tabularx}{\textwidth}{CCC}
\toprule
& \textbf{Steering} & \textbf{Velocity}\\ 
\midrule  
Mean 
 & $-$0.003923 & 20.310327 \\ 
\midrule
Std & 0.176994 & 2.924156 \\ 
\midrule 
Min & $-$1.000000 & 0.018407 \\
\midrule 
25\% & $-$0.069025 & 18.301480 \\
\midrule 
50\% & 0.000000 & 19.883244 \\
\midrule 
75\% & 0.058849 & 22.329700 \\
\midrule 
Max & 0.691892 & 29.936879 \\
\bottomrule
\end{tabularx}
\end{table}

\subsubsection{Data Balancing}

\begin{figure}[H]
\includegraphics[width=0.6\linewidth]{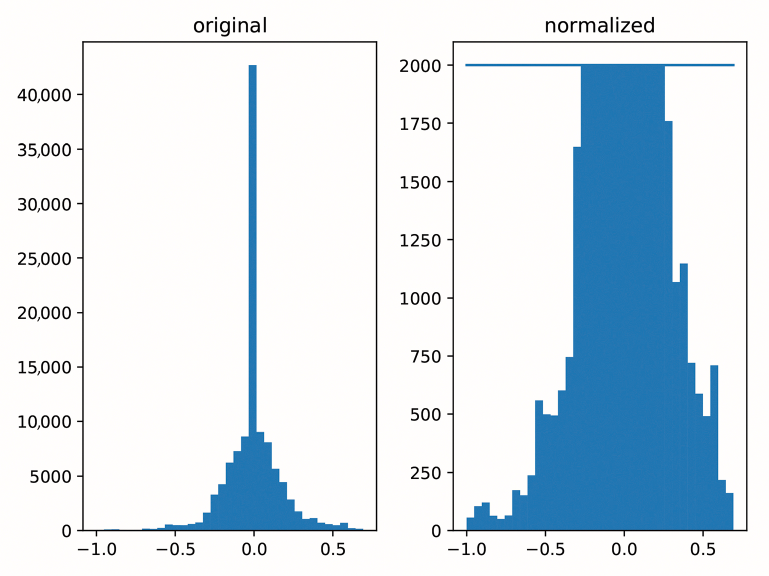}
\caption{
Histogram-based 
 data balancing based on steering angle values. The left image shows the steering histogram before data balancing and the right image shows the histogram after balancing.
}
\label{fig:histogram}
\end{figure}

As shown in Table \ref{tbl:dataset}, approximately $50\%$ of the steering values in our dataset are close to $0.0$, indicating the vehicle traveling on a straight road segment. Training a driving model solely on this dataset would introduce bias. To address this, we conducted histogram-based data balancing to reduce the skew towards zero steering values. Figure~\ref{fig:histogram} illustrates the steering angle histogram before (left) and after (right) the balancing process. Post-balancing, our dataset comprised approximately 67,000 data samples.

\subsubsection{Data Augmentation}
To improve the model's generalization capabilities and augment dataset diversity during training, we implemented data augmentation techniques. Each image presented to the network undergoes a random subset of transformations, including horizontal flipping, where the steering value is negated, and random changes in brightness, while preserving the original steering angle. These augmentation techniques have proven effective in enhancing the robustness of our model during training.

\subsection{Driving Performance Evaluation}
\label{subsec:driving-performance-evaluation}

While Section \ref{subsec:performance-metrics} explained the rationale behind choosing the performance metrics and their importance, this section discusses the adopted methods to measure them in our experiments.

\subsubsection{Steering Angle Similarity}
We analyzed the steering angle values under different conditions: BM driving without latency, BM driving with latency but without TAPM, and BM driving with latency and TAPM (utilizing PLM-Net). This analysis was conducted both qualitatively, through visual inspection, and quantitatively, by calculating metrics such as Mean Absolute Error (MAE), Mean Squared Error (MSE), and Root Mean Squared Error (RMSE).
These metrics quantify pointwise deviation between steering signals, where lower values indicate improved temporal alignment with the latency-free baseline controller.

\subsubsection{Trajectory Similarity}
The performance metrics used to compare the driving trajectories were adopted from~\cite{jekel2019similarity,kim2023opemi}. We measure the similarity between driving trajectories based on lane center positioning. We use partial curve mapping, Frechet distance, area between curves, curve length, and dynamic time warping from \cite{jekel2019similarity}, and the Driving Trajectory Stability Index (DTSI) from \cite{kim2023opemi}.
These trajectory metrics capture complementary aspects of spatial deviation, including geometric similarity, accumulated lateral error, temporal alignment of motion profiles, and overall path stability relative to the latency-free baseline.

All evaluations were conducted in a deterministic closed-loop simulation environment with fixed speed and predefined latency injection profiles. Under identical latency settings, repeated trials yield identical steering outputs and vehicle trajectories. Therefore, a single representative evaluation per latency condition is sufficient to characterize system behavior.

\subsection{Parameter Tuning}
For the TAPM, we used $N=5$ sub-models, meaning the TAPM predicts five future actions for five different latency values in \(\boldsymbol{\delta}^{\text{TAPM}}\).
Specifically, the latency values start with $\delta_1 = 0.15$ s, with increments of \(0.05\) seconds, resulting in $\boldsymbol{\delta}^{\text{TAPM}}~=~[0.15, 0.20, 0.25, 0.30, 0.35]$ seconds. Consequently, the predictive action vector $\boldsymbol{\widetilde{a}}^{\text{TAPM}}_t$ in \eqref{eq:a_t-tapm} becomes
\[
\boldsymbol{\widetilde{a}}^{\text{TAPM}}_t = [\widetilde{a}_{t+0.15}, \widetilde{a}_{t+0.20}, \widetilde{a}_{t+0.25}, \widetilde{a}_{t+0.30}, \widetilde{a}_{t+0.35}]
\]

The selected latency range was designed to cover reported steering actuation delays (0.2 s) \cite{xu2020preview}, as well as visual and network-induced delays, in connected and teleoperated autonomous vehicle systems, which frequently range from 100 to 350 ms and exhibit noticeable degradation around $300$ ms \cite{liu2018hierarchical,kamtam2024network}.

As detailed in \eqref{eq:algo}, the reference latency vector \(\delta_{ref}\) and the corresponding action vector \(\widetilde{a}_{\delta_{ref}}\) are formed by including the BM action for zero latency. Thus, $\boldsymbol{\delta}_{ref}$ and $\boldsymbol{\widetilde{a}}_{\delta_{ref}}$ become
\[
\boldsymbol{\delta}_{ref} = [0.0, 0.15, 0.20, 0.25, 0.30, 0.35].
\]
\[
\boldsymbol{\widetilde{a}}_{\delta_{ref}} = [\widetilde{a}_{t+0.0}, \widetilde{a}_{t+0.15}, \widetilde{a}_{t+0.20}, \widetilde{a}_{t+0.25}, \widetilde{a}_{t+0.30}, \widetilde{a}_{t+0.35}]
\]

These configurations enable the PLM-Net to handle both constant and time-varying perception latency within the range [0--0.35] s.
Latency values outside this predefined range were not evaluated because, beyond approximately $0.30$--$0.35$ s, the baseline model (BM) departed the lane and subsequently left the track, preventing meaningful trajectory-based evaluation. Extending the latency range would require additional latency heads and potentially a redesigned baseline controller.

Both models, the BM and the TAPM, were trained using the Adam optimizer \cite{kingma2014adam} with a batch size of $32$ and a learning rate of $0.001$.
Table \ref{tbl:trainable-param}, shows the number of trainable and non-trainable parameters for the BM and the TAPM. The BM has 6,006,191 parameters, in which all of them are trainable. The TAPM has 12,256,396 total parameters, where 6,250,205 are trainable and 6,006,191 are non-trainable since the BM layers are set to be not trainable when training the TAPM. 
For the performance metric \textit{DTSI}, we used the default parameters recommended in \cite{kim2023opemi}.

\begin{table}[H]
\caption{Trainable and non-trainable parameter counts for the BM and the TAPM, where the BM weights remained frozen during TAPM training.}
\label{tbl:trainable-param}

\begin{tabularx}{\textwidth}{LCCC}
\toprule
 & \textbf{Total} & \textbf{Trainable} & \textbf{Non-Trainable} \\
 & \textbf{Parameters} & \textbf{Parameters} & \textbf{Parameters} \\
\midrule
BM
   & 5,874,566  & 5,874,566  & 0 \\
TAPM & 12,256,396 & 6,250,205  & 6,006,191 \\
\bottomrule
\end{tabularx}
\end{table}

\subsection{Latency Knowledge and Modeling Assumptions}

In this study, perception latency is injected within the closed-loop simulation environment, allowing direct access to the delay value at runtime. The framework therefore assumes availability of a measurable scalar latency estimate rather than knowledge of its physical origin. In practical autonomous driving systems, such estimates can be obtained through timestamp synchronization across sensing, processing, communication, and actuation modules.

The injected delay represents aggregate perception-to-actuation latency rather than separating algorithmic, communication, and actuator components. This abstraction allows the mitigation mechanism to remain agnostic to the physical source of delay and focus on compensating its behavioral impact on steering control.

Linear interpolation between discrete latency-conditioned predictions is adopted because steering evolution remains locally smooth under moderate delay variation within the trained range. The discrete latency heads are ordered with respect to delay magnitude, enabling interpolation to approximate intermediate latency values without modifying the base controller.

\subsection{Ablation Study}

In our exploration of different model architectures for the TAPM network, we experimented with various configurations to optimize performance. Initially, we introduced an additional fully connected layer with $500$ neurons after the concatenation layer preceding the five sub-models. However, this adjustment overly complicated the learning process for the predictive action values. Since this layer was shared among all sub-models, it hindered their individual learning capacities, resulting in subpar outcomes.
Further experimentation involved modifying the number of layers within the sub-models. Reducing the layers to two, with $100$ and $50$ neurons respectively, led to the model's inability to effectively learn predictive actions. Similarly, adding an extra fully connected layer to each sub-model with 300 neurons yielded comparable (if not slightly inferior) results to the existing architecture, thus introducing unnecessary complexity without significant improvement.
Additionally, we explored the integration of Long Short-Term Memory (LSTM) \cite{graves2012long} layers into the sub-models to capture temporal information. However, this approach encountered substantial challenges. Firstly, the model's complexity increased significantly, impeding computational efficiency. Secondly, the nature of capturing temporal information hindered conventional data balancing techniques and random data sampling from the dataset to enhance batch~diversity.

During model training, we found that a batch size of $32$ yielded optimal results compared to smaller sizes such as $16$ and $8$. Furthermore, we experimented with adjusting the learning rate from $0.001$ to $0.0001$. However, the model failed to converge under the lower learning rate, suggesting that the original rate was more conducive to effective~training.

\subsection{Computational Environment and Machine Learning Framework}

Our machine learning framework is built upon TensorFlow and Keras libraries. Specifically, we utilized Keras version $2.2.5$ in conjunction with TensorFlow-GPU version $1.12.0$, leveraging CUDA $9$ and cuDNN $7.1.2$ for GPU acceleration. All computational experiments were conducted on a hardware setup featuring an Intel i7-10700K CPU, 32 GB of RAM, and an NVIDIA GeForce RTX $2080$ Ti with 11 GB of GPU memory. The operating system employed for these experiments was Ubuntu $18.04.6$.

\subsection{Computational Cost Analysis}

To assess the computational overhead introduced by PLM-Net at deployment, we report inference latency and GPU memory usage relative to the BM. Parameter counts and the trainable/non-trainable breakdown are provided in Table~\ref{tbl:trainable-param}. During inference, PLM-Net executes a forward pass through the BM, followed by a forward pass through the TAPM and a lightweight linear interpolation step (Algorithm~\ref{alg:linear_interpolation_latency}).

Inference-time measurements were performed with batch size 1 after discarding the first 50 samples to mitigate warm-up effects. The BM achieved an average inference time of 2.352~ms per frame (p95: 4.532~ms; $N$ = 670) and occupied approximately 650~MB of GPU memory. In comparison, PLM-Net required 5.909~ms per frame on average (p95: 9.880~ms; $N$ = 686) and occupied approximately 778~MB of GPU memory. This corresponds to an additional computational overhead of 3.557~ms per frame and an additional 128~MB of GPU memory (19.7\% increase) relative to the BM.

Given a 30~Hz control frequency (33.3~ms per cycle), the observed inference times confirm that PLM-Net maintains real-time feasibility with a substantial timing margin.
The additional computational overhead is justified by the improved trajectory tracking performance under latency conditions, as demonstrated in Section~\ref{sec:results}.

\section{Results}
\label{sec:results}

Our experimental design aimed to investigate the impact of perception latency on driving and assess the efficacy of our proposed solution, PLM-Net, in mitigating this effect (as detailed in Section \ref{subsec:latency-effect-on-av-lateral-control}). 
To simulate latency, we introduced delays in the input data and applied closed-loop velocity control to maintain a constant vehicle speed (approximately $60$ km/h),
reflecting realistic latency conditions observed in perception--control pipelines where perception and decision-making processes are delayed.
For instance, with a $0.2$ s latency, the available visual observation at time $t$ becomes $o_{t-0.2}$ instead of $o_t$, causing the baseline model (BM) to compute actions based on outdated perception. PLM-Net aims to mitigate this mismatch by predicting an action that better approximates the desired action at time $t$, despite the delayed observation.

We assess the impact of perception latency through a comparison of driving behaviors: BM without latency, BM with latency, and PLM-Net with latency. Successful mitigation by PLM-Net is indicated when its driving performance closely resembles that of the latency-free BM. Our evaluation involves analyzing steering angle similarity and driving trajectory similarity, as detailed in Section \ref{subsec:driving-performance-evaluation}, for both constant and time-variant perception latency.

{To examine how perception latency interacts with road geometry, trajectory similarity metrics are reported not only for the full test track but also for individual track segments, including straight sections, left turns, and right turns. This segment-level evaluation enables analysis of latency-induced deviations under different curvature conditions and provides additional insight into how the mitigation mechanism behaves across distinct driving scenarios.}

\subsection{Constant Perception Latency Mitigation}

\begin{figure}[H]
\includegraphics[width=0.55\linewidth]{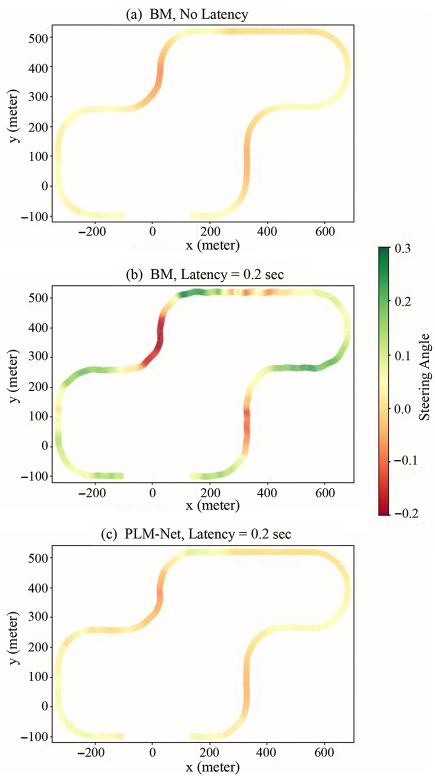}
\caption{
Vehicle 
trajectories 
 colored by steering angle. (\textbf{a}) The trajectory of the BM driving with no latency. (\textbf{b}) The trajectory of the BM driving with $0.2$ s latency. (\textbf{c}) The trajectory of the PLM-Net driving with $0.2$ s latency.
}
\label{fig:plm-delta-200-steering2}
\end{figure}

In evaluating PLM-Net under constant perception latency, we focus on a latency of $0.2$ s, with similar trends observed for other constant latency values, as illustrated in Appendix~\ref{apndx: constant-latency}. 
Figure~\ref{fig:plm-delta-200-steering1} provides a qualitative comparison of steering angles over time between the BM with and without latency and PLM-Net with the same latency. 
In this figure, the blue line represents the BM driving without latency, the green line represents the BM driving with $0.2$ s latency, and the red line represents PLM-Net driving with $0.2$ s latency. 
Additionally, Figure~\ref{fig:plm-delta-200-steering2} provides a visual representation of vehicle trajectories on the test track, colored based on steering angle, to further elucidate the comparative performance. 
Table \ref{tbl:PLM-delta-200-steering} quantifies steering angle errors, demonstrating PLM-Net's superior performance in reducing errors compared to the BM under identical latency conditions. 
Under a constant perception latency of 0.2 s, the performance of the BM degraded substantially.

\begin{figure}[H]
\hspace{-15pt}\includegraphics[width=1\linewidth]{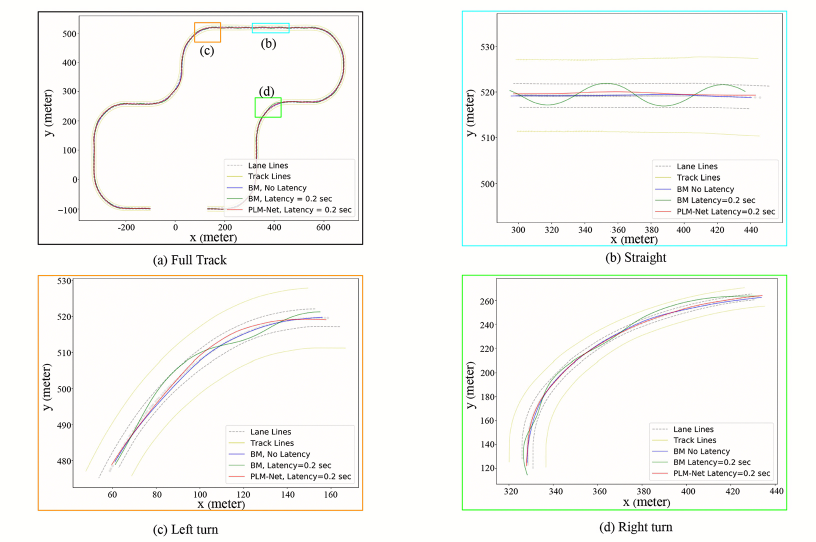}
\caption{
Qualitative 
comparison 
 of trajectories on test track. 
Comparison of driving trajectories on the test track, highlighting differences between the BM driving with no latency (blue line), the BM driving with $0.2$ s latency (green line), and the PLM-Net driving with $0.2$ s latency (red line). (\textbf{a}) The trajectories on the full test track. (\textbf{b}) The trajectories on a straight road segment. (\textbf{c}) The trajectories on a left turn. (\textbf{d}) The trajectories on a right turn.
}
\label{fig:plm-delta-200-traj}
\end{figure}

Furthermore, Figure~\ref{fig:plm-delta-200-traj} presents trajectory comparisons qualitatively, while Table~\ref{tbl:PLM-delta-200-traj} presents trajectory comparisons quantitatively, showing PLM-Net's ability to maintain accurate driving trajectories despite latency-induced challenges. 
Each color-coded trajectory corresponds to a different driving condition: blue for the BM driving without latency, green for the BM driving with $0.2$ s latency, and red for PLM-Net driving with $0.2$ s latency.
Examining the deviation from the lane center on the full track, the Partial Curve Mapping metric for the BM with $0.2$ s latency increased by $238.7\%$, while PLM-Net maintained a much smaller increase of only $11.1\%$. Similarly, improvements in Frechet distance, area between curves, curve length, and DTSI demonstrate that PLM-Net significantly reduces the trajectory deviation caused by latency. 
The additional segments of Figure~\ref{fig:plm-delta-200-traj} and Table \ref{tbl:PLM-delta-200-traj}, specifically parts (b), (c), and (d), which depict the trajectories on a straight road, right turn, and left turn, respectively, also demonstrate that PLM-Net effectively mitigates latency, similar to the results observed on the full track.

\begin{table}[H]
\caption{Steering 
 angle 
 similarity. Quantitative comparison of steering angle between the BM driving without latency, the BM driving with latency of $0.2$ s, and PLM-Net driving with the same latency. Downward arrows indicate that lower values correspond to better performance.
 }
\label{tbl:PLM-delta-200-steering}
\begin{tabularx}{\textwidth}{lCCC}
\toprule
\multicolumn{1}{l}{\textbf{BM, 
 No Latency vs.}} 
  & \textit{\textbf{MAE}} \(\downarrow\) 
  & \textit{\textbf{MSE}} \(\downarrow\) 
  & \textit{\textbf{RMSE}} \(\downarrow\) \\
\midrule
BM, Latency = $0.2$ s      & 0.1915          & 0.0716          & 0.2676 \\
PLM-Net, Latency = $0.2$ s & \textbf{0.0726
} & \textbf{0.0125} & \textbf{0.1119} \\
\bottomrule
\end{tabularx}
\end{table}
\vspace{-12pt}

\begin{figure}[H]
\includegraphics[width=0.55\linewidth]{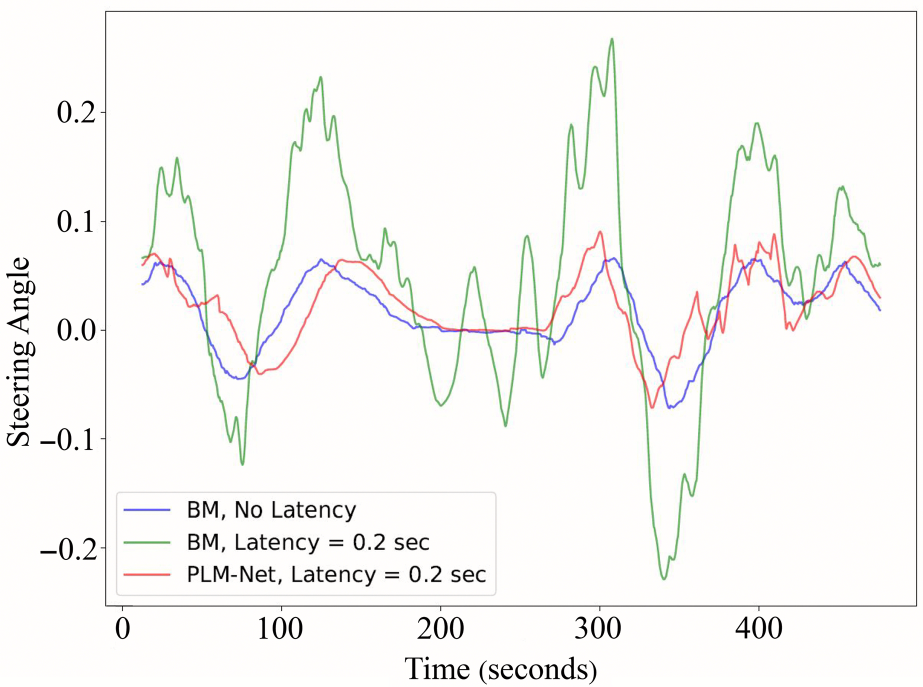}
\caption{
Qualitative 
comparison 
 of steering angle over time. 
The blue line represents the BM driving on the test track with no latency.
The green line represents the BM driving on the test track with $0.2$ s latency.
The red line represents the PLM-Net driving on the test track with the same latency value.
}
\label{fig:plm-delta-200-steering1}
\end{figure}
\vspace{-12pt}

\begin{table}[H]
    \caption{
    Quantitative 
comparison 
 of driving trajectory similarity, between the BM driving without latency, the BM driving with latency of $0.2$ s, and PLM-Net driving with the same latency.
 All driving models' trajectories are compared with the lane center on full track, straight road, left turn, and right turn. Downward arrows indicate that lower values correspond to better performance.
 }
    \label{tbl:PLM-delta-200-traj}

\footnotesize
    \begin{adjustwidth}{-\extralength}{0cm}
    \begin{tabularx}{\fulllength}{LL Cccccc}
    
    \toprule 
    \multirow{3}{*}{\textbf{Figure~\ref{fig:plm-delta-200-traj} Ref.}} & \multirow{3}{*}{\textbf{Driving Model}} 
                                    & \multicolumn{1}{c}{\textbf{Partial}}  
                                    & \multicolumn{1}{c}{\textbf{Frechet}}    
                                    & \multicolumn{1}{c}{\textbf{Area}}  & \multicolumn{1}{c}{\textbf{Curve}}          
                                    & \multicolumn{1}{c}{\textbf{Dynamic}} & \multicolumn{1}{c}{}
                                    \\ 
                                    & & \multicolumn{1}{c}{\textbf{Curve} \boldmath{$\downarrow$}} & \multicolumn{1}{c}{\textbf{Distance} \boldmath{$\downarrow$}} 
                                    & \multicolumn{1}{c}{\textbf{Between} \boldmath{$\downarrow$}}  & \multicolumn{1}{c}{\textbf{Length} \boldmath{$\downarrow$}}          
                                    & \multicolumn{1}{c}{\textbf{Time} \boldmath{$\downarrow$}} & \multicolumn{1}{c}{\textoverline{\textbf{DTSI}} \boldmath{$\downarrow$}}
                                    \\
                                    & & \multicolumn{1}{c}{\textbf{Mapping}} & \multicolumn{1}{c}{} 
                                    & \multicolumn{1}{c}{\textbf{Curves}}  & \multicolumn{1}{c}{}          
                                    & \multicolumn{1}{c}{\textbf{Warping}} & \multicolumn{1}{c}{}
                                    \\ \hline 
                                    
    \rowcolor{gray!30}
    \textbf{(a) 
 Full Track}                  & \multicolumn{1}{l}{\textbf{BM, No Latency (ref)}}                  
                                    & \multicolumn{1}{c}{\textbf{1.19}}     & \multicolumn{1}{c}{\textbf{2.176}}    
                                    & \multicolumn{1}{c}{\textbf{4658.15}}     & \multicolumn{1}{c}{\textbf{0.349}}      
                                    & \multicolumn{1}{c}{\textbf{1253.21}}     & \multicolumn{1}{c}{\textbf{0.034}} \\ \hline
                    
    (a) Full Track                  & \multicolumn{1}{l}{BM, Latency = $0.2$ s}         
                                    & \multicolumn{1}{c}{4.03}    & \multicolumn{1}{c}{5.469}    
                                    & \multicolumn{1}{c}{\textbf{23,045.1}}    & \multicolumn{1}{c}{0.767}      
                                    & \multicolumn{1}{c}{2594.39}    & \multicolumn{1}{c}{0.072}\\ \midrule
                    
    (a) Full Track                  & \multicolumn{1}{l}{PLM-Net, Latency = $0.2$ s}    
                                    & \multicolumn{1}{c}{\textbf{1.322}}    & \multicolumn{1}{c}{\textbf{3.418}}    
                                    & \multicolumn{1}{c}{39,523.2}     & \multicolumn{1}{c}{\textbf{0.374}}      
                                    & \multicolumn{1}{c}{\textbf{2299.84}}   & \multicolumn{1}{c}{\textbf{0.036}}\\ \hline 

    \rowcolor{cyan!10}
    \textbf{(b) Straight}                    & \multicolumn{1}{l}{\textbf{BM, No Latency (ref)}}                  
                                    & \multicolumn{1}{c}{\textbf{24.475}}     & \multicolumn{1}{c}{\textbf{5.095}}    
                                    & \multicolumn{1}{c}{\textbf{13.556}}     & \multicolumn{1}{c}{\textbf{0.077}}      
                                    & \multicolumn{1}{c}{\textbf{76.376}}     & \multicolumn{1}{c}{\textbf{0.012}} \\ \hline
                    
    (b) Straight                    & \multicolumn{1}{l}{BM, Latency = $0.2$ s }         
                                    & \multicolumn{1}{c}{351.807}    & \multicolumn{1}{c}{9.048}    
                                    & \multicolumn{1}{c}{227.119}    & \multicolumn{1}{c}{0.209}      
                                    & \multicolumn{1}{c}{338.861}    & \multicolumn{1}{c}{0.095}\\ \midrule
                    
    (b) Straight                    & \multicolumn{1}{l}{PLM-Net, Latency = $0.2$ s}    
                                    & \multicolumn{1}{c}{\textbf{137.353}}    & \multicolumn{1}{c}{\textbf{6.117}}    
                                    & \multicolumn{1}{c}{\textbf{66.545}}    & \multicolumn{1}{c}{\textbf{0.093}}      
                                    & \multicolumn{1}{c}{\textbf{136.861}}    & \multicolumn{1}{c}{\textbf{0.016}}\\ \hline 

    \rowcolor{orange!20}
    \textbf{(c) Left turn}                   & \multicolumn{1}{l}{\textbf{BM, No Latency (ref)}}     
                                    & \multicolumn{1}{c}{\textbf{0.895}}    & \multicolumn{1}{c}{\textbf{3.966}}    
                                    & \multicolumn{1}{c}{\textbf{41.402}}     & \multicolumn{1}{c}{\textbf{0.178}}      
                                    & \multicolumn{1}{c}{\textbf{67.892}}    & \multicolumn{1}{c}{\textbf{0.025}}\\ \hline
                    
    (c) Left turn                   & \multicolumn{1}{l}{BM, Latency = $0.2$ s }         
                                    & \multicolumn{1}{c}{1.3}    & \multicolumn{1}{c}{5.405}    
                                    & \multicolumn{1}{c}{141.041}     & \multicolumn{1}{c}{0.234}      
                                    & \multicolumn{1}{c}{156.02}    & \multicolumn{1}{c}{0.048}\\ \midrule
                    
    (c) Left turn                   & \multicolumn{1}{l}{PLM-Net, Latency = $0.2$ s}    
                                    & \multicolumn{1}{c}{\textbf{0.674}}    & \multicolumn{1}{c}{\textbf{1.5}}    
                                    & \multicolumn{1}{c}{\textbf{71.816}}     & \multicolumn{1}{c}{\textbf{0.05}}      
                                    & \multicolumn{1}{c}{\textbf{81.885}}    & \multicolumn{1}{c}{\textbf{0.01}}\\ \hline 

    \rowcolor{green!20}
    \textbf{(d) Right turn}                  & \multicolumn{1}{l}{\textbf{BM, No Latency (ref)}}                  
                                    & \multicolumn{1}{c}{\textbf{0.293}}    & \multicolumn{1}{c}{\textbf{1.546}}    
                                    & \multicolumn{1}{c}{\textbf{90.228}}     & \multicolumn{1}{c}{\textbf{0.048}}      
                                    & \multicolumn{1}{c}{\textbf{79.036}}     & \multicolumn{1}{c}{\textbf{0.037}}\\ \hline
                    
    (d) Right turn                  & \multicolumn{1}{l}{BM, Latency = $0.2$ s }         
                                    & \multicolumn{1}{c}{0.927}    & \multicolumn{1}{c}{5.357}    
                                    & \multicolumn{1}{c}{384.746}     & \multicolumn{1}{c}{0.124}      
                                    & \multicolumn{1}{c}{232.834}   & \multicolumn{1}{c}{\textbf{0.055}}\\ \midrule
                    
    (d) Right turn                  & \multicolumn{1}{l}{PLM-Net, Latency = $0.2$ s}    
                                    & \multicolumn{1}{c}{\textbf{0.745}}    & \multicolumn{1}{c}{\textbf{2.863}}    
                                    & \multicolumn{1}{c}{\textbf{45.779}}     & \multicolumn{1}{c}{\textbf{0.063}}
                                    & \multicolumn{1}{c}{\textbf{69.018}}   & \multicolumn{1}{c}{0.078}\\ \bottomrule 
                    
    \end{tabularx}
    \end{adjustwidth}
\end{table}

The Mean Absolute Error (MAE) between the BM without latency and the BM with latency  is $0.1915$, indicating a substantial degradation in steering angle accuracy. However, when using PLM-Net, the performance decline was mitigated, with the MAE reduced to $0.0726$. 
This corresponds to a $62.1\%$ reduction in MAE relative to the BM under the same latency condition.
Additionally, the Mean Squared Error (MSE) and Root Mean Squared Error (RMSE) were similarly improved with PLM-Net, showing reductions of $82.5\%$ and $58.2\%$, respectively.

\subsection{Time-Variant Perception Latency Mitigation}

\begin{figure}[H]
\includegraphics[width=0.5\linewidth]{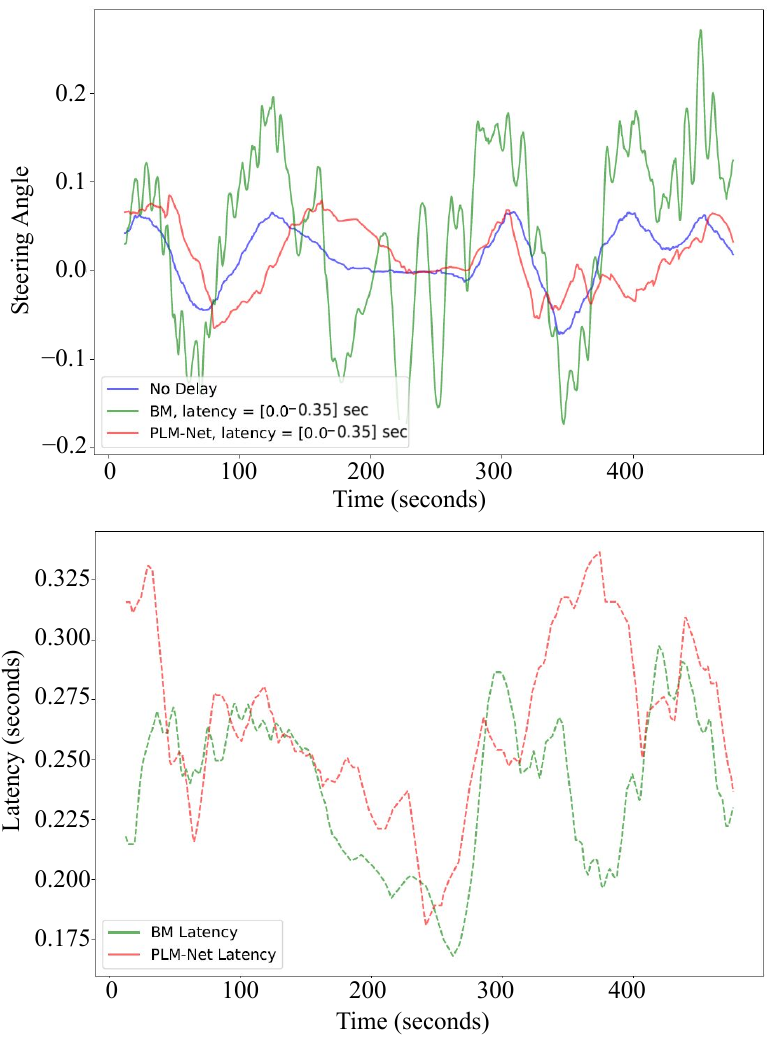}
\caption{
Qualitative 
 comparison 
 of 
 steering angle over time is depicted in the upper image. 
The blue line represents the BM driving without latency on the test track. 
Meanwhile, the green line illustrates the BM's performance under time-variant latency, ranging from $0.0$ to $0.35$ s. 
The red line showcases the PLM-Net's driving behavior under the same time-variant latency conditions. 
In the lower image, latency values against time are presented. The green line corresponds to the latency experienced by the BM, while the red line represents the latency encountered by the PLM-Net.
}
\label{fig:plm-delta-rand-steering1}
\end{figure}

\begin{figure}[H]
\includegraphics[width=0.5\linewidth]{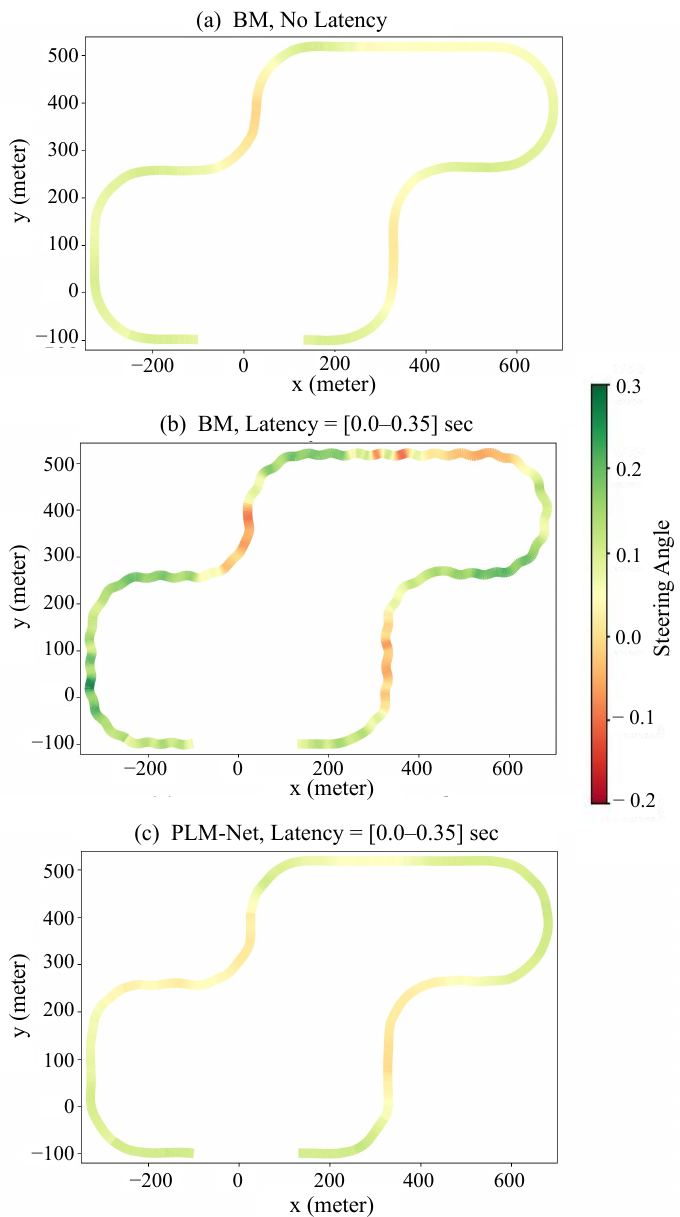}
\caption{
Colored 
trajectories 
 based on steering angle. (\textbf{a}) The trajectory of the BM driving with no latency. (\textbf{b}) The trajectory of the BM driving under time-variant latency [0.0--0.35] s. (\textbf{c}) The trajectory of the PLM-Net driving under time-variant latency [0.0--0.35] s.
}
\label{fig:plm-delta-rand-steering2}
\end{figure}

For time-variant perception latency, we evaluate PLM-Net against varying latency levels ([0.0--0.35] s). 
Similar to the constant latency scenario, the upper image in Figure~\ref{fig:plm-delta-rand-steering1} illustrates qualitative steering angle comparisons over time, with the blue line representing the BM driving without latency, the green line representing the BM driving with time-variant latency, and the red line representing PLM-Net driving with time-variant latency. The lower image illustrates the varying latency values experienced by both models over time. 
Additionally, Figure~\ref{fig:plm-delta-rand-steering2} provides a visual representation of vehicle trajectories on the test track, colored based on steering angle values. 
Table \ref{tbl:PLM-delta-rand-steering} quantifies steering angle errors, demonstrating PLM-Net's effectiveness in reducing errors compared to the BM under time-variant latency conditions. 
Under a time-variant perception latency of \mbox{[0.0--0.35] s,} the performance of the BM degraded substantially.
The Mean Absolute Error (MAE) between the BM without latency and the BM with latency  is $0.3336$, indicating a substantial degradation in steering angle similarity. However, when using PLM-Net, the performance decline was mitigated, with the MAE reduced to $0.0710$. This represents a $78.7\%$ improvement in MAE compared to the BM under the same latency condition. Additionally, the Mean Squared Error (MSE) and Root Mean Squared Error (RMSE) were similarly improved with PLM-Net, showing reductions of $94.2\%$ and $76.0\%$, respectively.

\vspace{-9pt}
\begin{table}[H]
\caption{Steering angle similarity. Quantitative comparison of steering angle between BM driving with no latency and with time-variant latency [0.0--0.35] s, and PLM-Net driving with the same time-variant latency. Downward arrows indicate that lower values correspond to better performance.}
\label{tbl:PLM-delta-rand-steering}
\begin{tabularx}{\textwidth}{Lccc}
\toprule
\multicolumn{1}{l}{\textit{\textbf{BM, No Latency vs.}}} 
  & \textit{\textbf{MAE}} \(\downarrow\) 
  & \textit{\textbf{MSE}} \(\downarrow\) 
  & \textit{\textbf{RMSE}} \(\downarrow\) \\
\midrule  

\textit{BM, Latency = [0.0--0.35] s} & 0.3336 & 0.1871 & 0.4326 \\ 
\midrule
\textit{PLM-Net, Latency = [0.0--0.35] s} & \textbf{0.0710} & \textbf{0.0108} & \textbf{0.1037} \\ 
\bottomrule
\end{tabularx}
\end{table}

\begin{figure}[H]
\hspace{-18pt}
\includegraphics[width=0.9\linewidth]{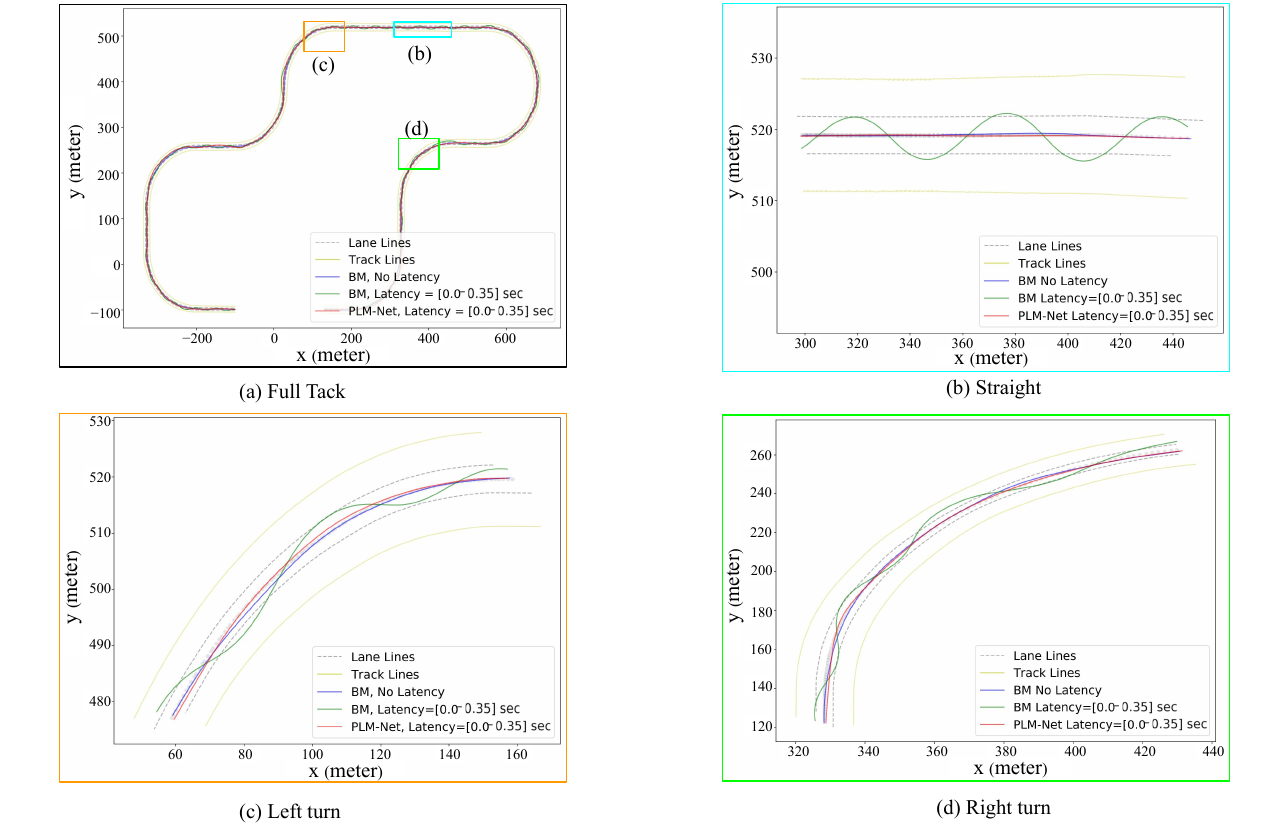}
\caption{
Trajectory 
comparison 
between 
 the BM driving with no latency (blue line), the BM driving with time-variant latency [0.0--0.35] s  (green line), and the PLM-Net driving with time-variant latency [0.0--0.35] s (red line). (\textbf{a}) The trajectories on the full test track. (\textbf{b}) The trajectories on a straight road segment. (\textbf{c}) The trajectories on a left turn. (\textbf{d}) The trajectories on a right turn.
}
\label{fig:plm-delta-rand-traj}
\end{figure}

Figure~\ref{fig:plm-delta-rand-traj} presents trajectory comparisons qualitatively, while Table \ref{tbl:PLM-delta-rand-traj} presents trajectory comparisons quantitatively, confirming PLM-Net's successful mitigation of time-variant perception latency.
Each color-coded trajectory corresponds to a different driving condition: blue for the BM driving without latency, green for the BM driving with time-variant latency, and red for PLM-Net driving with time-variant latency.
The Partial Curve Mapping metric revealed a $396.6\%$ increase in deviation from the lane center for the BM, with a [0.0--0.35] s time-variant latency on the full track, while PLM-Net showed a more modest increase of $114.0\%$. Similarly, the Frechet distance for the BM increased by $254.9\%$, compared to a $66.2\%$ increase for PLM-Net. This pattern of reduced deviation is also reflected in the improvements in area between curves, curve length, and DTSI, indicating that PLM-Net significantly mitigates the trajectory deviations caused by latency.
Parts (b), (c), and (d) of Figure~\ref{fig:plm-delta-rand-traj} and Table \ref{tbl:PLM-delta-rand-traj}, which illustrate the trajectories during a straight segment, a right turn, and a left turn, respectively, exhibit results consistent with the full track, confirming that PLM-Net successfully mitigates latency.

\vspace{-9pt}
\begin{table}[H]
    \caption{
    Comparative 
analysis 
 of driving trajectory similarity. The BM driving without latency, the BM driving under time-variant latency ranging from $0.0$ to $0.35$ s, and PLM-Net driving within the same time-variant latency range. 
    Trajectories of all driving models are evaluated against the lane center across various scenarios, including full track, straight road, left turn, and right turn.
     Downward arrows indicate that lower values correspond to better performance.
    }
    \label{tbl:PLM-delta-rand-traj}
    \begin{adjustwidth}{-\extralength}{0cm}
    \footnotesize
    \begin{tabularx}{\fulllength}{lLCCCCCC}
    
    \toprule 

    \multirow{3}{*}{\textbf{Figure~\ref{fig:plm-delta-rand-traj} ref.}} & \multirow{3}{*}{\textbf{Driving} \textbf{Model}} 
                                    & \multicolumn{1}{c}{\textbf{Partial}}  
                                    & \multicolumn{1}{c}{\textbf{Frechet}}    
                                    & \multicolumn{1}{c}{\textbf{Area}}  & \multicolumn{1}{c}{\textbf{Curve}}          
                                    & \multicolumn{1}{c}{\textbf{Dynamic}} & \multicolumn{1}{c}{}
                                    \\ 
                                    & & \multicolumn{1}{c}{\textbf{Curve} \boldmath{$\downarrow$}} & \multicolumn{1}{c}{\textbf{Distance} \boldmath{$\downarrow$}} 
                                    & \multicolumn{1}{c}{\textbf{Between} \boldmath{$\downarrow$}}  & \multicolumn{1}{c}{\textbf{Length} \boldmath{$\downarrow$}}          
                                    & \multicolumn{1}{c}{\textbf{Time} \boldmath{$\downarrow$}} & \multicolumn{1}{c}{\textoverline{\textbf{DTSI}} \boldmath{$\downarrow$}}
                                    \\
                                    & & \multicolumn{1}{c}{\textbf{Mapping}} & \multicolumn{1}{c}{} 
                                    & \multicolumn{1}{c}{\textbf{Curves}}  & \multicolumn{1}{c}{}          
                                    & \multicolumn{1}{c}{\textbf{Warping}} & \multicolumn{1}{c}{}
                                    \\ \hline
                                    
    \rowcolor{gray!30}
    \textbf{(a) Full Track}                  & \multicolumn{1}{l}{\textbf{BM, No Latency (ref)}}                  
                                    & \multicolumn{1}{c}{\textbf{1.19}}     & \multicolumn{1}{c}{\textbf{2.176}}    
                                    & \multicolumn{1}{c}{\textbf{4658.15}}     & \multicolumn{1}{c}{\textbf{0.349}}      
                                    & \multicolumn{1}{c}{\textbf{1253.21}}     & \multicolumn{1}{c}{\textbf{0.034}} \\ \hline
                    
    (a) Full Track                  & \multicolumn{1}{l}{BM, Latency = 0.0--0.35 s}         
                                    & \multicolumn{1}{c}{5.91}     & \multicolumn{1}{c}{7.723}    
                                    & \multicolumn{1}{c}{\textbf{26,498.2}}     & \multicolumn{1}{c}{1.245}      
                                    & \multicolumn{1}{c}{4268.24}     & \multicolumn{1}{c}{0.238}\\ \midrule
                    
    (a) Full Track                  & \multicolumn{1}{l}{PLM-Net, Latency = 0.0--0.35 s}    
                                    & \multicolumn{1}{c}{\textbf{2.547}}     & \multicolumn{1}{c}{\textbf{3.616}}    
                                    & \multicolumn{1}{c}{29,126.6}     & \multicolumn{1}{c}{\textbf{0.694}}      
                                    & \multicolumn{1}{c}{\textbf{2653.26}}     & \multicolumn{1}{c}{\textbf{0.028}}\\ \hline

    \rowcolor{cyan!10}
    \textbf{(b) Straight}                    & \multicolumn{1}{l}{\textbf{BM, No Latency (ref)}}                  
                                    & \multicolumn{1}{c}{\textbf{24.475}}     & \multicolumn{1}{c}{\textbf{5.095}}    
                                    & \multicolumn{1}{c}{\textbf{13.556}}     & \multicolumn{1}{c}{\textbf{0.077}}      
                                    & \multicolumn{1}{c}{\textbf{76.376}}     & \multicolumn{1}{c}{\textbf{0.012}} \\ \hline
                    
    (b) Straight                    & \multicolumn{1}{l}{BM, Latency = 0.0--0.35 s }         
                                    & \multicolumn{1}{c}{465.004}     & \multicolumn{1}{c}{4.026}    
                                    & \multicolumn{1}{c}{293.106}     & \multicolumn{1}{c}{0.08}      
                                    & \multicolumn{1}{c}{409.007}     & \multicolumn{1}{c}{0.109}\\ \midrule
                    
    (b) Straight                    & \multicolumn{1}{l}{PLM-Net, Latency = 0.0--0.35 s}    
                                    & \multicolumn{1}{c}{\textbf{21.428}}     & \multicolumn{1}{c}{\textbf{1.319}}    
                                    & \multicolumn{1}{c}{\textbf{12.097}}     & \multicolumn{1}{c}{\textbf{0.032}}      
                                    & \multicolumn{1}{c}{\textbf{49.84}}     & \multicolumn{1}{c}{\textbf{0.014}}\\ 
                                    \bottomrule

    \rowcolor{orange!20}
    \textbf{(c) Left turn}                   & \multicolumn{1}{l}{\textbf{BM, No Latency (ref)}}                  
                                    & \multicolumn{1}{c}{\textbf{0.895}}     & \multicolumn{1}{c}{\textbf{3.966}}    
                                    & \multicolumn{1}{c}{\textbf{41.402}}     & \multicolumn{1}{c}{\textbf{0.178}}      
                                    & \multicolumn{1}{c}{\textbf{67.892}}     & \multicolumn{1}{c}{\textbf{0.025}}\\ \hline
                    
    (c) Left turn                   & \multicolumn{1}{l}{BM, Latency = 0.0--0.35 s }         
                                    & \multicolumn{1}{c}{1.697}     & \multicolumn{1}{c}{4.449}    
                                    & \multicolumn{1}{c}{209.537}     & \multicolumn{1}{c}{0.284}      
                                    & \multicolumn{1}{c}{198.209}     & \multicolumn{1}{c}{0.107}\\ \midrule
                    
    (c) Left turn                   & \multicolumn{1}{l}{PLM-Net, Latency = 0.0--0.35 s}    
                                    & \multicolumn{1}{c}{\textbf{0.4}}     & \multicolumn{1}{c}{\textbf{1.527}}    
                                    & \multicolumn{1}{c}{\textbf{45.844}}     & \multicolumn{1}{c}{\textbf{0.087}}      
                                    & \multicolumn{1}{c}{\textbf{59.073}}     & \multicolumn{1}{c}{\textbf{0.017}}\\ \hline 

    \rowcolor{green!20}
    \textbf{(d) Right turn}                  & \multicolumn{1}{l}{\textbf{BM, No Latency (ref)}}                  
                                    & \multicolumn{1}{c}{\textbf{0.293}}     & \multicolumn{1}{c}{\textbf{1.546}}    
                                    & \multicolumn{1}{c}{\textbf{90.228}}     & \multicolumn{1}{c}{\textbf{0.048}}      
                                    & \multicolumn{1}{c}{\textbf{79.036}}     & \multicolumn{1}{c}{\textbf{0.037}}\\ \hline
                    
    (d) Right turn                  & \multicolumn{1}{l}{BM, Latency = 0.0--0.35 s }         
                                    & \multicolumn{1}{c}{0.744}     & \multicolumn{1}{c}{4.572}    
                                    & \multicolumn{1}{c}{420.415}     & \multicolumn{1}{c}{0.11}      
                                    & \multicolumn{1}{c}{228.825}     & \multicolumn{1}{c}{\textbf{0.097}}\\ \midrule
                    
    (d) Right turn                  & \multicolumn{1}{l}{PLM-Net, Latency = 0.0--0.35 s}    
                                    & \multicolumn{1}{c}{\textbf{0.327}}     & \multicolumn{1}{c}{\textbf{1.692}}    
                                    & \multicolumn{1}{c}{\textbf{89.126}}     & \multicolumn{1}{c}{\textbf{0.046}}      
                                    & \multicolumn{1}{c}{\textbf{92.033}}     & \multicolumn{1}{c}{0.056}\\ \bottomrule 
                    
    \end{tabularx}
    \end{adjustwidth}
\end{table}

Overall, the experimental findings underscore PLM-Net's robustness in mitigating both constant and time-variant perception latency, 
indicating its potential for improving robustness against perception latency in vision-based autonomous driving systems.

\section{Conclusions}
\label{sec:conclusion}

This paper introduced PLM-Net, a learning-based framework designed to mitigate the effect of perception latency in vision-based imitation-learning lateral control systems. By integrating a Timed Action Prediction Model (TAPM) with an existing Base Model (BM), PLM-Net anticipates latency-induced mismatch between perception and control without modifying the original controller architecture.

The TAPM predicts a discrete set of future steering actions corresponding to predefined latency values, and linear interpolation is used to generate the final control output according to the real-time latency. This design enables mitigation of both constant and time-varying perception latency within the modeled range.

Experimental results in a deterministic closed-loop simulation environment demonstrated that PLM-Net substantially reduces latency-induced steering and trajectory errors. Under a constant $0.2$ s latency, PLM-Net achieved a $62.1\%$ reduction in MAE compared to the baseline model. Under time-variant latency within [0.0--0.35] s, the MAE reduction reached $78.7\%$. Trajectory-based metrics further confirmed improved lane-following performance across straight and turning segments.

\textbf{Limitations:} This study was conducted in a deterministic closed-loop simulation environment with a fixed vehicle speed to isolate the effect of perception latency on lateral control behavior. 
{While this controlled setting enables systematic evaluation of latency mitigation behavior, broader validation across multiple drivers, stochastic latency realizations, and more diverse road environments remains an important direction for future work.}
The training dataset was collected from a single driver in a controlled simulator setting, consistent with common imitation-learning frameworks, and was evaluated on a separate test track to assess generalization under the studied conditions. The modeled latency range was bounded within [0.0--0.35] s, beyond which the baseline controller departed the track boundaries, preventing meaningful trajectory-based comparison. Additionally, the proposed framework was evaluated for lateral control only under aggregate perception-to-actuation delay modeling. These constraints define the scope of validation for the present study and motivate future investigation under broader operational, dynamic, and multi-sensor conditions. 
The present work focuses on architectural latency mitigation under bounded delay assumptions and does not claim replacement or superiority over model-based delay compensation methods; rather, it provides a complementary learning-based solution designed for vision-based imitation-learning control pipelines where explicit vehicle modeling or controller redesign may not be feasible.

Future work will focus on extending the modeled latency range, evaluating the framework under varying vehicle speeds and more complex traffic scenarios, integrating the approach into higher-fidelity simulation and real-world platforms, and exploring hybrid strategies that combine learning-based latency mitigation with classical delay-aware control methods to enhance robustness and theoretical guarantees.

\vspace{6pt} 
\authorcontributions{Conceptualization, J.K. and A.K.; methodology, A.K.; software, A.K.; validation, A.K.; formal analysis, A.K.; resources, J.K.; data curation, A.K.; writing---original draft preparation, A.K.; writing---review and editing, J.K.; visualization, A.K.; supervision, J.K.; project administration, J.K.; funding acquisition, J.K. All authors have read and agreed to the published version of the manuscript.}

\funding{This material is based upon work supported in part by the National Science Foundation under Grant No. 2500638, and in part by institutional support from the University of Michigan–Dearborn.}

\dataavailability{The project page includes the video, code, and dataset: 
\url{https://awskhalil.github.io/plm-net/} (accessed on 8 March 2026).}

\conflictsofinterest{The authors declare no conflicts of interest.}

\appendixtitles{yes} 
\appendixstart
\appendix

\section[\appendixname~\thesection]{Constant Perception Latency}
\label{apndx: constant-latency}

The following are the results from the different constant perception latency values that we tested to validate our model PLM-Net. The interpretation of these results is similar to that in Section \ref{sec:results}, so we directly show the relevant figures and tables for each latency value.

\subsection[\appendixname~\thesubsection]{Perception Latency = 0.15 s}
\begin{figure}[H]
\includegraphics[width=0.8\linewidth]{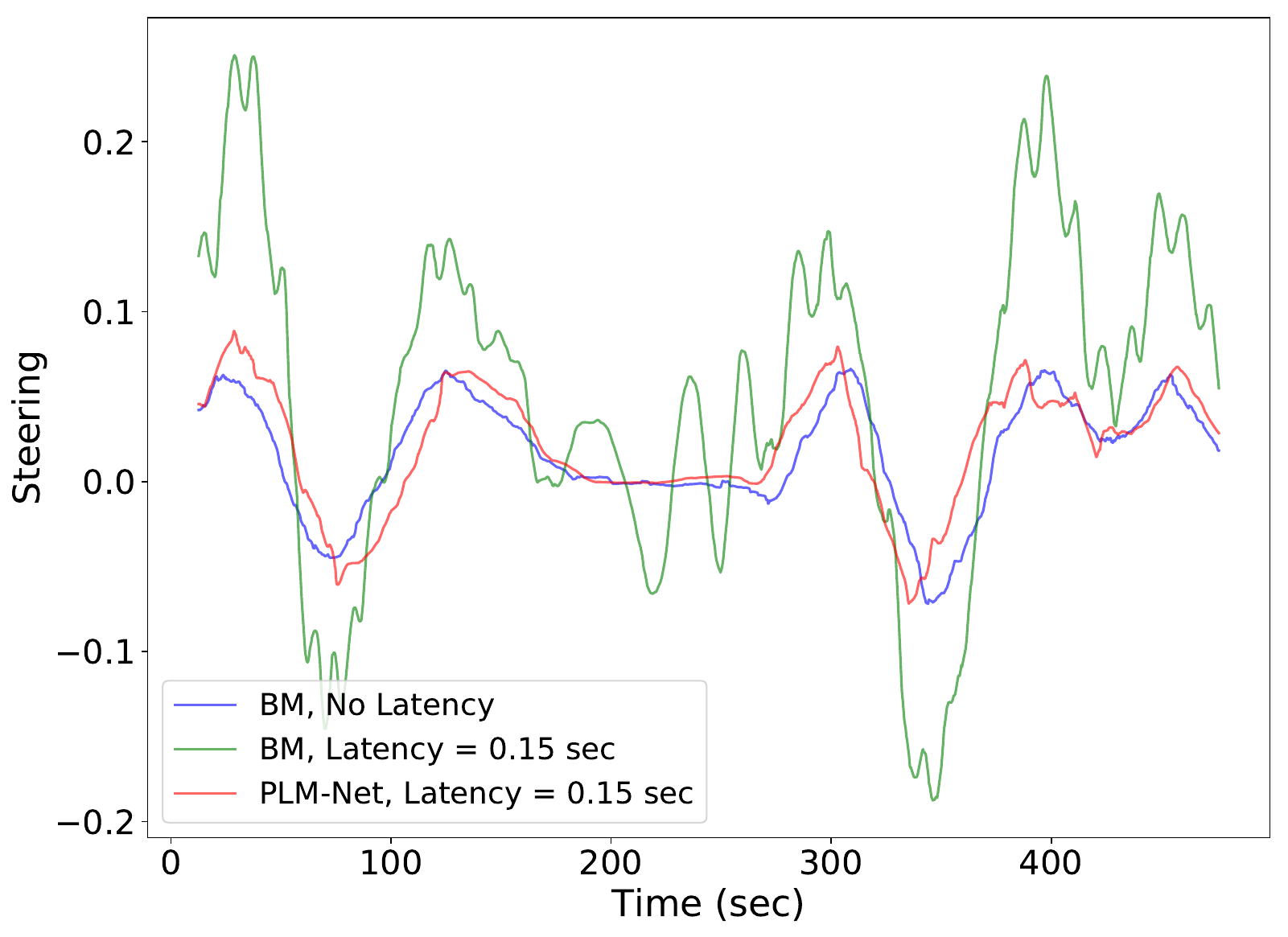}
\caption{
Qualitative 
 comparison of steering angle over time. 
The blue line represents the BM driving on the test track with no latency.
The green line represents the BM driving on the test track with $0.15$ s latency.
The red line represents the PLM-Net driving on the test track with the same latency value.
}
\label{fig:plm-delta-150-steering1}
\end{figure}

\begin{figure}[H]
\includegraphics[width=0.52\linewidth]{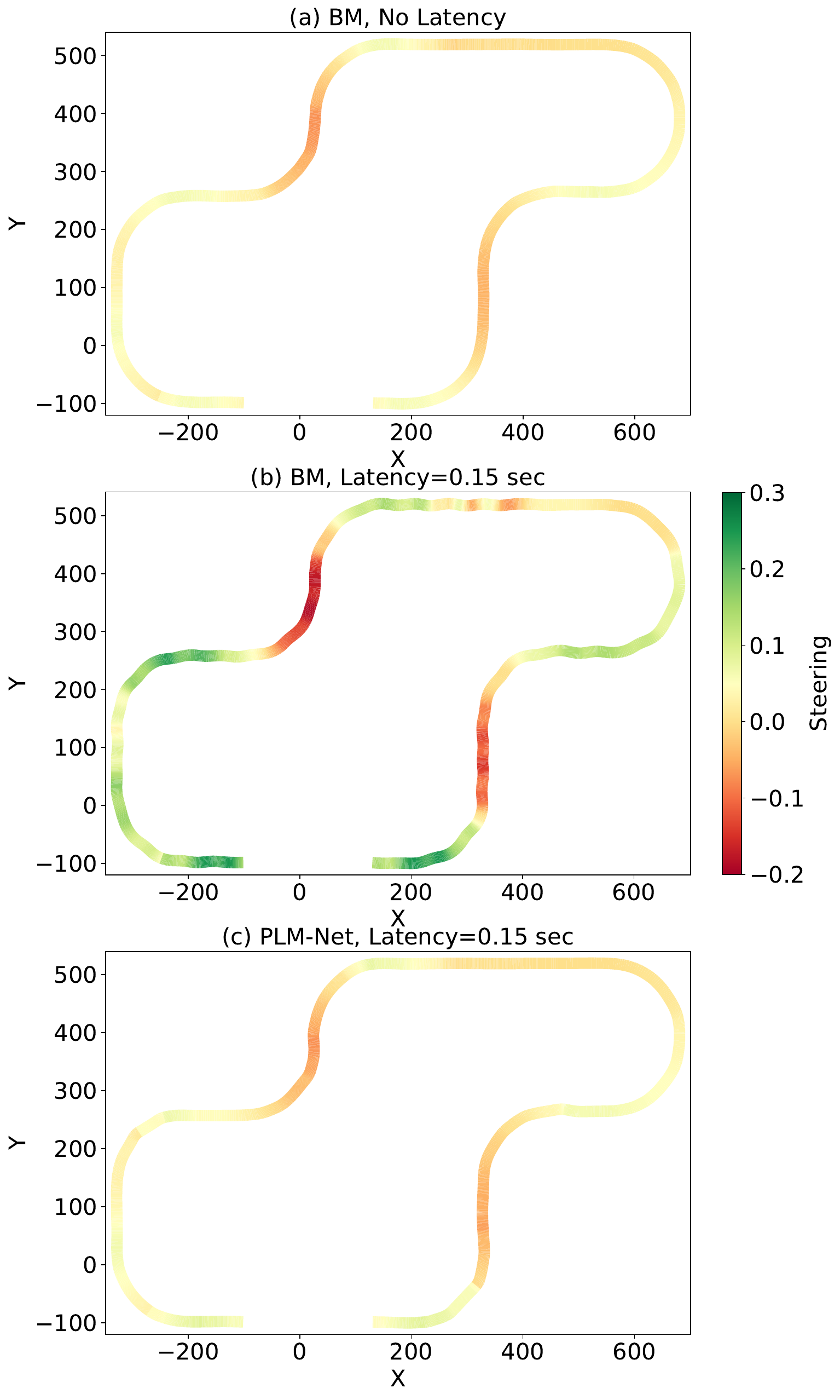}
\caption{
Vehicle trajectories colored by steering angle. (\textbf{a}) The trajectory of the BM driving with no latency. (\textbf{b}) The trajectory of the BM driving with $0.15$ s latency. (\textbf{c}) The trajectory of the PLM-Net driving with $0.15$ s latency.
}
\label{fig:plm-delta-150-steering2} 
\end{figure}

\begin{figure}[H]
\hspace{-18pt}
\includegraphics[width=1\linewidth]{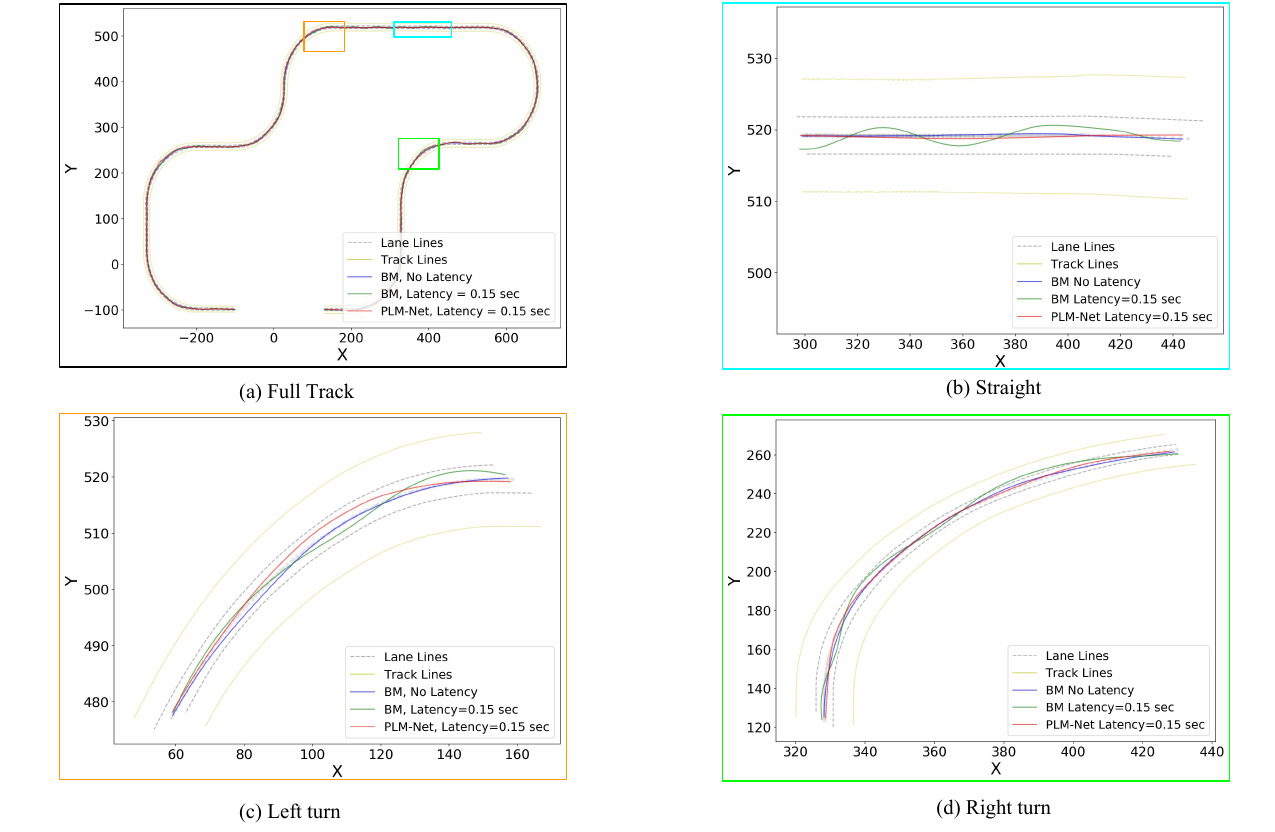}
\caption{
Trajectory 
 comparison between the BM driving with no latency (blue line), the BM driving with $0.15$ s latency (green line), and the PLM-Net driving with $0.15$ s latency (red line). (\textbf{a}) The trajectories on the full test track. (\textbf{b}) The trajectories on a straight road segment. (\textbf{c}) The trajectories on a left turn. (\textbf{d}) The trajectories on a right turn.
}
\label{fig:plm-delta-150-traj}
\end{figure}

This subsection illustrates the results when the perception latency was $0.15$ s. Figure~\ref{fig:plm-delta-150-steering1} and Table \ref{tbl:PLM-delta-150-steering} show the steering comparison against time. 
Figure~\ref{fig:plm-delta-150-steering2} compares driving trajectories colored based on steering angle values.
Figure~\ref{fig:plm-delta-150-traj} and  Table \ref{tbl:PLM-delta-150-traj} show the trajectory similarity.

\begin{table}[H]
\caption{ 
Steering Angle Similarity. Quantitative comparison of steering angle between the BM driving without latency, the BM driving with latency of $0.15$ s, and PLM-Net driving with the same latency.  Downward arrows indicate that lower values correspond to better performance.
}
\label{tbl:PLM-delta-150-steering}
\begin{tabularx}{\textwidth}{Lccc}
\toprule
\multicolumn{1}{l}{\textit{\textbf{BM, 
 No Latency vs.}}} 
  & \textbf{MAE} \(\downarrow\) 
  & \textbf{MSE} \(\downarrow\) 
  & \textbf{RMSE} \(\downarrow\) \\
\midrule  

\textit{BM, Latency = 0.15 s} & 0.171355 & 0.0570811 & 0.238916 \\ 
\midrule
\textit{PLM-Net, Latency = 0.15 s} & \textbf{0.053122 
} & \textbf{0.00691402} & \textbf{0.0831506} \\ 
\bottomrule
\end{tabularx}
\end{table}
\vspace{-9pt}
\vspace{-6pt}

\vspace{-6pt}

\begin{table}[H]
    \centering
    \caption{
    Quantitative 
 comparison 
 of driving trajectory similarity between the BM driving without latency, the BM driving with latency of $0.15$ s, and PLM-Net driving with the same latency.
 All driving models' trajectories are compared with the lane center on full track, straight road, left turn, and right~turn.   
 Downward arrows indicate that lower values correspond to better performance.
 }
    \label{tbl:PLM-delta-150-traj}
    \begin{adjustwidth}{-\extralength}{0cm}
    \footnotesize
    \begin{tabularx}{\fulllength}{lLCCCCCC}
    
    \toprule

    \multirow{3}{*}{\textbf{Figure~\ref{fig:plm-delta-150-traj} Ref.}} & \multirow{3}{*}{\textbf{Driving} \textbf{Model}} 
                                    & \multicolumn{1}{c}{\textbf{Partial}}  
                                    & \multicolumn{1}{c}{\textbf{Frechet}}    
                                    & \multicolumn{1}{c}{\textbf{Area}}  & \multicolumn{1}{c}{\textbf{Curve}}          
                                    & \multicolumn{1}{c}{\textbf{Dynamic}} & \multicolumn{1}{c}{}
                                    \\ 
                                    & & \multicolumn{1}{c}{\textbf{Curve} \boldmath{$\downarrow$}} & \multicolumn{1}{c}{\textbf{Distance} \boldmath{$\downarrow$}} 
                                    & \multicolumn{1}{c}{\textbf{Between} \boldmath{$\downarrow$}}  & \multicolumn{1}{c}{\textbf{Length} \boldmath{$\downarrow$}}          
                                    & \multicolumn{1}{c}{\textbf{Time} \boldmath{$\downarrow$}} & \multicolumn{1}{c}{\textoverline{\textbf{DTSI}} \boldmath{$\downarrow$}}
                                    \\
                                    & & \multicolumn{1}{c}{\textbf{Mapping}} & \multicolumn{1}{c}{} 
                                    & \multicolumn{1}{c}{\textbf{Curves}}  & \multicolumn{1}{c}{}          
                                    & \multicolumn{1}{c}{\textbf{Warping}} & \multicolumn{1}{c}{}
                                    \\ \hline
                                    
    \rowcolor{gray!30}
    \textbf{(a) Full Track}                  & \multicolumn{1}{l}{\textbf{BM, No Latency (ref)}}                  
                                    & \multicolumn{1}{c}{\textbf{1.19}}     & \multicolumn{1}{c}{\textbf{2.176}}    
                                    & \multicolumn{1}{c}{\textbf{4658.15}}     & \multicolumn{1}{c}{\textbf{0.349}}      
                                    & \multicolumn{1}{c}{\textbf{1253.21}}     & \multicolumn{1}{c}{\textbf{0.034}} \\ \hline
                    
    (a) Full Track                  & \multicolumn{1}{l}{BM, Latency = $0.15$ s}         
                                    & \multicolumn{1}{c}{1.931}     & \multicolumn{1}{c}{\textbf{4.087}}    
                                    & \multicolumn{1}{c}{15,946.7}     & \multicolumn{1}{c}{0.432}      
                                    & \multicolumn{1}{c}{2343.73}     & \multicolumn{1}{c}{0.266}\\ \midrule
                    
    (a) Full Track                  & \multicolumn{1}{l}{PLM-Net, Latency = $0.15$ s}    
                                    & \multicolumn{1}{c}{\textbf{1.579}}     & \multicolumn{1}{c}{4.543}    
                                    & \multicolumn{1}{c}{\textbf{13,035}}     & \multicolumn{1}{c}{\textbf{0.383}}      
                                    & \multicolumn{1}{c}{\textbf{2255.83}}     & \multicolumn{1}{c}{\textbf{0.041}}\\ \hline

    \rowcolor{cyan!10}
    \textbf{(b) Straight}                    & \multicolumn{1}{l}{\textbf{BM, No Latency (ref)}}                  
                                    & \multicolumn{1}{c}{\textbf{24.475}}     & \multicolumn{1}{c}{\textbf{5.095}}    
                                    & \multicolumn{1}{c}{\textbf{13.556}}     & \multicolumn{1}{c}{\textbf{0.077}}      
                                    & \multicolumn{1}{c}{\textbf{76.376}}     & \multicolumn{1}{c}{\textbf{0.012}} \\ \hline
                    
    (b) Straight                    & \multicolumn{1}{l}{BM, Latency = $0.15$ s }         
                                    & \multicolumn{1}{c}{229.411}     & \multicolumn{1}{c}{5.918}    
                                    & \multicolumn{1}{c}{123.118}     & \multicolumn{1}{c}{0.109}      
                                    & \multicolumn{1}{c}{213.709}     & \multicolumn{1}{c}{0.016}\\ \midrule
                    
    (b) Straight                    & \multicolumn{1}{l}{PLM-Net, Latency = $0.15$ s}    
                                    & \multicolumn{1}{c}{\textbf{63.767}}     & \multicolumn{1}{c}{\textbf{5.566}}    
                                    & \multicolumn{1}{c}{\textbf{34.741}}     & \multicolumn{1}{c}{\textbf{0.086}}      
                                    & \multicolumn{1}{c}{\textbf{97.41}}     & \multicolumn{1}{c}{\textbf{0.012}}\\ \hline

    \rowcolor{orange!20}
    \textbf{(c) Left turn}                   & \multicolumn{1}{l}{\textbf{BM, No Latency (ref)}}                  
                                    & \multicolumn{1}{c}{\textbf{0.895}}     & \multicolumn{1}{c}{\textbf{3.966}}    
                                    & \multicolumn{1}{c}{\textbf{41.402}}     & \multicolumn{1}{c}{\textbf{0.178}}      
                                    & \multicolumn{1}{c}{\textbf{67.892}}     & \multicolumn{1}{c}{\textbf{0.025}}\\ \hline
                    
    (c) Left turn                   & \multicolumn{1}{l}{BM, Latency = $0.15$ s }         
                                    & \multicolumn{1}{c}{0.932}     & \multicolumn{1}{c}{4.115}    
                                    & \multicolumn{1}{c}{98.63}     & \multicolumn{1}{c}{0.179}      
                                    & \multicolumn{1}{c}{113.299}     & \multicolumn{1}{c}{0.07}\\ \midrule
                    
    (c) Left turn                   & \multicolumn{1}{l}{PLM-Net, Latency = $0.15$ s}    
                                    & \multicolumn{1}{c}{\textbf{0.531}}     & \multicolumn{1}{c}{\textbf{2.126}}    
                                    & \multicolumn{1}{c}{\textbf{81.367}}     & \multicolumn{1}{c}{\textbf{0.141}}      
                                    & \multicolumn{1}{c}{\textbf{92.124}}     & \multicolumn{1}{c}{\textbf{0.058}}\\ \hline

    \rowcolor{green!20}
    \textbf{(d) Right turn}                  & \multicolumn{1}{l}{\textbf{BM, No Latency (ref)}}                  
                                    & \multicolumn{1}{c}{\textbf{0.293}}     & \multicolumn{1}{c}{\textbf{1.546}}    
                                    & \multicolumn{1}{c}{\textbf{90.228}}     & \multicolumn{1}{c}{\textbf{0.048}}      
                                    & \multicolumn{1}{c}{\textbf{79.036}}     & \multicolumn{1}{c}{\textbf{0.037}}\\ \hline
                    
    (d) Right turn                  & \multicolumn{1}{l}{BM, Latency = $0.15$ s }         
                                    & \multicolumn{1}{c}{0.746}     & \multicolumn{1}{c}{3.971}    
                                    & \multicolumn{1}{c}{285.641}     & \multicolumn{1}{c}{0.08}      
                                    & \multicolumn{1}{c}{169.875}     & \multicolumn{1}{c}{0.113}\\ \midrule
                    
    (d) Right turn                  & \multicolumn{1}{l}{PLM-Net, Latency = $0.15$ s}    
                                    & \multicolumn{1}{c}{\textbf{0.313}}     & \multicolumn{1}{c}{\textbf{1.882}}    
                                    & \multicolumn{1}{c}{\textbf{67.192}}     & \multicolumn{1}{c}{\textbf{0.05}}      
                                    & \multicolumn{1}{c}{\textbf{75.371}}     & \multicolumn{1}{c}{\textbf{0.098}}\\ \bottomrule
                    
    \end{tabularx}
    \end{adjustwidth}
\end{table}

\subsection[\appendixname~\thesubsection]{Perception Latency = 0.25 s}
\begin{figure}[H]
\includegraphics[width=0.7\linewidth]{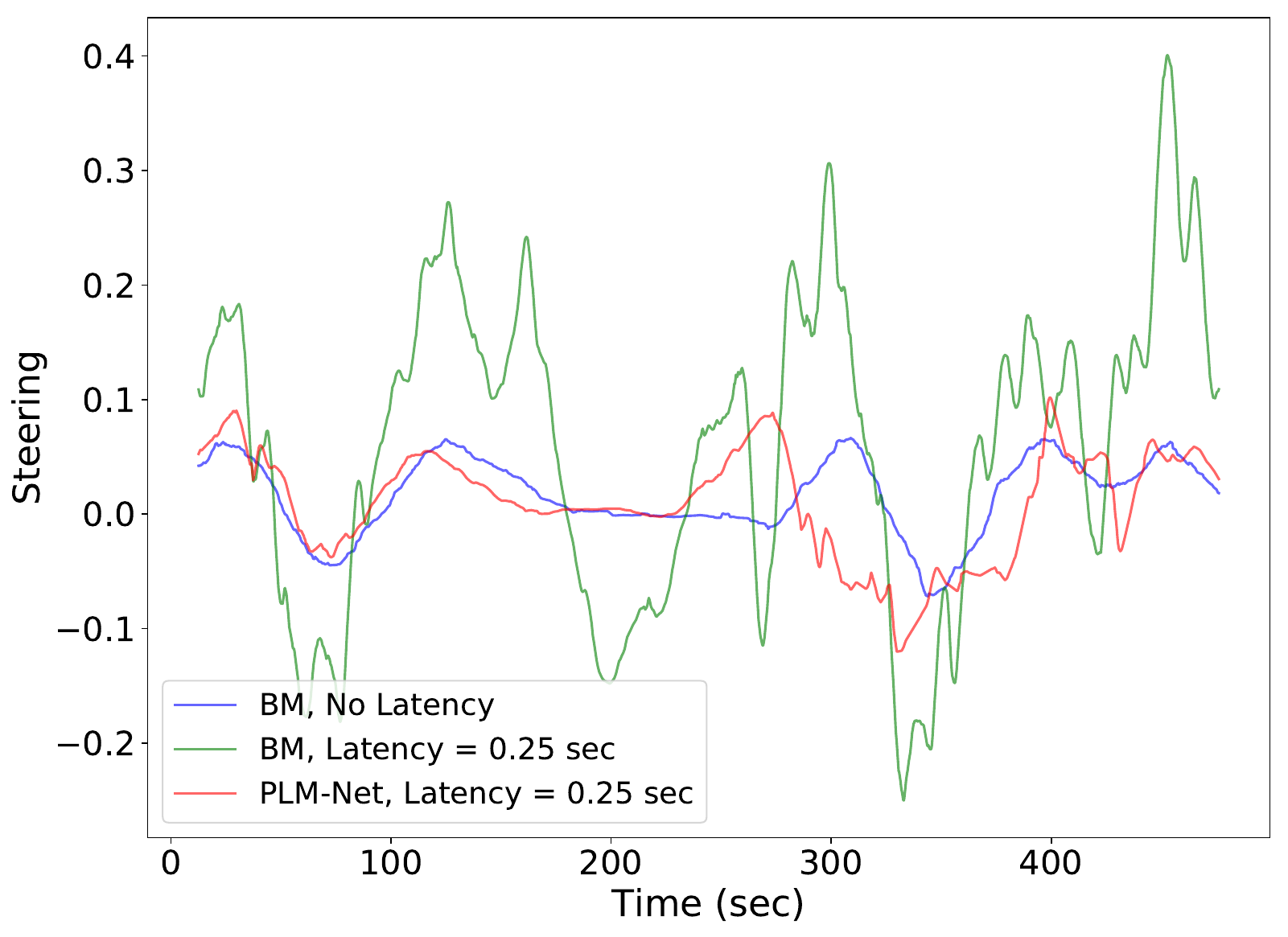}
\caption{
Qualitative 
 comparison of steering angle over time.  
The blue line represents the BM driving on the test track with no latency.
The green line represents the BM driving on the test track with $0.25$ s latency.
The red line represents the PLM-Net driving on the test track with the same latency value.
}
\label{fig:plm-delta-250-steering1}
\end{figure}

\begin{figure}[H]
\includegraphics[width=0.6\linewidth]{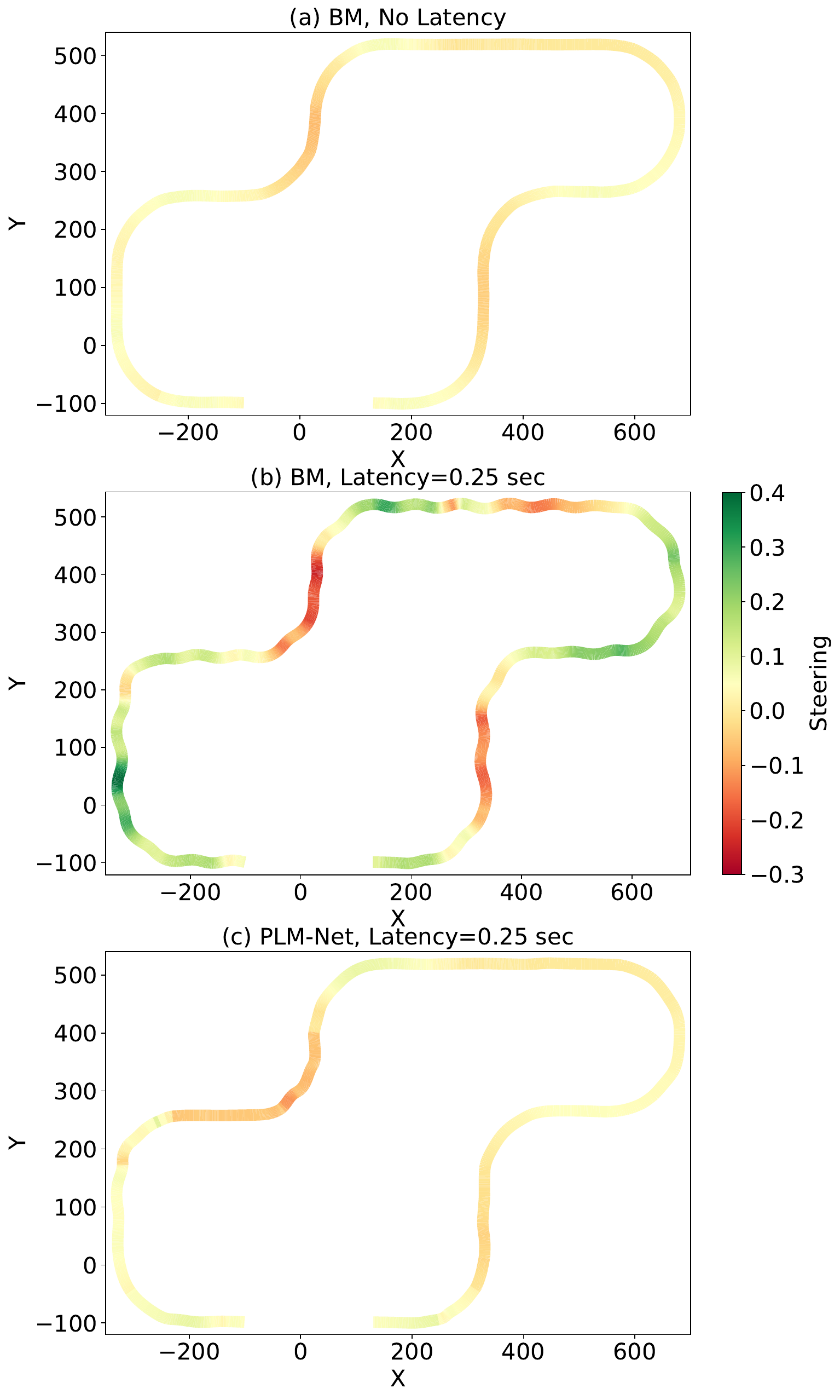}
\caption{
Vehicle trajectories colored by steering angle. (\textbf{a}) The trajectory of the BM driving with no latency. (\textbf{b}) The trajectory of the BM driving with $0.25$ s latency. (\textbf{c}) The trajectory of the PLM-Net driving with $0.25$ s latency.
}
\label{fig:plm-delta-250-steering2}
\end{figure}

\begin{figure}[H]
\hspace{-18pt}
\includegraphics[width=1.0\linewidth]{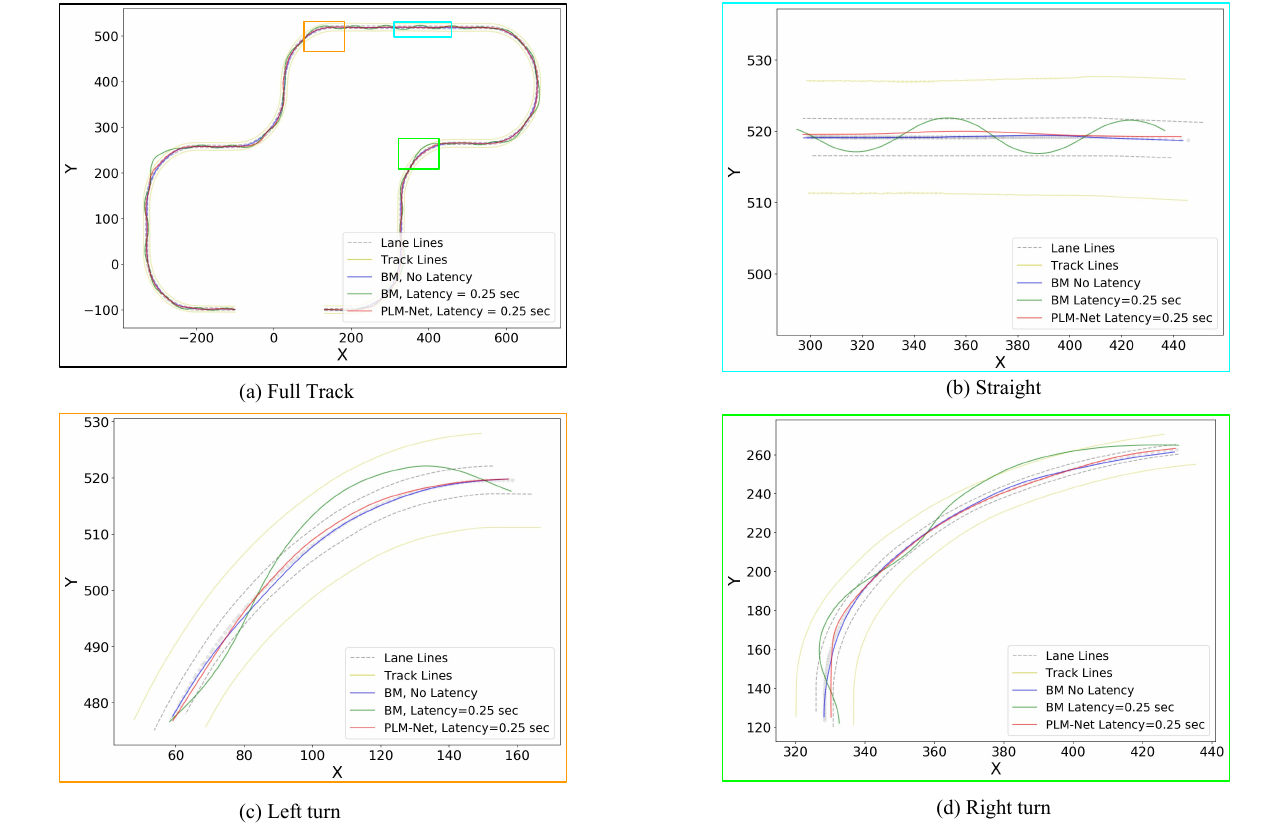}
\caption{
Trajectory 
 comparison between the BM driving with no latency (blue line), the BM driving with $0.25$ s latency (green line), and the PLM-Net driving with $0.25$ s latency (red line). (\textbf{a}) The trajectories on the full test track. (\textbf{b}) The trajectories on a straight road segment. (\textbf{c}) The trajectories on a left turn. (\textbf{d}) The trajectories on a right turn.
}
\label{fig:plm-delta-250-traj}
\end{figure}

This subsection illustrates the results when the perception latency was $0.25$ s. Figure~\ref{fig:plm-delta-250-steering1} and Table \ref{tbl:PLM-delta-250-steering} show the steering comparison against time. 
Figure~\ref{fig:plm-delta-250-steering2} compares driving trajectories colored based on steering angle values.
Figure~\ref{fig:plm-delta-250-traj} and  Table \ref{tbl:PLM-delta-250-traj} show the trajectory similarity.

\vspace{-6pt}
\begin{table}[H]
\caption{ 
Steering angle similarity. Quantitative comparison of steering angle between the BM driving without latency, the BM driving with latency of $0.25$ s, and PLM-Net driving with the same latency. Downward arrows indicate that lower values correspond to better performance.
}
\label{tbl:PLM-delta-250-steering}
\centering
\begin{tabularx}{\textwidth}{Lccc}
\toprule
\multicolumn{1}{l}{\textit{\textbf{BM, 
 No Latency vs.}}} 
  & \textbf{MAE} \(\downarrow\) 
  & \textbf{MSE} \(\downarrow\) 
  & \textbf{RMSE} \(\downarrow\) \\
\midrule  

\textit{BM, Latency = 0.25 s} & 0.334727 & 0.18581 & 0.431057 \\ 
\midrule
\textit{PLM-Net, Latency = 0.25 s} & \textbf{0.0921437 
} & \textbf{0.0192333} & \textbf{0.138684} \\ 
\bottomrule
\end{tabularx}
\end{table}

\vspace{-6pt}
\begin{table}[H]
    \caption{
    Quantitative 
 comparison 
 of driving trajectory similarity between the BM driving without latency, the BM driving with latency of $0.25$ s, and PLM-Net driving with the same latency. 
 All driving models' trajectories are compared with the lane center on full track, straight road, left turn, and right~turn.
 Downward arrows indicate that lower values correspond to better performance.
 }
    \label{tbl:PLM-delta-250-traj}
    \begin{adjustwidth}{-\extralength}{0cm}
    \footnotesize
    \begin{tabularx}{\fulllength}{lLCCCCCC}
    
    \toprule 

    \multirow{3}{*}{\textbf{Figure~\ref{fig:plm-delta-250-traj} ref.}} & \multirow{3}{*}{\textbf{Driving} \textbf{Model}} 
                                    & \multicolumn{1}{c}{\textbf{Partial}}  
                                    & \multicolumn{1}{c}{\textbf{Frechet}}    
                                    & \multicolumn{1}{c}{\textbf{Area}}  & \multicolumn{1}{c}{\textbf{Curve}}          
                                    & \multicolumn{1}{c}{\textbf{Dynamic}} & \multicolumn{1}{c}{}
                                    \\ 
                                    & & \multicolumn{1}{c}{\textbf{Curve} \boldmath{$\downarrow$}} & \multicolumn{1}{c}{\textbf{Distance} \boldmath{$\downarrow$}} 
                                    & \multicolumn{1}{c}{\textbf{Between} \boldmath{$\downarrow$}}  & \multicolumn{1}{c}{\textbf{Length} \boldmath{$\downarrow$}}          
                                    & \multicolumn{1}{c}{\textbf{Time} \boldmath{$\downarrow$}} & \multicolumn{1}{c}{\textoverline{\textbf{DTSI}} \boldmath{$\downarrow$}}
                                    \\
                                    & & \multicolumn{1}{c}{\textbf{Mapping}} & \multicolumn{1}{c}{} 
                                    & \multicolumn{1}{c}{\textbf{Curves}}  & \multicolumn{1}{c}{}          
                                    & \multicolumn{1}{c}{\textbf{Warping}} & \multicolumn{1}{c}{}
                                    \\ \hline
                                    
    \rowcolor{gray!30}
    \textbf{(a) Full Track}                  & \multicolumn{1}{l}{\textbf{BM, No Latency (ref)}}                  
                                    & \multicolumn{1}{c}{\textbf{1.19}}     & \multicolumn{1}{c}{\textbf{2.176}}    
                                    & \multicolumn{1}{c}{\textbf{4658.15}}     & \multicolumn{1}{c}{\textbf{0.349}}      
                                    & \multicolumn{1}{c}{\textbf{1253.21}}     & \multicolumn{1}{c}{\textbf{0.034}} \\ \hline
                    
    (a) Full Track                  & \multicolumn{1}{l}{BM, Latency = $0.25$ s}         
                                    & \multicolumn{1}{c}{8.077}     & \multicolumn{1}{c}{18.917}    
                                    & \multicolumn{1}{c}{\textbf{25,958.2}}     & \multicolumn{1}{c}{2.09}      
                                    & \multicolumn{1}{c}{7459.15}     & \multicolumn{1}{c}{0.36}\\ \midrule
                    
    (a) Full Track                  & \multicolumn{1}{l}{PLM-Net, Latency = $0.25$ s}    
                                    & \multicolumn{1}{c}{\textbf{3.47}}     & \multicolumn{1}{c}{\textbf{5.357}}    
                                    & \multicolumn{1}{c}{91,195.1}     & \multicolumn{1}{c}{\textbf{0.814}}      
                                    & \multicolumn{1}{c}{\textbf{3637.4}}     & \multicolumn{1}{c}{\textbf{0.046}}\\ \hline

    \rowcolor{cyan!10}
    \textbf{(b) Straight}                    & \multicolumn{1}{l}{\textbf{BM, No Latency (ref)}}                  
                                    & \multicolumn{1}{c}{\textbf{24.475}}     & \multicolumn{1}{c}{\textbf{5.095}}    
                                    & \multicolumn{1}{c}{\textbf{13.556}}     & \multicolumn{1}{c}{\textbf{0.077}}      
                                    & \multicolumn{1}{c}{\textbf{76.376}}     & \multicolumn{1}{c}{\textbf{0.012}} \\ \hline
                    
    (b) Straight                    & \multicolumn{1}{l}{BM, Latency = $0.25$ s }         
                                    & \multicolumn{1}{c}{680.624}     & \multicolumn{1}{c}{4.756}    
                                    & \multicolumn{1}{c}{359.311}     & \multicolumn{1}{c}{0.122}      
                                    & \multicolumn{1}{c}{550.559}     & \multicolumn{1}{c}{0.077}\\ \midrule
                    
    (b) Straight                    & \multicolumn{1}{l}{PLM-Net, Latency = $0.25$ s}    
                                    & \multicolumn{1}{c}{\textbf{150.696}}     & \multicolumn{1}{c}{\textbf{3.053}}    
                                    & \multicolumn{1}{c}{\textbf{78.985}}     & \multicolumn{1}{c}{\textbf{0.072}}      
                                    & \multicolumn{1}{c}{\textbf{122.401}}     & \multicolumn{1}{c}{\textbf{0.046}}\\ \hline

    \rowcolor{orange!20}
    \textbf{(c) Left turn}                   & \multicolumn{1}{l}{\textbf{BM, No Latency (ref)}}                  
                                    & \multicolumn{1}{c}{\textbf{0.895}}     & \multicolumn{1}{c}{\textbf{3.966}}    
                                    & \multicolumn{1}{c}{\textbf{41.402}}     & \multicolumn{1}{c}{\textbf{0.178}}      
                                    & \multicolumn{1}{c}{\textbf{67.892}}     & \multicolumn{1}{c}{\textbf{0.025}}\\ \hline
                    
    (c) Left turn                   & \multicolumn{1}{l}{BM, Latency = $0.25$ s }         
                                    & \multicolumn{1}{c}{2.999}     & \multicolumn{1}{c}{5.542}    
                                    & \multicolumn{1}{c}{349.446}     & \multicolumn{1}{c}{\textbf{0.156}}      
                                    & \multicolumn{1}{c}{326.643}     & \multicolumn{1}{c}{0.13}\\ \midrule
                    
    (c) Left turn                   & \multicolumn{1}{l}{PLM-Net, Latency = $0.25$ s}    
                                    & \multicolumn{1}{c}{\textbf{0.973}}     & \multicolumn{1}{c}{\textbf{4.765}}    
                                    & \multicolumn{1}{c}{\textbf{61.525}}     & \multicolumn{1}{c}{0.218}      
                                    & \multicolumn{1}{c}{\textbf{89.88}}     & \multicolumn{1}{c}{\textbf{0.093}}\\ \hline

    \rowcolor{green!20}
    \textbf{(d) Right turn}                  & \multicolumn{1}{l}{\textbf{BM, No Latency (ref)}}                  
                                    & \multicolumn{1}{c}{\textbf{0.293}}     & \multicolumn{1}{c}{\textbf{1.546}}    
                                    & \multicolumn{1}{c}{\textbf{90.228}}     & \multicolumn{1}{c}{\textbf{0.048}}      
                                    & \multicolumn{1}{c}{\textbf{79.036}}     & \multicolumn{1}{c}{\textbf{0.037}}\\ \hline
                    
    (d) Right turn                  & \multicolumn{1}{l}{BM, Latency = $0.25$ s }         
                                    & \multicolumn{1}{c}{\textbf{1.79}}     & \multicolumn{1}{c}{9.468}    
                                    & \multicolumn{1}{c}{810.979}     & \multicolumn{1}{c}{0.222}      
                                    & \multicolumn{1}{c}{438.145}     & \multicolumn{1}{c}{0.11}\\ \midrule
                    
    (d) Right turn                  & \multicolumn{1}{l}{PLM-Net, Latency = $0.25$ s}    
                                    & \multicolumn{1}{c}{0.199}     & \multicolumn{1}{c}{\textbf{1.971}}    
                                    & \multicolumn{1}{c}{\textbf{108.645}}     & \multicolumn{1}{c}{\textbf{0.032}}      
                                    & \multicolumn{1}{c}{\textbf{103.168}}     & \multicolumn{1}{c}{\textbf{0.065}}\\  \bottomrule
                    
    \end{tabularx}
    \end{adjustwidth}
\end{table}

\subsection[\appendixname~\thesubsection]{Perception Latency = 0.30 s}
\begin{figure}[H]
\includegraphics[width=0.6\linewidth]{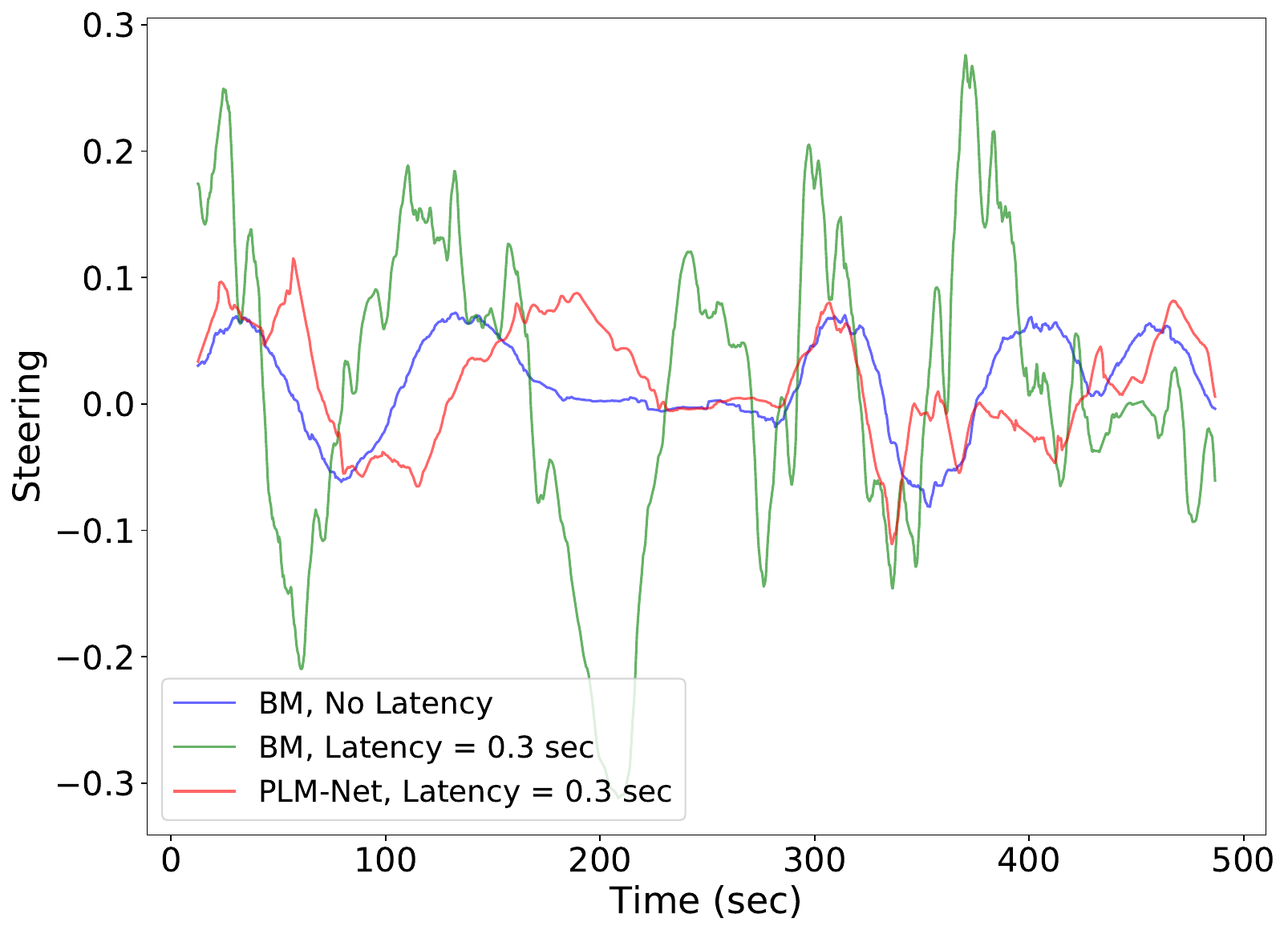}
\caption{
\textls[-15]{Qualitative 
 comparison of steering angle over time. 
The blue line represents the BM driving on the test track with no latency.
The green line represents the BM driving on the test track with $0.3$ s latency.
The red line represents the PLM-Net driving on the test track with the same latency value.}
}
\label{fig:plm-delta-300-steering1}
\end{figure}

\begin{figure}[H]
\includegraphics[width=0.5\linewidth]{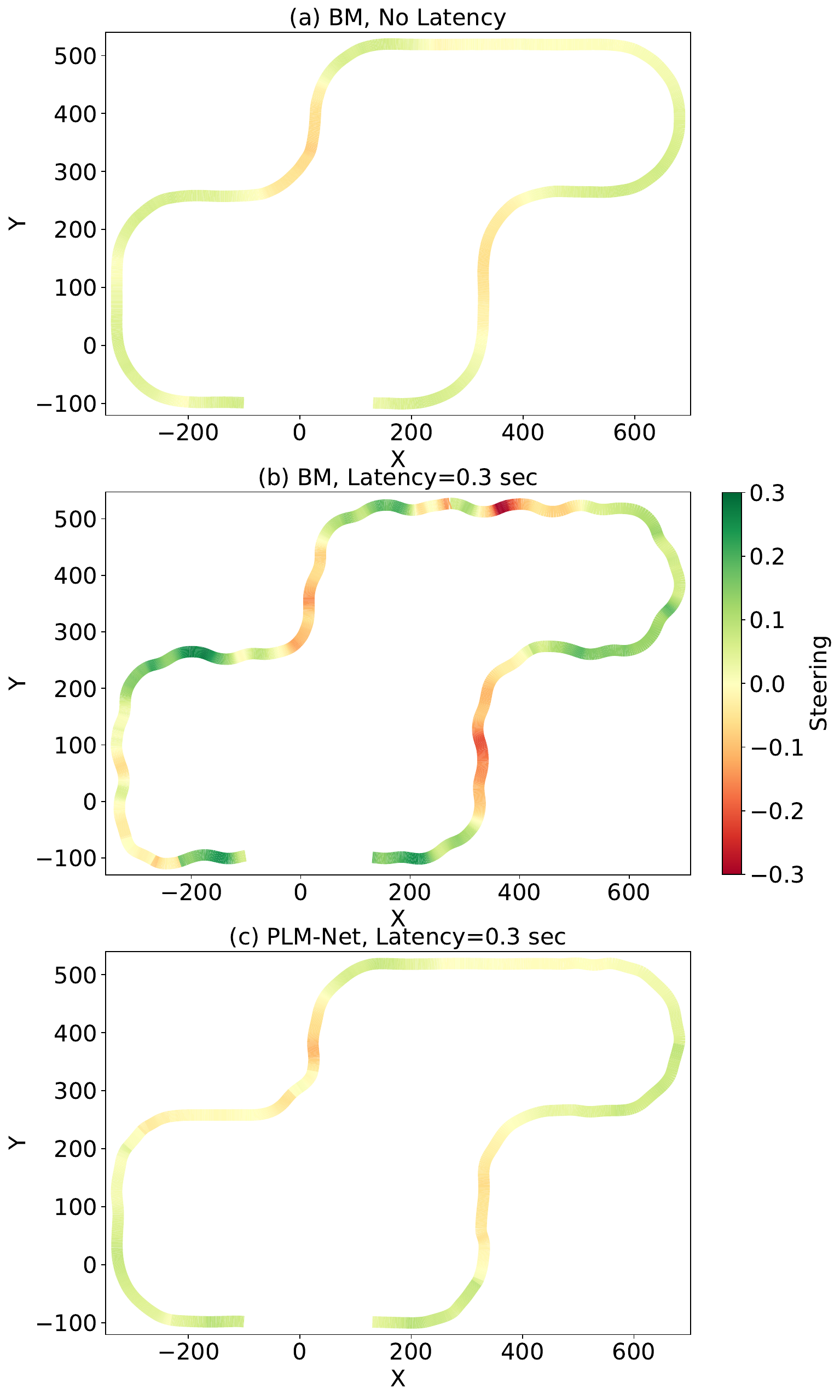}
\caption{
Vehicle trajectories colored by steering angle. (\textbf{a}) The trajectory of the BM driving with no latency. (\textbf{b}) The trajectory of the BM driving with $0.3$ s latency. (\textbf{c}) The trajectory of the PLM-Net driving with $0.3$ s latency.
}
\label{fig:plm-delta-300-steering2}
\end{figure}

\begin{figure}[H]

\hspace{-15pt}\includegraphics[width=1\linewidth]{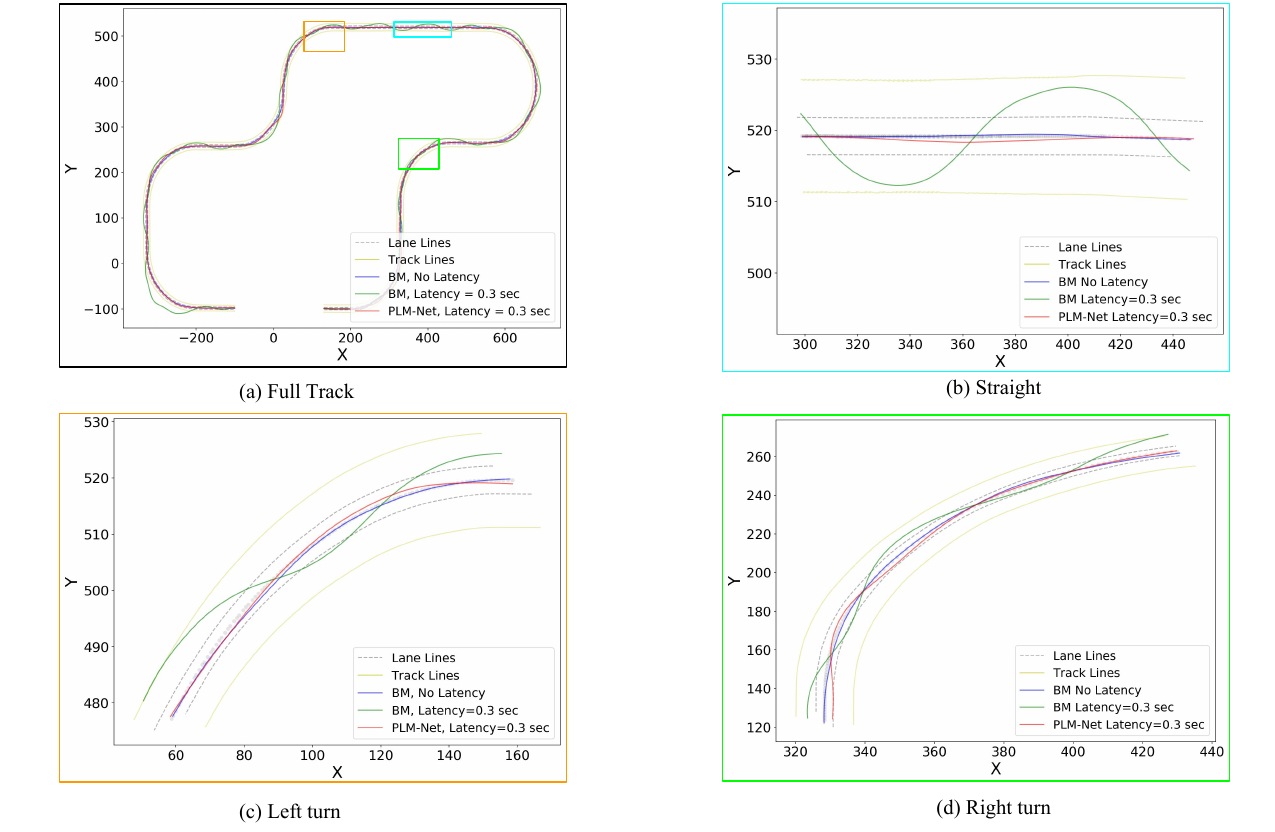}
\caption{
Trajectory 
 comparison between the BM driving with no latency (blue line), the BM driving with $0.3$ s latency (green line), and the PLM-Net driving with $0.3$ s latency (red line). (\textbf{a}) The trajectories on the full test track. (\textbf{b}) The trajectories on a straight road segment. (\textbf{c}) The trajectories on a left turn. (\textbf{d}) The trajectories on a right turn.
}
\label{fig:plm-delta-300-traj}
\end{figure}

This subsection illustrates the results when the perception latency was $0.30$ s. Figure~\ref{fig:plm-delta-300-steering1} and Table \ref{tbl:PLM-delta-300-steering} show the steering comparison against time. 
Figure~\ref{fig:plm-delta-300-steering2} compares driving trajectories colored based on steering angle values.
Figure~\ref{fig:plm-delta-300-traj} and  Table \ref{tbl:PLM-delta-300-traj} show the trajectory similarity.

\vspace{-6pt}

\begin{table}[H]
\caption{ 
Steering angle similarity. Quantitative comparison of steering angle error between the BM driving without latency, the BM driving with latency of $0.3$ s, and PLM-Net driving with the same latency.
 Downward arrows indicate that lower values correspond to better performance.
}
\label{tbl:PLM-delta-300-steering}
\begin{tabularx}{\textwidth}{Lccc}
\toprule
\multicolumn{1}{l}{\textit{\textbf{BM, 
 No Latency vs.}}} 
  & \textbf{MAE} \(\downarrow\) 
  & \textbf{MSE} \(\downarrow\) 
  & \textbf{RMSE} \(\downarrow\) \\ 
\midrule  

\textit{BM, Latency = 0.3 s} & 0.226148 & 0.0968093 & 0.311142 \\ 
\midrule
\textit{PLM-Net, Latency = 0.3 s} & \textbf{0.0710674 
} & \textbf{0.0107641} & \textbf{0.10375} \\ 
\bottomrule
\end{tabularx}
\end{table}
\vspace{-9pt}
\vspace{-9pt}
\begin{table}[H]
    \caption{
    Quantitative 
 comparison of driving trajectory similarity between the BM driving without latency, the BM driving with latency of $0.3$ s, and PLM-Net driving with the same latency. 
 All driving models' trajectories are compared with the lane center on full track, straight road, left turn, and right~turn.
 Downward arrows indicate that lower values correspond to better performance.
 }
    \label{tbl:PLM-delta-300-traj}
    \begin{adjustwidth}{-\extralength}{0cm}
    \footnotesize
    
    \begin{tabularx}{\fulllength}{lLCCCCCC}
    
    \toprule 

    \multirow{3}{*}{\textbf{Figure~\ref{fig:plm-delta-300-traj} Ref.}} & \multirow{3}{*}{\textbf{Driving} \textbf{Model}} 
                                    & \multicolumn{1}{c}{\textbf{Partial}}  
                                    & \multicolumn{1}{c}{\textbf{Frechet}}    
                                    & \multicolumn{1}{c}{\textbf{Area}}  & \multicolumn{1}{c}{\textbf{Curve}}          
                                    & \multicolumn{1}{c}{\textbf{Dynamic}} & \multicolumn{1}{c}{\textoverline{\textbf{DTSI}}}
                                    \\ 
                                    & & \multicolumn{1}{c}{\textbf{Curve} \boldmath{$\downarrow$}} & \multicolumn{1}{c}{\textbf{Distance} \boldmath{$\downarrow$}} 
                                    & \multicolumn{1}{c}{\textbf{Between} \boldmath{$\downarrow$}}  & \multicolumn{1}{c}{\textbf{Length} \boldmath{$\downarrow$}}          
                                    & \multicolumn{1}{c}{\textbf{Time} \boldmath{$\downarrow$}} & \multicolumn{1}{c}{\boldmath{$\downarrow$}}
                                    \\
                                    & & \multicolumn{1}{c}{\textbf{Mapping}} & \multicolumn{1}{c}{} 
                                    & \multicolumn{1}{c}{\textbf{Curves}}  & \multicolumn{1}{c}{}          
                                    & \multicolumn{1}{c}{\textbf{Warping}} & \multicolumn{1}{c}{}
                                    \\ \hline 
                                    
    \rowcolor{gray!30}
    \textbf{(a) Full Track}                  & \multicolumn{1}{l}{\textbf{BM, No Latency (ref)}}                  
                                    & \multicolumn{1}{c}{\textbf{1.19}}     & \multicolumn{1}{c}{\textbf{2.176}}    
                                    & \multicolumn{1}{c}{\textbf{4658.15}}     & \multicolumn{1}{c}{\textbf{0.349}}      
                                    & \multicolumn{1}{c}{\textbf{1253.21}}     & \multicolumn{1}{c}{\textbf{0.034}} \\ \hline
                    
    (a) Full Track                  & \multicolumn{1}{l}{BM, Latency = $0.3$ s}         
                                    & \multicolumn{1}{c}{16.097}     & \multicolumn{1}{c}{28.429}    
                                    & \multicolumn{1}{c}{93,980.4}     & \multicolumn{1}{c}{4.494}      
                                    & \multicolumn{1}{c}{15,235.3}     & \multicolumn{1}{c}{0.391}\\ \midrule
                    
    (a) Full Track                  & \multicolumn{1}{l}{PLM-Net, Latency = $0.3$ s}    
                                    & \multicolumn{1}{c}{\textbf{2.189}}     & \multicolumn{1}{c}{\textbf{7.033}}    
                                    & \multicolumn{1}{c}{\textbf{80,617.1}}     & \multicolumn{1}{c}{\textbf{0.606}}      
                                    & \multicolumn{1}{c}{\textbf{3524.91}}     & \multicolumn{1}{c}{\textbf{0.02}}\\ \hline

    \rowcolor{cyan!10}
    \textbf{(b) Straight}                    & \multicolumn{1}{l}{\textbf{BM, No Latency (ref)}}                  
                                    & \multicolumn{1}{c}{\textbf{24.475}}     & \multicolumn{1}{c}{\textbf{5.095}}    
                                    & \multicolumn{1}{c}{\textbf{13.556}}     & \multicolumn{1}{c}{\textbf{0.077}}      
                                    & \multicolumn{1}{c}{\textbf{76.376}}     & \multicolumn{1}{c}{\textbf{0.012}} \\ \hline
                    
    (b) Straight                    & \multicolumn{1}{l}{BM, Latency = $0.3$ s }         
                                    & \multicolumn{1}{c}{1046.2}     & \multicolumn{1}{c}{7.968}    
                                    & \multicolumn{1}{c}{631.198}     & \multicolumn{1}{c}{0.203}      
                                    & \multicolumn{1}{c}{911.435}     & \multicolumn{1}{c}{0.192}\\ \midrule
                    
    (b) Straight                    & \multicolumn{1}{l}{PLM-Net, Latency = $0.3$ s}    
                                    & \multicolumn{1}{c}{\textbf{97.525}}     & \multicolumn{1}{c}{\textbf{6.002}}    
                                    & \multicolumn{1}{c}{\textbf{59.361}}     & \multicolumn{1}{c}{\textbf{0.108}}      
                                    & \multicolumn{1}{c}{\textbf{117.782}}     & \multicolumn{1}{c}{\textbf{0.014}}\\ \hline

    \rowcolor{orange!20}
    \textbf{(c) Left turn}                   & \multicolumn{1}{l}{\textbf{BM, No Latency (ref)}}                  
                                    & \multicolumn{1}{c}{\textbf{0.895}}     & \multicolumn{1}{c}{\textbf{3.966}}    
                                    & \multicolumn{1}{c}{\textbf{41.402}}     & \multicolumn{1}{c}{\textbf{0.178}}      
                                    & \multicolumn{1}{c}{\textbf{67.892}}     & \multicolumn{1}{c}{\textbf{0.025}}\\ \hline
                    
    (c) Left turn                   & \multicolumn{1}{l}{BM, Latency = $0.3$ s }         
                                    & \multicolumn{1}{c}{3.088}     & \multicolumn{1}{c}{8.825}    
                                    & \multicolumn{1}{c}{422.69}     & \multicolumn{1}{c}{0.556}      
                                    & \multicolumn{1}{c}{398.425}     & \multicolumn{1}{c}{0.161}\\ \midrule
                    
    (c) Left turn                   & \multicolumn{1}{l}{PLM-Net, Latency = $0.3$ s}    
                                    & \multicolumn{1}{c}{\textbf{0.514}}     & \multicolumn{1}{c}{\textbf{1.047}}    
                                    & \multicolumn{1}{c}{\textbf{60.492}}     & \multicolumn{1}{c}{\textbf{0.019}}      
                                    & \multicolumn{1}{c}{\textbf{69.811}}     & \multicolumn{1}{c}{\textbf{0.034}}\\ \hline

    \rowcolor{green!20}
    \textbf{(d) Right turn}                  & \multicolumn{1}{l}{\textbf{BM, No Latency (ref)}}                  
                                    & \multicolumn{1}{c}{\textbf{0.293}}     & \multicolumn{1}{c}{\textbf{1.546}}    
                                    & \multicolumn{1}{c}{\textbf{90.228}}     & \multicolumn{1}{c}{\textbf{0.048}}      
                                    & \multicolumn{1}{c}{\textbf{79.036}}     & \multicolumn{1}{c}{\textbf{0.037}}\\ \hline
                    
    (d) Right turn                  & \multicolumn{1}{l}{BM, Latency = $0.3$ s }         
                                    & \multicolumn{1}{c}{1.235}     & \multicolumn{1}{c}{9.1}    
                                    & \multicolumn{1}{c}{1321.13}     & \multicolumn{1}{c}{0.2}      
                                    & \multicolumn{1}{c}{545.015}     & \multicolumn{1}{c}{0.482}\\ \midrule
                    
    (d) Right turn                  & \multicolumn{1}{l}{PLM-Net, Latency = $0.3$ s}    
                                    & \multicolumn{1}{c}{\textbf{0.21}}     & \multicolumn{1}{c}{\textbf{2.529}}    
                                    & \multicolumn{1}{c}{\textbf{162.868}}     & \multicolumn{1}{c}{\textbf{0.039}}      
                                    & \multicolumn{1}{c}{\textbf{120.484}}     & \multicolumn{1}{c}{\textbf{0.148}}\\ \bottomrule
                    
    \end{tabularx}
    \end{adjustwidth}
\end{table}

\section[\appendixname~\thesection]{Training Results}
\label{apndx: training-results}
\begin{figure}[H]
\hspace{-3.5em}\includegraphics[width=1\linewidth]{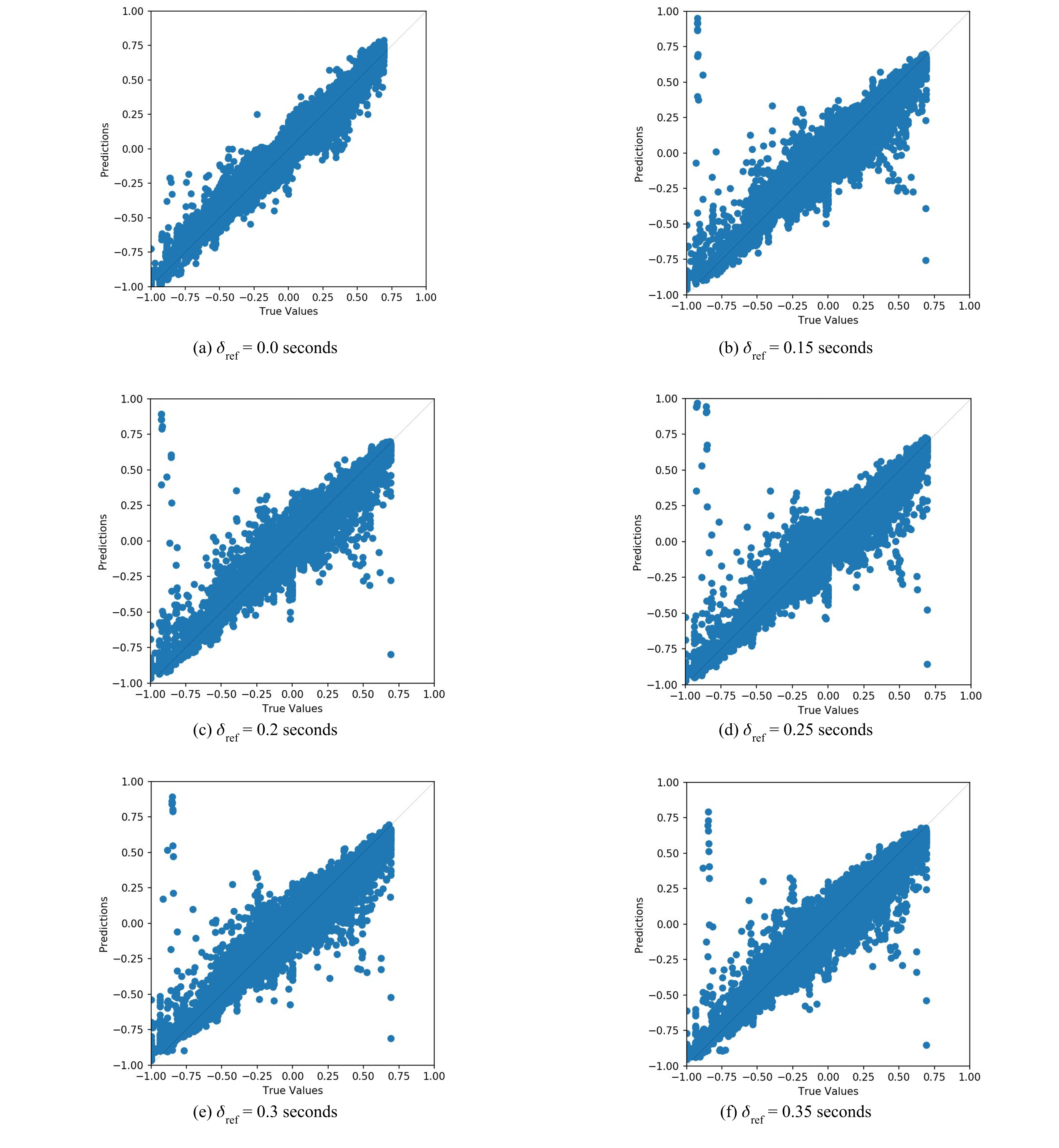}
\caption{
PLM-Net models training results. Each sub-figure shows the predicted vs ground truth steering angle. (\textbf{a}) BM training results where $\delta_{ref}=0.0$. 
(\textbf{b}--\textbf{f}) TAPM training results for all other values of $\delta_{ref}$, as the TAPM predicts five future steering angles.
}
\label{fig:training-results}
\end{figure}

Figure~\ref{fig:training-results} shows the training results for the PLM-Net models. 
Each sub-figure shows the predicted vs ground truth steering angle. 
Figure~\ref{fig:training-results}a show the BM training results where $\delta_{ref}=0.0$. 
Figure~\ref{fig:training-results}b--f show the TAPM training results for all other values of $\delta_{ref}$, as the TAPM predicts five future steering angles.

\begin{adjustwidth}{-\extralength}{0cm}

\reftitle{References}

\PublishersNote{}
\end{adjustwidth}

\begin{thebibliography}{999}
\bibitem[Kwon and Choe(2007)]{kwon2007enhanced}
Kwon, J.; Choe, Y.
\newblock Enhanced facilitatory neuronal dynamics for delay compensation.
\newblock In \emph{Proceedings of the 2007 International Joint Conference on Neural
Networks}; IEEE: Piscataway, NJ, USA, 2007; pp. 2040--2045.
\bibitem[Li et~al.(2020)Li, Wang, and Ramanan]{li2020towards}
Li, M.; Wang, Y.X.; Ramanan, D.
\newblock Towards streaming perception.
\newblock In \textit{Proceedings of the European Conference on Computer Vision};
Springer: Cham, Switzerland, 2020; pp. 473--488.
\bibitem[Ho and Griffiths(2022)]{ho2022cognitive}
Ho, M.K.; Griffiths, T.L.
\newblock Cognitive science as a source of forward and inverse models of human
decisions for robotics and control.
\newblock {\em Annu. Rev. Control Robot. Auton. Syst.} {\bf
2022}, {\em 5},~33--53. [\href{http://dx.doi.org/10.1146/annurev-control-042920-015547}{CrossRef}]
\bibitem[Xu et~al.(2020)Xu, Peng, and Tang]{xu2020preview}
Xu, S.; Peng, H.; Tang, Y.
\newblock Preview path tracking control with delay compensation for autonomous
vehicles.
\newblock {\em IEEE Trans. Intell. Transp. Syst.} {\bf
2020}, {\em 22},~2979--2989. [\href{http://dx.doi.org/10.1109/TITS.2020.2978417}{CrossRef}]
\bibitem[Liu et~al.(2018)Liu, Liu, Liu, Chen, Zhang, Li, and
Ji]{liu2018hierarchical}
Liu, Q.; Liu, Y.; Liu, C.; Chen, B.; Zhang, W.; Li, L.; Ji, X.
\newblock Hierarchical lateral control scheme for autonomous vehicle with
uneven time delays induced by vision sensors.
\newblock {\em Sensors} {\bf 2018}, {\em 18},~2544. [\href{http://dx.doi.org/10.3390/s18082544}{CrossRef}]
\bibitem[Chen and Huang(2017)]{chen_end--end_2017}
Chen, Z.; Huang, X.
\newblock End-to-end learning for lane keeping of self-driving cars.
\newblock In \emph{Proceedings of the IEEE Intelligent Vehicles Symposium (IV)};
IEEE: Piscataway, NJ, USA, 2017; pp. 1856--1860. [\href{http://dx.doi.org/10.1109/IVS.2017.7995975}{CrossRef}]
\bibitem[Khalil and Kwon(2023)]{khalil2023anec}
Khalil, A.; Kwon, J.
\newblock ANEC: Adaptive Neural Ensemble Controller for Mitigating Latency
Problems in Vision-Based Autonomous Driving.
\newblock In \emph{Proceedings of the 2023 IEEE/RSJ International Conference on
Intelligent Robots and Systems (IROS)};
IEEE: Piscataway, NJ, USA, 2023; pp.~9340--9346.
\bibitem[Codevilla et~al.(2018)Codevilla, Müller, López, Koltun, and
Dosovitskiy]{codevilla2018ConditionalImitation}
Codevilla, F.; Müller, M.; López, A.; Koltun, V.; Dosovitskiy, A.
\newblock End-to-End Driving Via Conditional Imitation Learning.
\newblock In \emph{Proceedings of the 2018 IEEE International Conference on Robotics
and Automation (ICRA)};
IEEE: Piscataway, NJ, USA, 2018; \mbox{pp. 4693--4700.}
\newblock [\href{http://dx.doi.org/10.1109/ICRA.2018.8460487}{CrossRef}]
\bibitem[Salvucci(2006)]{salvucci2006modeling}
Salvucci, D.D.
\newblock Modeling driver behavior in a cognitive architecture.
\newblock {\em Hum. Factors} {\bf 2006}, {\em 48},~362--380. [\href{http://dx.doi.org/10.1518/001872006777724417}{CrossRef}]
\bibitem[Berntorp et~al.(2018)Berntorp, Hoang, Quirynen, and
Di~Cairano]{berntorp2018control}
Berntorp, K.; Hoang, T.; Quirynen, R.; Di~Cairano, S.
\newblock Control architecture design for autonomous vehicles.
\newblock In \emph{Proceedings of the 2018 IEEE Conference on Control Technology and
Applications (CCTA)}; IEEE: Piscataway, NJ, USA, 2018; pp. 404--411.
\bibitem[Lee et~al.(2022)Lee, Choi, Han, and Kim]{lee2022probabilistically}
Lee, H.; Choi, Y.; Han, T.; Kim, K.
\newblock Probabilistically Guaranteeing End-to-end Latencies in Autonomous
Vehicle Computing Systems.
\newblock {\em IEEE Trans. Comput.} {\bf 2022}, \emph{71}, 3361--3374. [\href{http://dx.doi.org/10.1109/TC.2022.3152105}{CrossRef}]
\bibitem[Strobel et~al.(2020)Strobel, Zhu, Chang, and
Koppula]{strobel2020accurate}
Strobel, K.; Zhu, S.; Chang, R.; Koppula, S.
\newblock Accurate, low-latency visual perception for autonomous racing:
Challenges, mechanisms, and practical solutions.
\newblock In \emph{Proceedings of the 2020 IEEE/RSJ International Conference on
Intelligent Robots and Systems (IROS)}; IEEE: Piscataway, NJ, USA, 2020; pp. 1969--1975.
\bibitem[Vedder et~al.(2018)Vedder, Vinter, and Jonsson]{vedder2018low}
Vedder, B.; Vinter, J.; Jonsson, M.
\newblock A low-cost model vehicle testbed with accurate positioning for
autonomous driving.
\newblock {\em J. Robot.} {\bf 2018}, {\em 2018}, 4907536. [\href{http://dx.doi.org/10.1155/2018/4907536}{CrossRef}]
\bibitem[Ge(2019)]{ge2019ultra}
Ge, X.
\newblock Ultra-reliable low-latency communications in autonomous vehicular
networks.
\newblock {\em IEEE Trans. Veh. Technol.} {\bf 2019}, {\em
68},~\mbox{5005--5016.} [\href{http://dx.doi.org/10.1109/TVT.2019.2903793}{CrossRef}]
\bibitem[El~Marai and Taleb(2020)]{el2020smooth}
El~Marai, O.; Taleb, T.
\newblock Smooth and low latency video streaming for autonomous cars during
handover.
\newblock {\em IEEE Netw.} {\bf 2020}, {\em 34},~302--309. [\href{http://dx.doi.org/10.1109/MNET.011.2000258}{CrossRef}]
\bibitem[Pokhrel et~al.(2021)Pokhrel, Kumar, and Walid]{pokhrel2021towards}
Pokhrel, S.R.; Kumar, N.; Walid, A.
\newblock Towards ultra reliable low latency multipath TCP For connected
autonomous vehicles.
\newblock {\em IEEE Trans. Veh. Technol.} {\bf 2021}, {\em
70},~8175--8185. [\href{http://dx.doi.org/10.1109/TVT.2021.3093632}{CrossRef}]
\bibitem[Zhang et~al.(2020)Zhang, Zhong, Cui, Ren, and Shi]{zhang2020ac4av}
Zhang, Q.; Zhong, H.; Cui, J.; Ren, L.; Shi, W.
\newblock AC4AV: A flexible and dynamic access control framework for connected
and autonomous vehicles.
\newblock {\em IEEE Internet Things J.} {\bf 2020}, {\em
8},~1946--1958. [\href{http://dx.doi.org/10.1109/JIOT.2020.3016961}{CrossRef}]
\bibitem[Gorsich et~al.(2018)Gorsich, Jayakumar, Cole, Crean, Jain, and
Ersal]{gorsich2018evaluating}
Gorsich, D.J.; Jayakumar, P.; Cole, M.P.; Crean, C.M.; Jain, A.; Ersal, T.
\newblock Evaluating mobility vs. latency in unmanned ground vehicles.
\newblock {\em J. Terramech.} {\bf 2018}, {\em 80},~11--19. [\href{http://dx.doi.org/10.1016/j.jterra.2018.10.001}{CrossRef}]
\bibitem[Yaqoob et~al.(2019)Yaqoob, Khan, Kazmi, Imran, Guizani, and
Hong]{yaqoob2019autonomous}
Yaqoob, I.; Khan, L.U.; Kazmi, S.A.; Imran, M.; Guizani, N.; Hong, C.S.
\newblock Autonomous driving cars in smart cities: Recent advances,
requirements, and challenges.
\newblock {\em IEEE Netw.} {\bf 2019}, {\em 34},~174--181. [\href{http://dx.doi.org/10.1109/MNET.2019.1900120}{CrossRef}]
\bibitem[Kalaria et~al.(2022)Kalaria, Lin, and Dolan]{kalaria2022delay}
Kalaria, D.; Lin, Q.; Dolan, J.M.
\newblock Delay-aware robust control for safe autonomous driving.
\newblock In \emph{Proceedings of the 2022 IEEE Intelligent Vehicles Symposium (IV)};
IEEE: Piscataway, NJ, USA, 2022; pp. 1565--1571.
\bibitem[Kalaria et~al.(2024)Kalaria, Lin, and Dolan]{kalaria2024delay}
Kalaria, D.; Lin, Q.; Dolan, J.M.
\newblock Delay-aware robust control for safe autonomous driving and racing.
\newblock {\em IEEE Trans. Intell. Transp. Syst.} {\bf
2024}, {\em 25},~7140--7150. [\href{http://dx.doi.org/10.1109/TITS.2023.3339708}{CrossRef}]
\bibitem[Kamtam et~al.(2024)Kamtam, Lu, Bouali, Haas, and
Birrell]{kamtam2024network}
Kamtam, S.B.; Lu, Q.; Bouali, F.; Haas, O.C.; Birrell, S.
\newblock Network latency in teleoperation of connected and autonomous
vehicles: A review of trends, challenges, and mitigation strategies.
\newblock {\em Sensors} {\bf 2024}, {\em 24},~3957. [\href{http://dx.doi.org/10.3390/s24123957}{CrossRef}]
\bibitem[Han and Kim(2023)]{han2023minimizing}
Han, T.; Kim, K.
\newblock Minimizing probabilistic end-to-end latencies of autonomous driving
systems.
\newblock In \emph{Proceedings of the 2023 IEEE 29th Real-Time and Embedded
Technology and Applications Symposium (RTAS)}; IEEE: Piscataway, NJ, USA, 2023; pp. 27--39.
\bibitem[Pomerleau(1989)]{pomerleau_alvinn_1989}
Pomerleau, D.
\newblock ALVINN: An Autonomous Land Vehicle In a Neural Network.
\newblock In \emph{Proceedings of the Advances in Neural Information Processing Systems};
Morgan Kaufmann: San Mateo, CA, USA, 1989; pp. 305--313.
\bibitem[Net-Scale(2004)]{net-scale_autonomous_2004}
Lecun, Y.; Cosatto, E.; Ben, J.; Muller, U.; Flepp, B.
\newblock Autonomous {Off}-{Road} {Vehicle} {Control} {Using} {End}-to-{End}
{Learning}. \emph{Darpa-Ipto Final. Rep.} \textbf{2004}, \emph{36}, 1.
\bibitem[Bojarski et~al.(2017)]{bojarski2017explaining}
Bojarski, M.; Yeres, P.; Choromanska, A.; Choromanski, K.; Firner, B.; Jackel, L.; Muller, U.
\newblock Explaining How a Deep Neural Network Trained with End-to-End Learning
Steers a Car.
\newblock {\em arXiv} {\bf 2017}, arXiv:1704.07911.
\bibitem[Wang et~al.(2019)Wang, Chen, Tian, Tian, Li, and
Cao]{wang_end--end_2019-1}
Wang, Q.; Chen, L.; Tian, B.; Tian, W.; Li, L.; Cao, D.
\newblock End-to-End Autonomous Driving: An Angle Branched Network Approach.
\newblock {\em IEEE Trans. Veh. Technol.} {\bf 2019}, {\em 68},~11599--11610. [\href{http://dx.doi.org/10.1109/TVT.2019.2921918}{CrossRef}]
\bibitem[Wu et~al.(2019)Wu, Luo, Huang, Cheng, and Zhao]{wu_end--end_2019}
Wu, T.; Luo, A.; Huang, R.; Cheng, H.; Zhao, Y.
\newblock End-to-End Driving Model for Steering Control of Autonomous Vehicles
with Future Spatiotemporal Features.
\newblock In \emph{Proceedings of the IEEE/RSJ International Conference on Intelligent
Robots and Systems (IROS)}; IEEE: Piscataway, NJ, USA, 2019; pp. 950--955. [\href{http://dx.doi.org/10.1109/IROS40897.2019.8968453}{CrossRef}]
\bibitem[Kwon et~al.(2022)Kwon, Khalil, Kim, and Nam]{kwon2022incremental}
Kwon, J.; Khalil, A.; Kim, D.; Nam, H.
\newblock Incremental end-to-end learning for lateral control in autonomous
driving.
\newblock {\em IEEE Access} {\bf 2022}, {\em 10},~33771--33786. [\href{http://dx.doi.org/10.1109/ACCESS.2022.3160655}{CrossRef}]
\bibitem[Weiss and Behl(2020)]{weiss2020deepracing}
Weiss, T.; Behl, M.
\newblock Deepracing: Parameterized trajectories for autonomous racing.
\newblock {\em arXiv} {\bf 2020}, arXiv:2005.05178. [\href{http://dx.doi.org/10.48550/arXiv.2005.05178}{CrossRef}]
\bibitem[Cultrera et~al.(2023)Cultrera, Becattini, Seidenari, Pala, and
Del~Bimbo]{cultrera2023addressing}
Cultrera, L.; Becattini, F.; Seidenari, L.; Pala, P.; Del~Bimbo, A.
\newblock Addressing limitations of state-aware imitation learning for
autonomous driving.
\newblock {\em IEEE Trans. Intell. Veh.} {\bf 2023}, {\em
9},~2946--2955. [\href{http://dx.doi.org/10.1109/TIV.2023.3336063}{CrossRef}]
\bibitem[Jamgochian et~al.(2023)Jamgochian, Buehrle, Fischer, and
Kochenderfer]{jamgochian2023shail}
Jamgochian, A.; Buehrle, E.; Fischer, J.; Kochenderfer, M.J.
\newblock Shail: Safety-aware hierarchical adversarial imitation learning for
autonomous driving in urban environments.
\newblock In \emph{Proceedings of the 2023 IEEE International Conference on Robotics
and Automation (ICRA)}; IEEE: Piscataway, NJ, USA, 2023; pp. 1530--1536.
\bibitem[Cheng et~al.(2024)Cheng, Chen, and Chen]{cheng2024pluto}
Cheng, J.; Chen, Y.; Chen, Q.
\newblock Pluto: Pushing the limit of imitation learning-based planning for
autonomous driving.
\newblock {\em arXiv} {\bf 2024}, arXiv:2404.14327. [\href{http://dx.doi.org/10.48550/arXiv.2404.14327}{CrossRef}]
\bibitem[Delavari et~al.(2025)Delavari, Khalil, and Kwon]{delavari2025caril}
Delavari, E.; Khalil, A.; Kwon, J.
\newblock CARIL: Confidence-Aware Regression in Imitation Learning for
Autonomous Driving.
\newblock {\em arXiv} {\bf 2025}, arXiv:2503.00783. [\href{http://dx.doi.org/10.48550/arXiv.2503.00783}{CrossRef}]
\bibitem[Mao et~al.(2019)Mao, Yang, and Dally]{mao2019delay}
Mao, H.; Yang, X.; Dally, W.J.
\newblock A delay metric for video object detection: What average precision
fails to tell.
\newblock In \emph{Proceedings of the IEEE/CVF International Conference on Computer Vision (ICCV)};
IEEE: Piscataway, NJ, USA, 2019; pp. 573--582.
\bibitem[Kocić et~al.(2019)Kocić, Jovičić, and
Drndarević]{kocic_end--end_2019}
Kocić, J.; Jovičić, N.; Drndarević, V.
\newblock An End-to-End Deep Neural Network for Autonomous Driving Designed for
Embedded Automotive Platforms.
\newblock {\em Sensors} {\bf 2019}, {\em 19}, 2064. [\href{http://dx.doi.org/10.3390/s19092064}{CrossRef}]
\bibitem[Wu et~al.(2022)Wu, Jia, Chen, Yan, Li, and Qiao]{wu2022trajectory}
Wu, P.; Jia, X.; Chen, L.; Yan, J.; Li, H.; Qiao, Y.
\newblock Trajectory-guided Control Prediction for End-to-end Autonomous
Driving: A Simple yet Strong Baseline.
\newblock {\em arXiv} {\bf 2022}, arXiv:2206.08129.
\bibitem[Popov et~al.(2024)Popov, Degirmenci, Wehr, Hegde, Oldja, Kamenev,
Douillard, Nist{\'e}r, Muller, Bhargava, et~al.]{popov2024mitigating}
Popov, A.; Degirmenci, A.; Wehr, D.; Hegde, S.; Oldja, R.; Kamenev, A.;
Douillard, B.; Nist{\'e}r, D.; Muller, U.; Bhargava, R.;  et~al.
\newblock Mitigating covariate shift in imitation learning for autonomous
vehicles using latent space generative world models.
\newblock {\em arXiv} {\bf 2024}, arXiv:2409.16663. [\href{http://dx.doi.org/10.48550/arXiv.2409.16663}{CrossRef}]
\bibitem[Tampuu et~al.(2024)Tampuu, Roosild, and Uduste]{tampuu2024effects}
Tampuu, A.; Roosild, K.; Uduste, I.
\newblock The Effects of Speed and Delays on Test-Time Performance of
End-to-End Self-Driving.
\newblock {\em Sensors} {\bf 2024}, {\em 24},~1963. [\href{http://dx.doi.org/10.3390/s24061963}{CrossRef}] [\href{http://www.ncbi.nlm.nih.gov/pubmed/38544226}{PubMed}]
\bibitem[Bojarski et~al.(2016)Bojarski, Del~Testa, Dworakowski, Firner, Flepp,
Goyal, Jackel, Monfort, Muller, Zhang, et~al.]{bojarski2016end}
Bojarski, M.; Del~Testa, D.; Dworakowski, D.; Firner, B.; Flepp, B.; Goyal, P.;
Jackel, L.D.; Monfort, M.; Muller, U.; Zhang, J.;  et~al.
\newblock End to end learning for self-driving cars.
\newblock {\em arXiv} {\bf 2016}, arXiv:1604.07316. [\href{http://dx.doi.org/10.48550/arXiv.1604.07316}{CrossRef}]
\bibitem[Kwon(2021)]{kwon_oscar_2021}
Kwon, J.
\newblock OSCAR: Open-Source Robotic Car Architecture for Research and Education.
\newblock Available online: \url{https://github.com/jrkwon/oscar} (accessed on 8 March 2026).
\bibitem[Quigley et~al.(2009)Quigley, Conley, Gerkey, Faust, Foote, Leibs,
Wheeler, Ng]{quigley2009ros}
Quigley, M.; Conley, K.; Gerkey, B.; Faust, J.; Foote, T.; Leibs, J.; Wheeler,
R.; Ng, A.Y.
\newblock ROS: An open-source Robot Operating System.
\newblock  \emph{ICRA Workshop Open Source Softw.}
\textbf{2009}, \emph{3}, 5.
\bibitem[Koenig and Howard(2004)]{koenig2004design-gazebo}
Koenig, N.; Howard, A.
\newblock Design and use paradigms for gazebo, an open-source multi-robot
simulator.
\newblock In \emph{Proceedings of the 2004 IEEE/RSJ International Conference on
Intelligent Robots and Systems (IROS)}; IEEE: Piscataway, NJ, USA, 2004; Volume~3,
pp.~2149--2154.
\bibitem[Dosovitskiy et~al.(2017)Dosovitskiy, Ros, Codevilla, Lopez, and
Koltun]{Dosovitskiy17}
Dosovitskiy, A.; Ros, G.; Codevilla, F.; Lopez, A.; Koltun, V.
\newblock CARLA: An Open Urban Driving Simulator.
\newblock In \emph{Proceedings of the 1st Annual Conference on Robot Learning (CoRL)};
PMLR: New York, NY, USA, 2017; pp. 1--16.
\bibitem[{NVIDIA Corporation}(2024)]{nvidia_drive_sim}
{NVIDIA Corporation}.
\newblock NVIDIA DRIVE Sim. 2024.
\newblock  Available online: \url{https://developer.nvidia.com/drive/simulation}  (accessed on 9 May 2024).
\bibitem[Jekel et~al.(2019)Jekel, Venter, Venter, Stander, and
Haftka]{jekel2019similarity}
Jekel, C.F.; Venter, G.; Venter, M.P.; Stander, N.; Haftka, R.T.
\newblock Similarity measures for identifying material parameters from
hysteresis loops using inverse analysis.
\newblock {\em Int. J. Mater. Form.} {\bf 2019}, {\em
12},~355--378. [\href{http://dx.doi.org/10.1007/s12289-018-1421-8}{CrossRef}]
\bibitem[Kim et~al.(2023)Kim, Khalil, Nam, and Kwon]{kim2023opemi}
Kim, D.; Khalil, A.; Nam, H.; Kwon, J.
\newblock OPEMI: Online Performance Evaluation Metrics Index for Deep
Learning-Based Autonomous Vehicles.
\newblock {\em IEEE Access} {\bf 2023}, {\em 11},~16951--16963. [\href{http://dx.doi.org/10.1109/ACCESS.2023.3246104}{CrossRef}]
\bibitem[Kingma and Ba(2014)]{kingma2014adam}
Kingma, D.P.; Ba, J.
\newblock Adam: A method for stochastic optimization.
\newblock {\em arXiv} {\bf 2014}, arXiv:1412.6980.
\bibitem[Graves and Graves(2012)]{graves2012long}
Graves, A.; Graves, A.
\newblock Long short-term memory.
\newblock In {\em Supervised Sequence Labelling with Recurrent Neural Networks};
Springer: Berlin/Heidelberg, Germany, 2012; pp. 37--45.
\end{thebibliography}
\end{document}